# Multi-Task Learning for Argumentation Mining

**Multi-Task Learning für Argumentation Mining**
Master-Thesis von Tobias Kahse aus Langen (Hessen)
Tag der Einreichung:

1. Gutachten: Dr. Steffen Eger
2. Gutachten: Prof. Dr. Iryna Gurevych

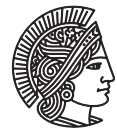

Computer Science Department
Ubiquitous Knowledge Processing

Multi-Task Learning for Argumentation Mining
Multi-Task Learning für Argumentation Mining

Vorgelegte Master-Thesis von Tobias Kahse aus Langen (Hessen)

1. Gutachten: Dr. Steffen Eger
2. Gutachten: Prof. Dr. Iryna Gurevych

Tag der Einreichung:

# Erklärung zur Master-Thesis

Hiermit versichere ich, die vorliegende Master-Thesis ohne Hilfe Dritter nur mit den angegebenen Quellen und Hilfsmitteln angefertigt zu haben. Alle Stellen, die aus Quellen entnommen wurden, sind als solche kenntlich gemacht. Diese Arbeit hat in gleicher oder ähnlicher Form noch keiner Prüfungsbehörde vorgelegen.

Darmstadt, den 28. September 2017

______________________________
(Tobias Kahse)

# Abstract

Multi-task learning has recently become a very active field in deep learning research. In contrast to learning a single task in isolation, multiple tasks are learned at the same time, thereby utilizing the training signal of related tasks to improve the performance on the respective machine learning tasks.

Related work shows various successes in different domains when applying this paradigm and this thesis extends the existing empirical results by evaluating multi-task learning in four different scenarios: argumentation mining, epistemic segmentation, argumentation component segmentation, and grapheme-to-phoneme conversion. We show that multi-task learning can, indeed, improve the performance compared to single-task learning in all these scenarios, but may also hurt the performance.

Therefore, we investigate the reasons for successful and less successful applications of this paradigm and find that dataset properties such as entropy or the size of the label inventory are good indicators for a potential multi-task learning success and that multi-task learning is particularly useful if the task at hand suffers from data sparsity, i.e. a lack of training data. Moreover, multi-task learning is particularly effective for long input sequences in our experiments. We have observed this trend in all evaluated scenarios.

Finally, we develop a highly configurable and extensible sequence tagging framework which supports multi-task learning to conduct our empirical experiments and to aid future research regarding the multi-task learning paradigm and natural language processing.

# Zusammenfassung

Im Forschungsfeld Deep Learning zeigt sich Multi-Task Learning in letzter Zeit als sehr aktives Teilgebiet. Anstatt ein einziges Problem komplett isoliert zu lösen, werden mehrere Probleme gleichzeitig durch ein Machine Learning System gelöst. Dadurch werden Reize aus dem Lernprozess verwandter Probleme beim Lernprozess eines Problems genutzt, um eine bessere Lösung für dieses Problem zu finden.

In der Literatur konnte dieses Paradigma bereits mehrfach erfolgreich in verschiedenen Domänen angewandt werden. Diese Thesis ergänzt die bisher erlangten empirischen Ergebnisse, indem der Multi-Task Learning Ansatz in vier weiteren Szenarien evaluiert wird: Argumentation Mining, epistemische Segmentierung, Segmentierung von Argumentationskomponenten und Konvertierung von Graphemen zu Phonemen. Wir zeigen, dass Mutli-Task Learning tatsächlich in allen Szenarien bessere Ergebnisse als Single-Task Learning erreichen kann, es jedoch auch möglich ist, dass sich die Ergebnisse verschlechtern.

Aus diesem Grund ermitteln wir die Ursachen für einen Erfolg oder Misserfolg bei der Anwendung von Multi-Task Learning. Dabei finden wir heraus, dass Eigenschaften von Datensätzen wie zum Beispiel Entropie oder die Anzahl an verfügbaren Labels gute Idikatoren für den möglichen Erfolg von Multi-Task Learning sind. Zudem erkennen wir, dass Multi-Task Learning besonders nützlich ist, wenn das zu lösende Machine Learning Problem an einem Mangel an Trainingsdaten leidet. Des Weiteren haben wir bei allen evaluierten Szenarios beobachtet, dass Multi-Task Learning gerade bei langen Eingabesequenzen effektiv ist.

Zuletzt haben wir noch ein in hohem Maße konfigurierbares und erweiterbares Sequence Tagging Framework, welches Multi-Task Learning unterstützt, entwickelt, um unsere empirischen Experimente durchzuführen und die zukünftige Forschung im Bereich Multi-Task Learning und Natural Language Processing zu unterstützen.

# Contents

# List of Figures

# List of Tables

# Acronyms

**ACS**  Argument Component Segmentation

**ACI**  Argument Component Identification

**ARS**  Argument Relation Segmentation

**ARI**  Argument Relation Identification

**ADB**  AraucariaDB

**AM**  Argumentation Mining

**API**  Application Programming Interface

**BPTT**  Backpropagation Through Time

**CCG**  Combinatory Categorial Grammar

**CNN**  Convolutional Neural Network

**CoNLL**  Conference on Computational Natural Language Learning

**CRF**  Conditional Random Field

**CSV**  Comma Separated Values

**ED**  Edit Distance

**EDU**  Elementary Discourse Unit

**ES**  Epistemic Segmentation

**TS**  Teacher Students

**SW**  Social Workers

**MED**  Medicine

**G2P**  Grapheme-to-Phoneme

**GPU**  Graphics Processing Unit

**GRU**  Gated Recurrent Unit

**HMM**  Hidden Markov Model

**ILP**  Integer Linear Programming

**IPA**  International Phonetic Alphabet

**KBC**  Keyphrase Boundary Detection

**LSTM**  Long-Short Term Memory

**NLP**  Natural Language Processing

**MEMM**  Maximum-Entropy Markov Model

**MLP**  Multi-Layer Perceptron

**MTL**  Multi-Task Learning

**NER**  Named Entity Recognition

**NLTK**  Natural Language Toolkit

**OOV**  Out-of-Vocabulary

**PE**  Persuasive Essays

**PE:ACS**  Persuasive Essays

**PCN**  POS tagging, syntactic chunking, and NER

**PDTB**  Penn Discourse Treebank

**POS**  Part-Of-Speech

**PTB**  Penn Treebank

**RNN**  Recurrent Neural Network

**RST**  Rhetorical Structure Theory

**RST-DT**  Rhetorical Structure Theory Discourse Treebank

**S2S**  Sequence-to-Sequence

**SRA**  Scientific Reasoning and Argumentation

**STL**  Single-Task Learning

**TUD**  Technische Universität Darmstadt

**UKP**  Ubiquitous Knowledge Processing

**WACC**  Word Accuracy

**WSJ**  Wall Street Journal

**WTP**  Wikipedia Talk Pages

**YAML**  YAML Ain't Markup Language

# 1 Introduction

## 1.1 Motivation

First introduced by Caruana (1993), Multi-Task Learning (MTL) has recently become a very active field in deep learning research. In contrast to learning a single task in isolation, multiple tasks are learned at the same time. The motivation for this approach is that neural networks benefit from sharing parameters between tasks similarly to humans leveraging knowledge from related problems to find the solution for a new one (Caruana 1993; Ruder 2017).

While there are only a few theoretical findings w.r.t. MTL (e.g. Bingel and Søgaard 2017; Ruder et al. 2017), many empirical results in various fields such as Natural Language Processing (NLP) (e.g. Collobert et al. 2011), pharmaceutics (e.g. Ramsundar et al. 2015), and computer vision (e.g. T. Zhang et al. 2012; C. Zhang and Z. Zhang 2014) show that performance improvements can be achieved with MTL.

Argumentation Mining (AM) is a relatively new field in NLP that "focuses on the the analysis of arguments in natural language texts" (Stab and Gurevych 2017). Among other contributions, the MTL paradigm is applied to AM in this thesis.

The goal of this thesis is the application of MTL in different scenarios and thereby a further investigation of whether, when, how, and why MTL is beneficial for a machine learning system. In particular, MTL is applied to tasks for which MTL so far has been used only to a small degree or not at all.

The results shall extend the state-of-the-art knowledge in this research field with more empirical evidence for the success or failure of MTL and guide future work with recommendations and insights.

Moreover, we develop a novel, highly-configurable, and extensible implementation that utilizes the state-of-the-art MTL techniques with the purpose of enabling further empirical investigations and aiding future research in the MTL research field by providing an implementation upon which new techniques can be built.

## 1.2 Contributions

The contributions of this thesis are as follows:

- **Overview of the status quo of MTL**: MTL recently gained a lot of traction. The thesis provides a survey of MTL and the latest advances in this research field with a focus on NLP.

- **Generic MTL-capable sequence tagging framework**: A major contribution of this thesis is the development of a highly configurable and thereby flexible and generic sequence tagging framework which utilizes the MTL paradigm.

- **Empirical evaluation of the MTL paradigm** Using the developed sequence tagging framework and architectures from related work, we evaluate the MTL paradigm empirically by means of experiments with multiple NLP tasks and MTL variations.

- **Analysis of the MTL paradigm and best practice recommendations**: Finally, we analyze the MTL paradigm with the given experiment results and the findings from related work.

## 1.3 Thesis Structure

This thesis consists of eight chapters. The next chapter provides an overview of related work which, on the one hand, briefly introduces related topics that are not in the scope of this thesis and, on the other hand, provides a survey of the MTL literature and a short introduction to AM.

The third chapter, then, presents theoretical background knowledge which is necessary to understand the thesis. The section on MTL in this chapter builds upon the related work chapter.

In chapter four, we describe our MTL sequence tagging framework and the fifth chapter discusses the conducted experiments by listing and explaining the used corpora, describing the experiment setups, and presenting the experiments' results.

Chapter six, afterwards, continues with a discussion of the experiment results which includes an analysis of the MTL paradigm within the scope of the conducted experiments.

Finally, the seventh chapter summarizes the thesis' results and the eighth chapter discusses opportunities for future work.

# 2 Related Work

Our work is closely related to three fields of research outlined in the following.

## 2.1 Sequence Tagging

*Sequence tagging* or *sequence labeling* refers to the task of finding an optimal label sequence $l = l_1, \ldots, l_n$ for an input sequence $t = t_1, \ldots, t_n$ of $n$ tokens where the labels $l_i$ are from a predefined label set (Z. Li et al. 2015).

In the field of NLP, solutions to the generic problem of *sequence tagging* can be applied to many specific problems such as Part-Of-Speech (POS) tagging, chunking, Named Entity Recognition (NER), error detection in learner writing, and spelling normalization (Cuong et al. 2014; Z. Huang, Wei Xu, and K. Yu 2015; Bollmann and Søgaard 2016; Ma and E. Hovy 2016; Rei, Crichton, and Pyysalo 2016; Rei and Yannakoudakis 2016; Yang, Salakhutdinov, and Cohen 2016).

### 2.1.1 Non-Neural Network Approaches

Non-neural network approaches to sequence tagging are usually based on linear statistical models such as Hidden Markov Models (HMMs) (Rabiner 1989) and Conditional Random Fields (CRFs) (Lafferty, McCallum, and Pereira 2001). These approaches traditionally rely on feature engineering (Collobert et al. 2011) and task-, domain-, and/or language-specific resources (Lample et al. 2016).

Particularly interesting for this thesis are CRFs because they can also be used with neural network approaches. This will be discussed in the next section.

Z. Li et al. (2015) use a CRF-based model for Chinese POS tagging. They leverage data from multiple corpora with heterogeneous labels by jointly predicting the heterogeneous annotations and thereby improve the performance for Chinese POS tagging. This approach is also related to MTL in that it jointly learns to predict tags for the different corpora.

While CRFs are theoretically capable of modeling dependencies between arbitrarily many consecutive labels ($n^{\text{th}}$-order), *usually* only first-order dependencies, i.e. dependencies between adjacent labels, are modeled in practice. This is due to the high complexity of inference for CRFs. Müller, Schmid, and Schütze (2013) and Cuong et al. (2014), however, discuss implementations of higher-order CRFs for sequence tagging. Both introduced techniques to reduce the computation time for training higher-order CRFs in order to make it feasible to use them.

Müller, Schmid, and Schütze (2013) reduce the computation time by pruning candidate states to prevent a polynomial increase of the state space. Cuong et al. (2014) exploit the "label pattern sparsity", i.e. the number of observed patterns within sequences of consecutive labels is much smaller than the number of possible label patterns. This sparsity is often encountered in real data. Moreover, the authors show that using higher-order CRFs is especially beneficial for data with long range dependencies.

### 2.1.2 Neural Network Approaches

In neural networks for sequence tagging, Recurrent Neural Networks (RNNs) are prevalent (e.g. Z. Huang, Wei Xu, and K. Yu 2015; Yang, Salakhutdinov, and Cohen 2016; John 2017). In particular, Long-Short Term Memory (LSTM) cells are used frequently (e.g. Lample et al. 2016; Ma and E. Hovy 2016; Rei and Yannakoudakis 2016), but Gated Recurrent Units (GRUs) are used as well (Yang, Salakhutdinov, and Cohen 2016). Both techniques allow to model long-range dependencies in input sequences. LSTM cells, however, have a considerably more complex structure. More detailed information is provided in section 3.1.3. Usually, RNNs do not only read the input sequence from left to right, but also from right to left, i.e. the RNNs are bidirectional (e.g. Lample et al. 2016).

Non-neural network approaches often rely on feature engineering and task-specific resources (e.g. McCallum and W. Li 2003; Müller, Schmid, and Schütze 2013; Z. Li et al. 2015). When using neural networks, many authors focus on being independent of such measures. Lample et al. (2016) and Ma and E. Hovy (2016) incorporate character-level information in their networks by using a bidirectional LSTM and a Convolutional Neural Network (CNN) respectively. This information is highly beneficial when the network encounters Out-of-Vocabulary (OOV) words, for example.

Rei, Crichton, and Pyysalo (2016) further improve the use of character-level information by introducing a novel, attention-based approach for combining word embeddings and character level information. Instead of concatenating word- and character-level information as it is done in Lample et al. (2016), for instance, they build the weighted sum of word vector and character representation. This allows the network to dynamically define the mixing ratio between both representations.

Since sequence tagging tasks frequently have label sets where labels may depend on each other, e.g. in chunking and NER, using a CRF layer on top of the neural network has become as prevalent as using bidirectional RNNs. For example, Z. Huang, Wei Xu, and K. Yu (2015), Lample et al. (2016), Ma and E. Hovy (2016), and Yang, Salakhutdinov, and Cohen (2016) all report that they achieve the best results in their experiments when using a CRF layer.

Compared to the previously presented papers, Strubell et al. (2017) have a quite different take on sequence tagging because their goal is developing fast and resource-efficient methods for sequence tagging which still achieve competitive prediction performance. They use a CNN which allows them to better exploit parallelization with Graphics Processing Units (GPUs). Moreover, the authors make use of "dilated convolutions" which are able to skip over inputs without a loss in resolution. This results in even faster networks, i.e. networks that require less computation time for training and prediction.

## 2.2 Multi-Task Learning (MTL)

Although there are non-neural network systems which use the MTL paradigm (Ruder 2017), only neural network approaches will be discussed in the following because non-neural MTL is not in the scope of this thesis.

### 2.2.1 Use Cases

MTL has been introduced by Caruana (1993). In his paper, he lists reducing the need for codifying domain experts' knowledge and improving the generalization capability of a system as major use cases. Later, he describes eight prototypical uses of MTL, e.g. using extra tasks to focus attention to details in the training data that might otherwise be overlooked (Caruana 1996). In his survey on MTL, Ruder (2017) revisits these use cases and adds "eavesdropping", i.e. learning features that are difficult to learn for one task through another task that learns these features more easily. For instance, J. Yu and Jiang (2016) predict whether a sentence contains a positive or negative domain-independent sentiment word as an auxiliary task to aid their main task which is sentence-level sentiment classification. Learning to find occurrences of such sentiment words is much easier for the auxiliary task than for the main tasks because it is only focused on domain-independent sentiment words.

In MTL literature, the most stated motivation for using the MTL paradigm is *fighting data sparseness*. Data sparseness refers to a lack of data for the machine-learning task at hand. Benton, M. Mitchell, and D. Hovy (2017) conclude that MTL is especially beneficial for tasks with very sparse data. Augenstein and Søgaard (2017), for example, improve the performance of Keyphrase Boundary Detection (KBC) by leveraging training data from tasks related to KBC. Since KBC is an uncommon task in NLP, data sparsity affects it particularly.

Instead of utilizing data from related tasks, Bollmann and Søgaard (2016) and N. Peng and Dredze (2016) use data from different domains, but for the same task. Thereby, they are able to achieve improvements in the tasks of historical spelling normalization (Bollmann and Søgaard 2016) and Chinese word segmentation and named entity recognition (N. Peng and Dredze 2016).

Similarly, one can use data from different languages as auxiliary tasks. Plank, Søgaard, and Goldberg (2016) use this for POS tagging across 22 languages including English, French, German, Hebrew, and Finnish. Yang, Salakhutdinov, and Cohen (2016) utilize this approach for NER in English, Dutch, and Spanish.

Braud, Lacroix, and Søgaard (2017) even leverage both, data from different domains and languages, in order to improve the performance for discourse segmentation.

While empirical results suggest that MTL in NLP is not only beneficial for tasks that are syntactic in nature such as chunking, but also for semantic problems (e.g. Liu, Qiu, and X. Huang 2016; Augenstein and Søgaard 2017; Braud, Lacroix, and Søgaard 2017; Hashimoto et al. 2017), Alonso and Plank (2017) find that in four out of five semantic tasks in their empirical study MTL is not able to substantially improve the performance and even decreases the performance for some problems.

They identify differences in dataset properties including dataset size, entropy, and kurtosis[1] between syntactic and semantic tasks as the main cause for these results.

### 2.2.2 Network Architecture

While many MTL network architectures feed their task classifiers from the outermost layer of the shared layers (e.g. Collobert et al. 2011; Luong et al. 2016; Yang, Salakhutdinov, and Cohen 2016), Søgaard and Goldberg (2016) propose modeling low-level tasks at lower levels of the shared layers. In their experiments, they obtain performance improvements due to adjusting the network architecture based on the "natural order among the different tasks". The authors use POS tagging as a lower level task to improve the performance of chunking and Combinatory Categorial Grammar (CCG) supertagging. CCG refers to the task of assigning lexical categories, so-called supertags, to tokens. Since there are significantly more lexical categories than POS tags, this assignment is more challenging than POS tagging. Tagging tokens with supertags by treating it as a sequence tagging problem is referred to as supertagging (Wenduan Xu, Auli, and S. Clark 2015).

Hashimoto et al. (2017) extend this idea by building hierarchical architecture for five different NLP tasks where each successive task feeds from a deeper layer. In addition to Søgaard and Goldberg (2016), they use their network not only for syntactic tasks (POS tagging, chunking, and dependency parsing), but also for semantic tasks (semantic relatedness and textual entailment).

Furthermore, they implement more sophisticated architectural features such as "shortcut connections", i.e. connecting the embedding layer with each shared layer instead of just the first one, "output label embeddings", i.e. feeding classifier output of lower layers into deeper layers, and utilizing character-level information.

H. Peng, Thomson, and Smith (2017) do not feed their classifiers directly from the shared layers, but rather use several independent layers in between. They achieve performance improvements for semantic dependency parsing with their MTL architectures.

### 2.2.3 Parameter Sharing

Parameter sharing is crucial in MTL (Ruder 2017). Liu, Qiu, and X. Huang (2016) take a closer look on how to share parameters in an MTL setup. They propose extending LSTM cells with external memory in addition to the existing internal memory. The external memory is then shared between tasks while the internal memory is not. The authors find that this separation is always beneficial as it enables the network to differentiate between shared and task-specific features.

---
[1] Kurtosis is a measure for the skewness of a distribution. This measure is explained in section 3.3.5.

This idea is further refined in Liu, Qiu, and X. Huang (2017). While the previous model allows the network to have task-specific and task-invariant feature spaces, there is no guarantee for the purity of these spaces. To encourage more pure spaces, the authors introduce adversarial loss and orthogonality constraints which prevent task-specific features from being in the task-invariant feature space and vice versa.

In both aforementioned papers, the authors perform binary sentiment classification on the document level for movie and product reviews, i.e. one review is one document. They evaluate their systems on 16 review domains, always using one domain as the main task and the remaining 15 domains as auxiliary tasks.

Recently, Ruder et al. (2017) proposed "Sluice Networks", an MTL architecture with shared and private feature spaces as in Liu, Qiu, and X. Huang (2016), but with the added benefit of trainable parameters which control the amount of sharing. That is, the parameters control for each layer which of the features are considered private and which can be shared between tasks.

Moreover, further parameters automatically determine which features from which shared layers feed each task classifier, thereby learning hierarchical relations rather than hardwiring them in the network architecture as in Søgaard and Goldberg (2016).

### 2.2.4 Auxiliary Tasks

For a successful application of the MTL paradigm, choosing the right auxiliary tasks is essential (Benton, M. Mitchell, and D. Hovy 2017).

Both, Caruana (1996) and Ruder (2017) present different types of auxiliary tasks such as learning the same task with a different error metric or leveraging the expertise of a domain expert by letting her define an auxiliary task by means of examples rather than having to codify this human expertise in the machine learning algorithm. While analyzing the prerequisites for an effective application of MTL, Alonso and Plank (2017) find that it is better to use auxiliary tasks with compact and more uniform label distributions. For instance, they compare the effect of using different POS tagging datasets as auxiliary tasks and observe that the dataset with the smallest label inventory systematically outperforms the other datasets.

Bingel and Søgaard (2017) investigate which relations between main and auxiliary tasks are beneficial for MTL. Their results support the finding of Alonso and Plank (2017). Moreover, they conclude that it is possible to predict MTL gains from characteristics of the datasets and the Single-Task Learning (STL) curves of the auxiliary tasks. The latter are particularly good indicators because MTL often helps the main task to escape local minima.

Ramsundar et al. (2015) show continuous performance gains when adding more and more auxiliary tasks for their experiments in the drug discovery domain. They use up to 259 tasks and conclude that increasing the total amount of data as well as having more tasks both contribute substantially to the MTL effect. Their architecture is a pyramidal multi-layer neural network with an individual softmax classifier for each task. There is no hierarchy among the tasks and the number of units in the layers decreases with each layer, thereby forming a pyramid.

Finally, Kaiser et al. (2017) present an MTL architecture which jointly learns various, conceptually different tasks including syntactic chunking, machine translation, image captioning, and speech recognition. In contrast to other related work in which the tasks are selected carefully w.r.t. to their relatedness or suitability as auxiliary tasks (e.g. Hashimoto et al. 2017), the authors continuously extend their model to further tasks and find that the performance on the existing tasks is never decreased and frequently increased. Since their system operates on different modalities, e.g. images and sound waves, they use sub-networks which suit their respective modality best, e.g. CNNs for images, to map the input into a unified representation space and also to convert this unified representation to an output. In the unified representation space, they rely on several techniques such as CNN and an attention mechanism that have proven successful in various domains.

### 2.2.5 Natural Subtasks

*Natural subtasks* are a special type of auxiliary task that is directly derived from the main task. Braud, Plank, and Søgaard (2016) refer to using such subtasks as "multi-view training" because it utilizes different views on the same data.

For instance, they extract subtasks from the complex Rhetorical Structure Theory (RST) label by removing one label information, thus resulting in tasks with fewer and simpler labels. Furthermore, the labels become less sparse, i.e. there are less labels with only few occurrences. Similarly, Eger, Daxenberger, and Gurevych (2017) derive subtasks from complex AM labels.

Ruder (2017) refers to this type of auxiliary task as "hints". He claims that solving the subtask is beneficial for extracting features which are difficult to learn when using solely the main task.

## 2.3 Argumentation Mining (AM)

As mentioned in section 1.1, AM is a relatively new field in NLP. Hence, a lot of related work revolves around creating new corpora. For instance, Kirschner, Eckle-Kohler, and Gurevych (2015) create a corpus for AM in scientific publications. The corpus by Reed et al. (2008) contains documents from newspapers, parliamentary records, judicial summaries, and discussion boards which have been annotated with AM labels.

Stab and Gurevych (2017) build a corpus of persuasive student essays. Moreover, they provide an annotation scheme that is derived from argumentation theory and implement a parser for argumentation structures. The latter relies on a pipeline approach which combines several subtasks of AM that are trained separately and recombined at successive steps in the training process.

A similar, but less holistic model is proposed by Persing and Ng (2016) for an earlier version of the persuasive essays corpus by Stab and Gurevych (2017). Both models use Integer Linear Programming (ILP) to model the different tasks of AM, e.g. argument component identification, jointly which is beneficial as they are interdependent.

While the previously presented models in this section all use non-neural network approaches, Eger, Daxenberger, and Gurevych (2017) propose neural network-based end-to-end models for AM. They investigate performance differences when framing AM as a dependency parsing or a sequence tagging problem and find that the latter is more suitable.

Furthermore, their neural networks outperform the previous pipeline approaches, but do not require feature- or constraint-engineering. The authors use the architectures by Ma and E. Hovy (2016) and Miwa and Bansal (2016). In addition to that, they successfully apply MTL to AM with the architecture of Søgaard and Goldberg (2016) and using natural subtasks of AM.

Daxenberger et al. (2017) analyze the differences of the concept of a claim across AM corpora. Their results show that there is no consensus on the conceptualization of a claim and the annotation schemes vary from corpus to corpus. These differences impede cross-domain training in the field of AM to a certain degree.

Related work in the field of AM not only attempts to model the argumentation structure within a natural language text (e.g. Stab and Gurevych 2017), but, among other tasks, also strives to assess the quality of arguments (Habernal and Gurevych 2016), check the verifiability of a proposition (Park and Cardie 2014), and score essays (Farra, Somasundaran, and Burstein 2015).

It is particularly difficult to perform AM in user-generated web discourse due to the variety of domains, lack of formal structures in contrast to student essays (Stab and Gurevych 2017), and noisy language, e.g. spelling and grammatical errors and use of colloquialisms. This application area is, among others, investigated by Habernal, Eckle-Kohler, and Gurevych (2014), Park and Cardie (2014) and Habernal and Gurevych (2017). Habernal, Eckle-Kohler, and Gurevych (2014) motivate this by attempting to help the user in finding the most relevant information regarding a controversy and thereby reducing the information overload experienced by users in the web. In this thesis, however, AM is limited to student essays.

# 3 Theory

This chapter provides theoretical background knowledge for the upcoming chapters. For an introduction to neural networks, their mechanics, and how to use them in an NLP context refer to the works by Goodfellow, Bengio, and Courville (2016) and Goldberg (2017).

## 3.1 Recurrent Neural Networks

Section 2.1.2 already introduced RNNs as the network architecture of choice for modeling sequence tagging problems. The theoretical foundation for this choice, challenges of using them, and variations of the RNN cell will be discussed in the remainder of this section.

### 3.1.1 Introduction

RNNs are specialized at processing sequences of values $x^{(1)}, \ldots, x^{(T)}$ which may be of arbitrary length. In fact, they model sequentiality explicitly (Lipton, Berkowitz, and Elkan 2015). Each value is considered a *time step* within a sequence of time steps. By sharing parameters between these time steps, a learned model can be applied to different input lengths and generalize across them. This parameter sharing is implemented by calculating the output $h^{(t)}$ as a function of all previous outputs (Goodfellow, Bengio, and Courville 2016, pp. 373–378). This is shown in equation (3.1) where $x^{(t)}$ is the input at time step $t$ for $1 \leq t \leq T$ and $\theta$ is a set of parameters.

$$h^{(t)} = f(h^{(t-1)}, x^{(t)}; \theta) \qquad (3.1)$$

When thinking of the neural network as a graph, this formulation allows cyclical connections due to the recursion in equation (3.1). To remove these cycles, the graph can be "unfolded" in time, i.e. along the input sequence (Graves 2012, p. 20). Figure 3.1 visualizes the process of unfolding the network and explicitly shows the weight sharing between time steps.

Unfolding the network in time or, more specifically, "creating a copy of the model for each time step" (Pascanu, Mikolov, and Bengio 2013) is also helpful for training the network because the RNN can be seen as a very deep neural network without cycles and thus Backpropagation Through Time (BPTT) (Werbos 1990) can be applied (Goldberg 2016).

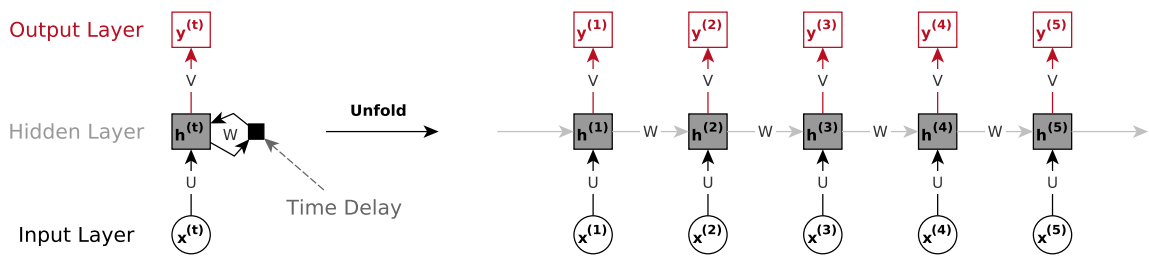

**Figure 3.1.:** The RNN with a recurrent connection ("time delay"; black rectangle) is unfolded in time resulting in a graph that maps the input sequence $x^{(1)} \ldots x^{(5)}$ to the output sequence $y^{(1)} \ldots y^{(5)}$. The weights $U, V,$ and $W$ are shared between the different inputs of the sequence.

RNNs are particularly useful due to their capability of memorizing previous inputs through the aforementioned parameter sharing and thereby influencing the network output in subsequent steps. Hence, structured properties in the input are taken into account and statistical regularities can be captured (Goldberg 2017).

Since they do not rely on a finite-length context window, RNNs are capable of modeling even dependencies between distant input signals which cannot be handled by other architectures with a fixed window size such as feedforward networks (Lipton, Berkowitz, and Elkan 2015).

Moreover, RNNs are more feasible w.r.t. their computation requirements when modeling longer sequences with dependencies compared to other sequential models including HMMs (Graves, Wayne, and Danihelka 2014; Lipton, Berkowitz, and Elkan 2015).

Goodfellow, Bengio, and Courville (2016, pp. 378–381) elaborate on RNN design patterns. In general, one can distinguish between two major architecture decisions: either the network produces an output at each time step as shown in figure 3.1 or the network reads the entire sequence to perform a single prediction for the complete sequence. The latter approach is often used in sentence classification (e.g. Balikas, Moura, and Amini 2017) while the former is especially suitable for sequence tagging as it maps a token sequence (input) to a label sequence (output) of the same length.

RNNs cannot only be used to incorporate past context into a prediction, but also future context. This is achieved by using two separate recurrent layers one of which reads the input sequence while the other reads it in reverse. The outputs of both RNNs are concatenated at each time step (Graves 2012; Goodfellow, Bengio, and Courville 2016; Goldberg 2016). *Bidirectional RNNs* have proven to be highly effective in sequence tagging tasks (cf. section 2.1.2).

### 3.1.2 Challenges

The major challenge in training RNNs is handling *long-range dependencies* (Graves 2012; Pascanu, Mikolov, and Bengio 2013; Lipton, Berkowitz, and Elkan 2015; Goodfellow, Bengio, and Courville 2016; Goldberg 2016). Bengio, Simard, and Frasconi (1994) describe a long-range dependency as follows:

> *A task displays long-term dependencies if prediction of the desired output at time $t$ depends on input presented at an earlier time $\tau \ll t$.*

Gradient-based techniques for training the network lose efficiency when the dependencies in the task become more distant (Bengio, Simard, and Frasconi 1994). This is attributed to two phenomena which may occur when training an RNN: *vanishing* and *exploding gradients* (Pascanu, Mikolov, and Bengio 2013). Both are caused by backpropagating errors across many time steps (Lipton, Berkowitz, and Elkan 2015).

The vanishing gradient problem is visualized in figure 3.2. It refers to the behavior that long-term components in the gradient go exponentially fast to norm 0 (Pascanu, Mikolov, and Bengio 2013) preventing the RNN from learning correlations between distant inputs.

Similarly, the exploding gradient problem refers to an exponential increase in the long-term components of the gradient causing its norm to explode (Pascanu, Mikolov, and Bengio 2013). This results in weight updates which render the network's weights useless (Mikolov 2012).

Pascanu, Mikolov, and Bengio (2013) suggest normalizing the gradient with its norm if the norm exceeds a specified threshold. They refer to this process as *norm clipping*. Other techniques for coping with exploding gradients are, for instance, truncating gradient components element-wise (Mikolov 2012) and using an L1 or L2 penalty on the recurrent weights (Pascanu, Mikolov, and Bengio 2013).

Which of both problems occurs or whether they occur at all depends upon various factors, e.g. the choice of activation function (Lipton, Berkowitz, and Elkan 2015). Pascanu, Mikolov, and Bengio (2013) analyze both problems and their causes in greater detail.

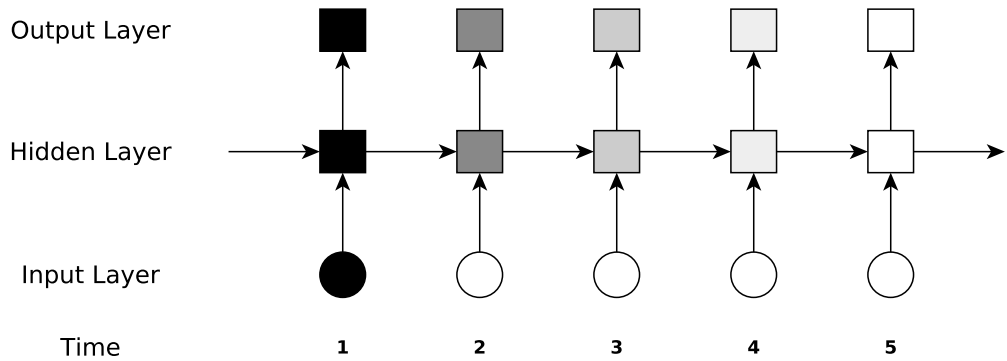

**Figure 3.2.:** Visualization of the vanishing gradient problem for RNNs adapted from Graves (2012). The shading of the nodes represents their sensitivity to the first input signal. The sensitivity decays (indicated by lighter shading) along the input sequence due to new inputs overwriting the activations of the hidden layer.

Both, however, hinder standard RNN architectures from truly taking into account a context window of arbitrary length in practice (Graves 2012). Therefore, more advanced architectures have been proposed which will be discussed in the next section.

### 3.1.3 Gated Architectures

In order to overcome the challenges, in particular vanishing gradients, presented in the previous section, more advanced recurrent architectures have been designed (Lipton, Berkowitz, and Elkan 2015).

Gated RNNs are able to learn when to use the state from a previous time step, when to ignore it, and when to reset the current state (Goodfellow, Bengio, and Courville 2016, pp. 408–409).

Besides gated architectures, other techniques for solving the challenges of RNNs have been proposed (e.g. Le, Jaitly, and Hinton 2015), but will not be discussed further because they are not relevant for this thesis.

**Long-Short Term Memory (LSTM)**

One of the oldest and arguably most used (cf. section 2.1.2) gated architecture is the LSTM architecture by Hochreiter and Schmidhuber (1997). It has been specifically designed to overcome the issues which arise when training an RNN with BPTT.

The main idea of their proposal is replacing the recurrent units of the standard RNN architecture with more advanced "memory cells" (further referred to as *LSTM cells*) which can preserve gradients over time (Goldberg 2016, p. 399).

An LSTM cell has a *cell state* $c^{(t)}$ which is influenced by the current input $x^{(t)}$ at time $t$, the previous output $h^{(t-1)}$, and the previous cell state $c^{(t-1)}$.

Hence, the cell state adds another means of recurrence to the network in addition to the recurrence which is inherent to RNNs (Goodfellow, Bengio, and Courville 2016, p. 410).

Moreover, three gates control the flow of information[2] in LSTM cells: the *input gate* controls the influence of $x^{(t)}$ and $h^{(t-1)}$ on $c^{(t)}$, the *output gate* determines the degree to which $c^{(t)}$ affects $h^{(t)}$, and the *forget gate*[3] allows to ignore $c^{(t-1)}$ or parts of it, thereby controlling the influence of the past state on the current state (Lipton, Berkowitz, and Elkan 2015; Goodfellow, Bengio, and Courville 2016).

---

[2] The equations which are used to model the LSTM cell and the gating mechanism are thoroughly explained in the literature, e.g. Goodfellow, Bengio, and Courville (2016, pp. 410-411) or Greff et al. (2016). Since they are not relevant for understanding this thesis, they have been omitted.
[3] The forget gate was not included in the original design by Hochreiter and Schmidhuber (1997). It has been added later by Gers, Schmidhuber, and Cummins (2000) (Goldberg 2016).

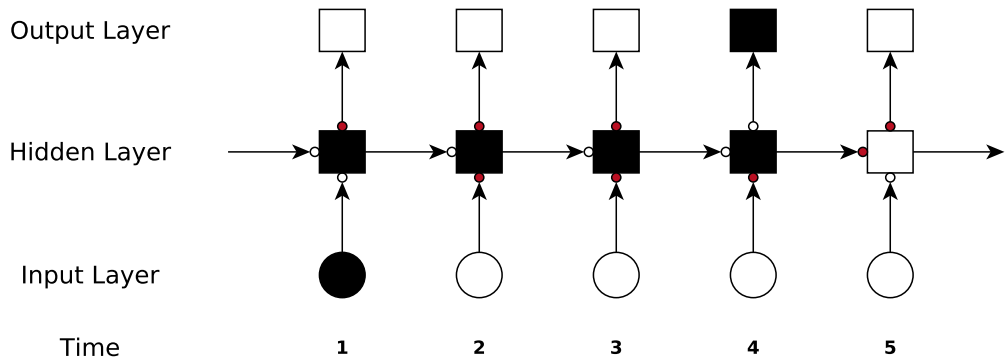

**Figure 3.3.:** As in figure 3.2, the shading of the network's nodes indicates their sensitivity to the first input. In difference to the previous visualization, the gradient information is preserved due to using the LSTM architecture. Each hidden layer node has an input (below), output (above), and forget (left) gate represented by a small circle. For simplicity, the gates are either completely open (not filled) or closed (filled with red). As long as the forget gate is open and the input gate is closed, the first input is remembered by the LSTM cell. This visualization is adapted from Graves (2012).

Figure 3.3 visualizes the intuition behind LSTM networks. By using the gating mechanism and the additional cell state, LSTM cells can store and retrieve information over long periods of time. This allows to train an RNN for a task with long-range dependencies more effectively because the vanishing gradient effect is mitigated (Graves 2012).

Since the initial proposal of the LSTM architecture, various extensions have been added and successfully improved the performance for solving NLP tasks with LSTM networks. Greff et al. (2016) provide an overview of multiple LSTM architectures and the history of the LSTM architecture.

**Gated Recurrent Unit**
While the use of LSTM cells is effective, they are also highly complex (Goldberg 2016). Therefore, Cho et al. (2014) proposed a novel hidden unit for RNNs which uses gates as well, but is less complex and therefore less expensive to compute. These hidden units are called Gated Recurrent Unit (GRU).

A GRU has no cell state $c^{(t)}$ and uses only two gates. The *update gate* controls the influence of the current input on the unit's activation and the *reset gate* allows to ignore the activation at the previous time step when calculating the current activation (Cho et al. 2014; Goodfellow, Bengio, and Courville 2016; Goldberg 2016). In contrast to an LSTM cell, the GRU cannot limit the degree to which the activation is exposed to other hidden units because it does not have an output gate (Chung et al. 2014).

For implementation details see the original proposal by Cho et al. (2014). While empirical evaluations have been performed (e.g. Chung et al. 2014), no conclusive results regarding which type of gated architecture is superior have been reported (Chung et al. 2014; Jozefowicz, Zaremba, and Sutskever 2015; Goodfellow, Bengio, and Courville 2016; Goldberg 2016).

## 3.2 Conditional Random Field

CRFs have been proposed by Lafferty, McCallum, and Pereira (2001) as a framework for modeling sequences with interdependent labels without suffering from the "label bias problem", i.e. the tendency of previous probabilistic and non-probabilistic sequence tagging models such as Maximum-Entropy Markov Models (MEMMs) to be biased towards states with fewer outgoing transitions.

The cause for this bias is that for a given state, transitions compete only against each other. For example, states with only one outgoing transition will always choose this transition independent of the observation (Lafferty, McCallum, and Pereira 2001).

CRFs are discriminative, undirected Markov models representing the conditional probability distribution of labels in a label sequence given a sequence of observations (Cuong et al. 2014). Instead of applying per-state normalization, i.e. the sum of the transition probabilities for any given state has to be 1, a CRF applies a per-sequence normalization, thus being conditioned on the whole sequence of observations (Do and Artieres 2010; Sutton and Mccallum 2012). Thereby, the label bias problem is solved (Lafferty, McCallum, and Pereira 2001).

While CRF classifiers can be used with handcrafted features such as capitalization (e.g. McCallum and W. Li 2003), they are often used in combination with neural networks which automatically extract suitable features (cf. section 2.1.2).

As briefly discussed in section 2.1.1, inference for CRFs is computationally expensive. In fact, it is NP-hard for non-linear, high-order CRFs (Sutton and Mccallum 2012). Hence, it is common to use only first-order CRFs. Higher-order CRFs usually require optimizations in order to make training and decoding feasible as shown by Müller, Schmid, and Schütze (2013) and Cuong et al. (2014).

## 3.3 Multi-Task Learning

Multi-Task Learning (MTL) has already been introduced in section 2.2. This section presents the intuition of MTL and provides more details on some of the MTL-related techniques which have been mentioned briefly in chapter 2. However, the scope of this section is limited to what is relevant for the implementation (cf. chapter 4) and potential future work (cf. chapter 8).

### 3.3.1 Intuition

In 1993, Caruana first formulated the intuition and motivation behind MTL. Traditional machine learning methodology splits complex problems into smaller, reasonably independent subproblems. For each subproblem, a separate model is learned in isolation. Afterwards, the models are recombined in order to solve the initial task.

Caruana (1993), however, claims that this methodology ignores the inherent similarity of related task which may serve as a crucial source of *inductive bias*. Inductive bias is a set of assumptions which allow a learner to generalize beyond the training samples by enabling it to make predictions that are not solely based on consistency with the training instances (T. M. Mitchell 1980).

Usually, inductive bias is created by codifying domain experts' knowledge into a learner. With MTL this domain-specific bias can be acquired by collecting additional training signals through auxiliary tasks (Caruana 1993). Caruana (1998) summarizes this as follows: "MTL improves generalization by leveraging the domain-specific information contained in the training signals of related tasks."

In addition to this motivation from the machine learning perspective, Caruana (1993) sees MTL as inspired by human learning. Humans are constantly tasked to learn various related things and are able to transfer knowledge from one task to a similar task. In other words, improvement in one task directly translates to improvements in several related tasks. This enables human beings to succeed even in complex task with only little prior experience.

Ruder (2017) provides a highly intuitive example for this biological motivation of MTL:

> *In the movie [The Karate Kid (1984)], sensei Mr Miyagi teaches the karate kid seemingly unrelated tasks such as sanding the floor and waxing a car. In hindsight, these, however, turn out to equip him with invaluable skills that are relevant for learning karate.*

In conclusion, MTL is a form of inductive transfer with the goal of leveraging additional sources of information for improvements when learning the current task. These sources include, but are not limited to, background domain knowledge, training signals for related tasks, and models for the same task learned with other learners or from different distributions (Caruana 1996).

### 3.3.2 Benefits

Section 2.2.1 already presented different use cases in which MTL was effective. Therefore, this section does not discuss specific use cases, but rather overarching benefits of using MTL reported by various authors and theoretical benefits claimed by Caruana (1993), for instance.

**Fighting Data Sparseness**

The most frequently mentioned benefit of MTL is being able to mitigate the issue of data sparseness (Ramsundar et al. 2015; Augenstein and Søgaard 2017; Benton, M. Mitchell, and D. Hovy 2017).

One option to achieve this is leveraging different views on the existing data, e.g. via natural subtasks (cf. section 2.2.5) or using different metrics or encodings for the same task (Caruana 1996).

The other option is using additional data for auxiliary tasks. The additional data may be obtained from common NLP tasks such as POS tagging which do not suffer from the data sparsity issue (Marcus, Santorini, and Marcinkiewicz 1993; Hashimoto et al. 2017), other domains (N. Peng and Dredze 2016), and other languages (Yang, Salakhutdinov, and Cohen 2016; Braud, Lacroix, and Søgaard 2017).

**Regularizing Effect**

Furthermore, MTL acts as a powerful regularizer, thereby preventing overfitting (Ruder 2017). This regularizing effect can be observed because a learner that is trained with MTL is biased towards hypotheses which are useful for multiple tasks instead of a single task as in STL (Caruana 1993; Bollmann and Søgaard 2016).

**Focus Attention**

MTL can also enable a learner to focus its attention on patterns in the input which would be ignored otherwise. This is done with the help of an auxiliary task (Caruana 1996).

For instance, an automated steering system may ignore lane markings as they are not visible at all times and only make up a fraction of the whole road image when they are present. However, adding the task of detecting lane markings as an auxiliary task, pressures the learner to include lane markings in its learned representation, thus making this previously ignored feature available to the main task, i.e. steering (Caruana 1998).

**Knowledge-Based Inductive Bias**

Finally, the previous example also demonstrates the integration of domain expert knowledge, i.e. the knowledge that lane markings are helpful for making steering decisions, in the learner without having to change the learner's algorithm. Caruana (1993) refers to this benefit of MTL as "knowledge-based inductive bias".

### 3.3.3 Parameter Sharing

While section 2.2.3 discussed advanced parameter sharing techniques, this section explains the traditional parameter sharing approach in MTL (Caruana 1993) which is still prevalent (Ruder 2017) and used in this thesis. Ruder (2017) refers to this approach as "hard parameter sharing" because the layers of the network are shared by all tasks to full extent.

As shown in figure 3.4a, the task-specific classifiers feed from the same shared layers. Therefore, the error for each task is backpropagated through the shared layers and thus the weights are adapted through the loss of all tasks. This ensures that the shared layers capture a representation that suits all tasks. Moreover, a knowledge transfer between the tasks is possible since "what is learned for each task can benefit others" (Caruana 1996).

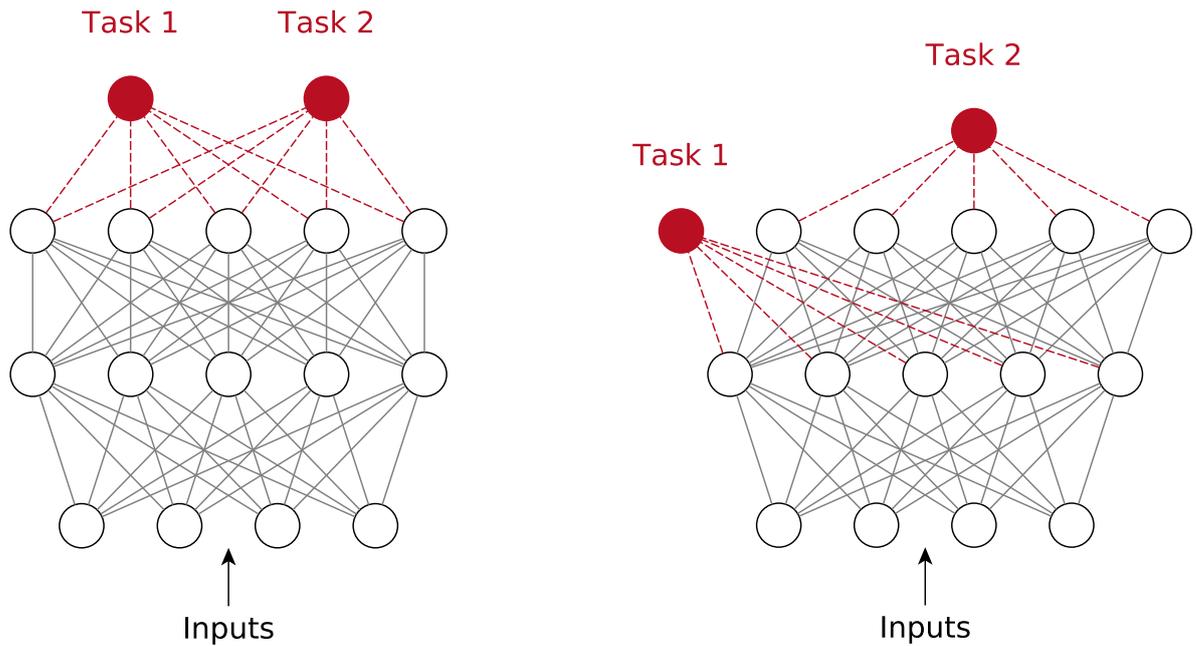

**(a)** Visualization of hard parameter sharing adapted from Caruana (1993). The classifiers of *task 1* and *2* feed from the same shared layers in the neural network.

**(b)** As in Søgaard and Goldberg (2016), the tasks feed from different levels of the shared layers. *Task 1* is considered a lower-level task and thus is fed from a lower level than *task 2*.

**Figure 3.4.:** Parameter sharing in a hierarchical and non-hierarchical MTL architecture. The task-specific classifiers are visualized as red circles. Every time the loss for a prediction is calculated at either classifier, this error is backpropagated through all previous layers.

### 3.3.4 Network Architecture

Figure 3.4a, shows an architecture which feeds all task classifiers from the same shared layer. Thereby, all shared layers are influenced by the loss of all tasks.

Another approach has been suggested by Søgaard and Goldberg (2016) (cf. section 2.2.2). Given some hierarchy between the main and auxiliary tasks, it may be beneficial to feed different task classifiers from different layers of the shared space as visualized in figure 3.4b.

In the remainder of this thesis, *terminating a task at layer X* refers to feeding the task's classifier from the shared layer $X$.

This architecture allows the learner to learn a more suitable representation for the higher-level and often more complex task because the higher layer is only influenced by the loss of the higher-level task. The low-level task, however, is still able to influence the classification for the high-level task through the low-level shared layers.

Similarly, task-specific hidden layers can be used after the final shared layer as suggested by H. Peng, Thomson, and Smith (2017). This enables the learner to build models with higher complexity for difficult tasks, too. Furthermore, both approaches are similar to the idea of separating shared and private feature spaces which has been discussed in section 2.2.3.

Two other architecture variants, namely *shortcut connections* and *label embeddings*, by Hashimoto et al. (2017) have been introduced briefly in section 2.2.2. In the following, they will be explained more thoroughly. A visual comparison of these features is presented in figure 3.5.

Usually, the word embeddings which are created from the word indices in the input via an embeddings layer are only used as input for the first shared layer. *Shortcut connections* (see figure 3.5b), however, allow to use them as the input for the higher levels of the shared layers as well by concatenating the word vectors with the output of the previous shared layer.

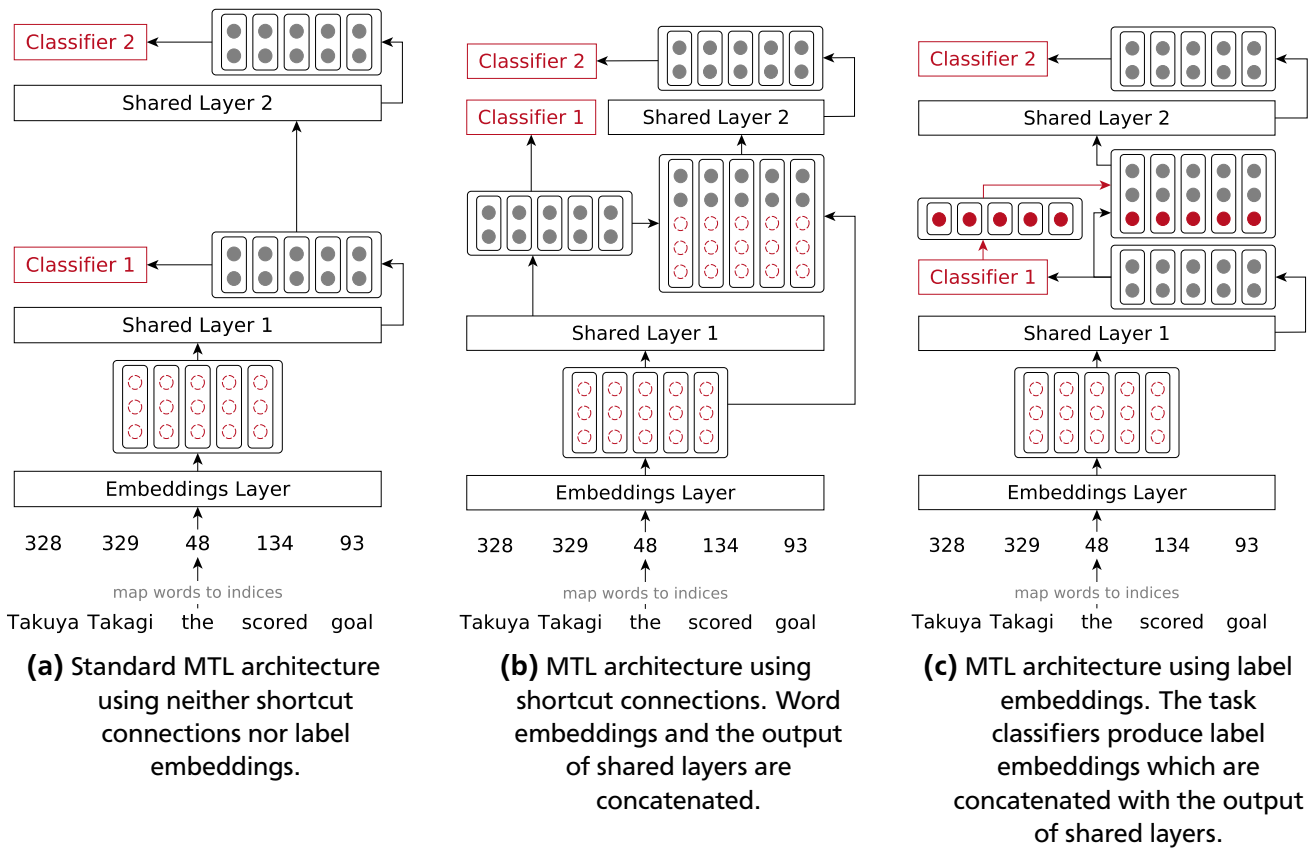

Figure 3.5.: Comparison between different MTL networks using the advanced architectural features introduced by Hashimoto et al. (2017). Word embeddings are represented by vectors of dashed circles, label embeddings by vectors of red circles, and the output of shared layers as vectors of gray circles. *Classifier 1*, *2*, and *3* are classifiers for different tasks which have a natural hierarchy.

*Label embeddings* (see figure 3.5c) enable higher-level shared layers to utilize the classification results of lower-level tasks. This is done by mapping the predicted label indices to dense vector representations, i.e. *embeddings*, which are concatenated with the other inputs of the higher-level shared layers.

### 3.3.5  Suitable Auxiliary Tasks

Whether or not MTL is successful, largely depends on finding suitable auxiliary tasks (cf. section 2.2.4). However, determining which properties of a task are responsible for making it a *suitable auxiliary task* remains an open research question (Ruder 2017).

Despite the importance of finding good auxiliary tasks, little research has been conducted in this area (Alonso and Plank 2017; Bingel and Søgaard 2017). Usually, authors rely on the intuition that main and auxiliary tasks should be related because MTL is a method of inductive transfer which does not seem to be reasonable for unrelated tasks (Caruana 1998).

In most cases, authors use domain-specific knowledge for justifying their choice of auxiliary tasks (e.g. Søgaard and Goldberg 2016; Hashimoto et al. 2017). A systematic notion of relatedness between tasks, however, remains to be discovered, too (Ruder 2017).

According to Caruana (1998), an auxiliary task is related to a main task if it improves the learners performance on the main task. It is important to note that the learner does not need to be good at solving the auxiliary task because MTL strives to improve the performance on the main task, "not to reduce the number of models" (Caruana 1996).

Caruana (1998) further claims that related tasks must share input features, but also finds that a main and an auxiliary task which are related when using one learning procedure may be unrelated with another learning procedure and that it is difficult, but not impossible to determine task relatedness before the actual training process.

Since this training process is expensive, it is preferable to assess the suitability of auxiliary tasks before the training. Hence, recent work (cf. section 2.2.4) attempts to predict which tasks are suitable by means of dataset properties (Alonso and Plank 2017) and learning curves of the respective auxiliary task when trained in isolation (Bingel and Søgaard 2017).

Alonso and Plank (2017) use the information-theoretic metrics *entropy* and *kurtosis* to quantify the properties of the label distributions in different datasets:

*Entropy* – Entropy indicates the amount of uncertainty in a distribution $p(x)$. It is low for imbalanced distributions and maximized if all values $x$ in the distribution are equally likely. It is calculated as follows:

$$-\sum_{x} p(x) \log_2 p(x) \qquad (3.2)$$

(Bishop 2006, pp. 48-51)

*Kurtosis* – Kurtosis $g_2$ is a measure to describe the shape of a distribution. The kurtosis of a normal distribution is $g_2 = 3$. On the one hand, distributions with flat-topped frequency curves have $g_2 < 3$ and are "said to have kurtosis". On the other hand, $g_2 > 3$ indicates a narrow and peaked frequency curve. The distribution "is said to lack kurtosis" (Kenney 1939, pp. 104-106).

Kurtosis is a measure of tail weight as well as peakedness (Ruppert 1987). Alonso and Plank (2017) use it to reason about the tailedness of label distributions in different datasets. A distribution which *lacks kurtosis* is very tailed while a distribution that *has kurtosis* is more uniform.

Kurtosis (equation (3.3)) is based on the second (equation (3.4)) and fourth (equation (3.5)) moment about the mean $\bar{x}$. The notation has been obtained from Zwillinger and Kokoska (1999). The second moment is usually referred to as *variance*.

$$g_2 = \frac{m_4}{m_2^2} \qquad (3.3) \qquad m_2 = \frac{1}{n}\sum_{i=1}^{n}(x_i - \bar{x})^2 \qquad (3.4) \qquad m_4 = \frac{1}{n}\sum_{i=1}^{n}(x_i - \bar{x})^4 \qquad (3.5)$$

Alonso and Plank (2017) conclude that auxiliary tasks with datasets that have kurtosis, i.e. $g_2 < 3$, and fairly high entropy are superior to tasks that do not have such characteristics. Moreover, tasks with a smaller label space are preferable as auxiliary tasks.

Bingel and Søgaard (2017) support the conclusions made by Alonso and Plank (2017). In addition to that, they find that the learning curves of auxiliary tasks in a STL setting are good indicators for MTL gains.

They observe that auxiliary tasks can help the main task to escape a local minimum when its learning curve is flat while their learning curves remain steep. Their summarization of this observation is:

> *Multi-task gains are more likely for [main] tasks that quickly plateau with non-plateauing auxiliary tasks.*

As introduced in section 2.2.5, natural subtasks empirically proved to be useful auxiliary tasks. Given the previously explained theoretical findings, this success can be tied to the properties of these subtasks: by definition their label inventory is smaller (cf. Alonso and Plank (2017)) and they operate on the same input features as the main task (cf. Caruana (1998)).

### 3.3.6 Criticism

Despite various promising results, MTL does not always yield performance gains (cf. section 2.2.1). Reimers and Gurevych (2017a) point out that MTL research frequently lacks a comparison with a pipeline approach in which the labels of the auxiliary tasks are used as features for the main task. They hypothesize that this pipeline approach might outperform the MTL setup.

The challenge of finding good auxiliary tasks has been discussed in the previous section. This challenge is still frequently solved with empirical evidence, domain-specific knowledge or intuition (cf. section 3.3.5). Therefore, the application of MTL can be seen as a type of manual feature engineering. Most neural network approaches, however, aim to reduce the amount of feature engineering (e.g. Lample et al. 2016; Ma and E. Hovy 2016; Yang, Salakhutdinov, and Cohen 2016).

Finally, MTL training is more complex for the learner and thus is very likely to increase the training time, i.e. make training more expensive (Caruana 1993).

## 3.4 Label Scheme for Sequence Tagging

For sequence tagging tasks in NLP (cf. section 2.1) various tagging schemes are used to encode sequences which span over multiple tokens in the label set. An overview of the different tagging schemes is provided by Collobert et al. (2011, p. 2505).

Most authors define the schemes themselves without any references (e.g. Collobert et al. 2011; Lample et al. 2016; M. Li et al. 2017) or do not provide an explanation or a reference at all (e.g. Z. Huang, Wei Xu, and K. Yu 2015; Hashimoto et al. 2017). While this can be considered a minor issue due to the familiarity of the NLP community with these schemes, inconsistencies between the definitions are a far more serious issue which may prevent reproducibility of experiments.

This issue is primarily caused by authors referring to IOB2 when using the abbreviation IOB (Nugues 2014, p. 299). Hence, two publications may not refer to the same tagging scheme although both use the abbreviation IOB. For example, this can be seen when comparing Collobert et al. (2011) and Reimers and Gurevych (2017a). Reimers and Gurevych (2017a) use the abbreviation IOB correctly and to refer to IOB2 with "BIO".

To prevent non-reproducibility, this thesis explicitly defines the used tagging scheme. The BIO scheme as used by Eger, Daxenberger, and Gurevych (2017) and Reimers and Gurevych (2017a), for instance, will be used in the following.

> *The BIO scheme marks the beginning of a segment with a `B-` tag and all other tokens of the same span with a `I-` tag. The `O` tag is used to tokens that are outside of a segment. (Reimers and Gurevych, 2017)*

## 3.5 Argumentation Mining

AM has already been introduced in section 2.3. This research field is concerned with the automated identification of argumentative structures within natural language documents (Green et al. 2014).

Common tasks are segmenting a text into argumentative and non-argumentative components, identifying the type, e.g. premise or claim, of a component, and analyzing the relationships between argumentation components (Persing and Ng 2016; Stab and Gurevych 2017).

Analyzing argumentative structures can be beneficial in many applications such as visualizing and summarizing arguments in texts to improve comprehensibility for end-users, information retrieval and extraction, and text summarization in general (Green et al. 2014; Persing and Ng 2016; Habernal and Gurevych 2017).

For a more thorough introduction to AM, its theoretical background, and different argumentation models and theories see the journal articles by Habernal and Gurevych (2017) and Stab and Gurevych (2017).

There is no consensus on which argumentation theory or model is best (Habernal and Gurevych 2017). In this thesis, the model used by Stab and Gurevych (2017) for their *Persuasive Essays* dataset is used. An argumentation structure is modeled as a tree with a *major claim* as its root. This shows the author's stance on the topic of interest. The arguments in the text can support or attack the major claim. An argument consists of a claim and several premises. The claim has a stance attribute which indicates whether it supports ("for") or attacks ("against") the major claim. Premises are reasons for a claim or another premise, i.e. premise chains are possible. A premise has exactly one outgoing relation to a claim or premise which is either a *support* or *attack* relation and may have multiple incoming relations.

A simple example is provided by Eger, Daxenberger, and Gurevych (2017):

> *Since it killed many marine lives*$_{Premise}$, *tourism has threatened nature*$_{Claim}$.

In this example, the premise $P$ supports the claim $C$. A complete example with all types of components and relations is given by Stab and Gurevych (2017). An excerpt of this example is shown in appendix A.

### 3.5.1 Labels

Following Eger, Daxenberger, and Gurevych (2017), AM is framed as a sequence tagging problem in this thesis. Their label scheme for AM will be used as well.

Each label is a four-tuple $(b, t, d, s)$ where $b$ determines whether or not a token belongs to an argumentative component using the BIO scheme and $t$ classifies the component as major claim MC, claim C or premise P.

The outgoing relation of a premise is represented by $d$ and $s$. While $d$ encodes the relative distance to the related component, $s$ refers to the type of relation (stance), i.e. support (Supp) or attack (Att). The distance is measured in number of components.

Since claims have a relation with the major claim, they also have a stance value $s$. For them, it is either for (For) or against (Ag) depending on whether the claim supports or attacks the major claim.

A special symbol $\perp$ is used when a component of the four-tuple cannot be filled for a specific token, e.g. a claim cannot have a distance as it has no outgoing relation except for the implicit relation to the text's major claim.

This set of labels $Y$ is summarized in equation (3.6).

$$\begin{aligned} Y = \{(b, t, d, s) \,|\, &b \in \{B, I, O\}, \\ &t \in \{P, C, MC, \perp\}, \\ &d \in \{\ldots, -2, -1, 1, 2, \ldots, \perp\} \\ &s \in \{Supp, Att, For, Ag, \perp\}\} \end{aligned} \quad (3.6)$$

Assuming that the claim supports the text's major claim, e.g. "tourism is not sustainable", the previous example is labeled as follows when framing the task of AM as sequence tagging:

| O | B:P:1:Supp | I:P:1:Supp | I:P:1:Supp | I:P:1:Supp | I:P:1:Supp |
|---|---|---|---|---|---|
| Since | it | killed | many | marine | lives |

| B:C:$\perp$:For | I:C:$\perp$:For | I:C:$\perp$:For | I:C:$\perp$:For | O |
|---|---|---|---|---|
| tourism | has | threatened | nature | . |

This label scheme is also very suitable for deriving natural subtasks (cf. section 2.2.5) as the four-tuple can be converted into a smaller tuple for a specific subtask.

### 3.5.2 Metrics

To evaluate the AM predictions, the metric introduced by Persing and Ng (2016) is used. It allows to calculate component and relation F1 scores by computing true positives TP, false positives FP, and false negatives FN on the component and relation level.

Moreover, they define the notion of an *exact match* and an *approximate match*. A predicted component/relation exactly matches a true component/relation if all their constituents, i.e. tokens, match. If at least half of the tokens are shared between gold standard and prediction, an approximate match exists.

Equations 3.7 – 3.9 describe the computation of TP, FP, and FN while equation (3.10) shows the calculation of the F1 score. Persing and Ng (2016) have defined all of these equations.

$$\text{TP} = |\{j \mid \exists i \; gl(i) = pl(i) \land i \doteq j\}| \quad (3.7)$$

$$\text{FP} = |\{i \mid pl(i) \neq n \land \nexists j \; gl(i) = pl(i) \land i \doteq j\}| \quad (3.8)$$

$$\text{FN} = |\{j \mid \nexists i \; gl(i) = pl(i) \land i \doteq j\}| \quad (3.9)$$

$$\text{F1} = \frac{2\text{TP}}{2\text{TP} + \text{FP} + \text{FN}} \quad (3.10)$$

To calculate the component F1 score, the four-tuple label (cf. section 3.5.1) is reduced to $b$ and $t$ in order to describe only argumentation components without their relations. While $b$ is used to describe the spans of components, i.e. which tokens constitute a component, $t$ is the actual label of a component.

The gold standard components are $j$ and the predicted components are $i$. The gold and predicted label are retrieved by $gl(j)$ and $pl(i)$ respectively. $n$ is the non-argumentative class, i.e. $b = O$, and $i \doteq j$ denotes the matching relation. That is, $i$ is an exact or approximate match for $j$.

According to equation (3.7), true positives are all true components $j$ for which a matching predicted component $i$ of the same type $t$ exists.

The relation F1 score is calculated analogously, but the full four-tuple is taken into account. Hence, the relation F1 score depends on the component F1 score because correct relations must have correct arguments, i.e. components (Eger, Daxenberger, and Gurevych 2017). In case of the relation F1 score, $n$ denotes the no-relation class and $j$ and $i$ represent true and predicted relations respectively.

In the following, the different scores are referred to as C-F1 (50%), C-F1 (100%), R-F1 (50%), and R-F1 (100%) for the component (C-F1) and relation (R-F1) scores with exact (100%) and approximate (50%) matches.

### 3.6 Sequence-to-Sequence (S2S) Learning

Sequence-to-Sequence (S2S) learning is tasked with mapping an input sequence to an output sequence. In difference to sequence tagging (cf. section 2.1), however, both sequences can be of variable length. In particular, input and output do not necessarily have to be of the same length (Luong et al. 2016; Kreutzer, Sokolov, and Riezler 2017).

This property of S2S learning is a problem for deep neural network approaches because they operate on vectors of fixed dimensionality (Sutskever, Vinyals, and Le 2014). To solve this issue, *encoder-decoder* architectures which consist of two RNNs have been introduced (Cho et al. 2014). The first RNN maps the variable length input sequence to a fixed length vector representation. The second one maps this representation into another variable length sequence (Gehring et al. 2017).

Using encoder-decoder architectures, S2S learning can be used to solve various NLP tasks such as machine translation (Cho et al. 2014; Sutskever, Vinyals, and Le 2014) and multimodal translation (Libovický and Helcl 2017).

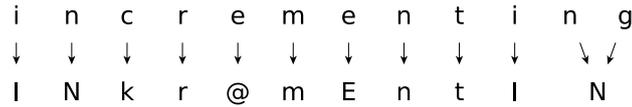

**Figure 3.6.:** Alignment of "incrementing" in the Combilex pronunciation lexicon (Richmond, R. A. J. Clark, and Fitt 2009) produced with the aligner by Jiampojamarn, Kondrak, and Sherif (2007).

### 3.6.1 Grapheme-to-Phoneme Conversion

Grapheme-to-Phoneme (G2P) conversion is a common S2S problem. The task is the conversion of a grapheme (letter) string into a phoneme string (Eger 2015a). While graphemes represent words in written language, phonemes represent them in spoken language. G2P conversion is, for instance, important for speech synthesis systems because it is a necessary intermediate step for producing sounds (Jiampojamarn, Kondrak, and Sherif 2007).

### 3.6.2 S2S as a Sequence Tagging Problem

As previously mentioned, the input and output sequences of S2S problems do not have to be of the same length which prevents the application of traditional sequence tagging architectures such as LSTMs.

Nevertheless, S2S problems can be framed as sequence tagging problems by *aligning* the items in the input sequence with the items in the output sequence (Eger 2015b). Note that the alignment may only be performed on the training data (Eger 2015a).

As shown in the example in figure 3.6, a simple one-to-one alignment is not always possible because a grapheme may produce two phonemes or two graphemes may be combined into a single phoneme. Thus, an aligner which is capable of producing many-to-many alignments is necessary (Jiampojamarn, Kondrak, and Sherif 2007).

However, not all S2S problems can be aligned easily. In particular, the task has to be monotone (Eger 2015b). For a monotone S2S task matching relations of subsequences between the input and output sequence obey monotonicity. In other words, there are no "crossing edges" (see figure 3.6) in the alignments (Schnober et al. 2016).

After the alignment, the input and output sequences of the S2S problem have the same length which allows the application of sequence tagging architectures by using the input sequence as observed variables and the output sequence as labels (Jiampojamarn, Kondrak, and Sherif 2007; Eger 2015b).

### 3.6.3 Metrics

The only S2S problem which is covered in this thesis is G2P. Therefore, only G2P metrics are used. In particular, the metrics *word accuracy* and *edit distance* will be used in this thesis.

*Word accuracy* – Word Accuracy (WACC) is calculated as the percentage of correctly converted grapheme sequences among all converted sequences in the test dataset (Eger 2015a).

*Edit distance* – The Edit Distance (ED) or Levenshtein distance (Levenshtein 1965) measures the similarity of strings (Apostolico and Galil 1997). Given two strings, it is calculated as "the minimal number of insertions, deletions and substitutions to make two strings equal". Each of these operations has a cost of one (Navarro 2001).

Since the edit distance is calculated on a per-word basis, the *mean edit distance* is used as a measure for the performance on a complete dataset.

# 4 Multi-Task Learning Sequence Tagging Framework

## 4.1 Requirements

The development of an *MTL sequence tagging framework* is one major contribution of this thesis. This system has to be *general* and *expressive* so that it can be used for various sequence tagging tasks. It is based on existing work and associated code which has been presented in chapter 2.

In particular, it combines existing sequence tagging approaches (section 2.1) with MTL (section 2.2). In this regard, the system is similar to the work of Reimers and Gurevych (2017a), but it also adds new building blocks such as shortcut connections (cf. section 4.3.2) and focuses on configurability and extensibility.

## 4.2 Technology

The system is based on TensorFlow[4] which "is an interface for expressing machine learning algorithms, and an implementation for executing such algorithms" (Abadi et al. 2015).

TensorFlow's programming model is based on a computational graph consisting of nodes that represent operations. From the graph's input nodes to its output nodes tensors flow along the edges and are modified by the graph's operations. A tensor is an array of arbitrary dimensionality (Abadi et al. 2015).

TensorFlow has been used to solve several machine learning problems such as object recognition (Frome, Gs Corrado, and Shlens 2013), machine translation (Sutskever, Vinyals, and Le 2014), and playing Go (Maddison et al. 2015).

The MTL sequence tagging framework uses TensorFlow's Python Application Programming Interface (API) with Python 2.7.x.

## 4.3 Features

This section presents all implemented features of the MTL sequence tagging framework. Due to time constraints not all potentially useful features have been implemented at the time of writing. Features which we have left for future work will be discussed in chapter 8.

### 4.3.1 Configuration

To ensure that the system is *general* and *expressive* as required, comprehensive configuration options have been implemented to modify the behavior of the underlying network.

Configuration options for the MTL sequence tagging framework are provided in a configuration file in YAML format. The format was chosen because it is easily readable by humans, agnostic of the used programming language, and supports comments (Ben-Kiki, Evans, and Ingerson 2009, p. 1).

The configuration file is used to design 1) the training process, 2) tasks to learn, 3) which input files to use, 4) the use of pre-trained word embeddings, 5) the network architecture including shared and private layers, 6) the application of regularization techniques, 7) and the evaluation process including post-processing and metrics.

---

[4] `https://www.tensorflow.org/`

For example, the user can choose between several optimizers[5] including Adam (Kingma and Ba 2015), Adadelta (Zeiler 2012), and Adagrad (Duchi, Hazan, and Singer 2011) for the training process.

A full explanation of all configuration options can be found alongside the code[6]. In short, the configuration allows a user to modify almost any aspect of the network and thus provides high flexibility. Due to this flexibility, all types of sequence tagging problems can be solved. This makes the MTL sequence tagging framework extremely expressive.

### 4.3.2 Network Architecture

As mentioned in section 4.1, the MTL sequence tagging framework combines existing approaches. Hence, all papers referenced in this section have already been presented in chapter 2.

**Recurrent Neural Network (RNN)**

The preference for utilizing RNNs to solve sequence tagging tasks in the NLP community has been shown in section 2.1.2. Moreover, their suitability for these tasks has been justified in section 3.1. The system, therefore, uses bidirectional RNN layers with either simple RNN cells, LSTM cells or GRU cells depending on the configuration chosen by the user.

**Parameter Sharing**

The MTL sequence tagging framework uses hard parameter sharing (cf. section 3.3.3) as introduced by Caruana (1993) and used by most existing MTL architectures (Ruder 2017). The recurrent layers are thus shared by all tasks.

**Task Termination at Different Layers**

Motivated by Søgaard and Goldberg (2016) and Hashimoto et al. (2017), however, the system allows to terminate tasks at different levels of the shared layers. That is, different tasks can be supervised on different layers depending on the chosen configuration. The shared layers, thereby, are not necessarily completely shared by all tasks. The number of shared layers depends on the highest level of supervision among all tasks and the level of supervision can be specified individually for each task. In figure 4.1, for example, one task is terminated at the first level and all other tasks are terminated at the second level.

**Private Layers**

In addition, each task may have an arbitrary number of private layers between its termination layer and its classifier. These additional layers shall provide more capacity for task-specific features. This architectural decision was inspired by H. Peng, Thomson, and Smith (2017). The NER task in figure 4.1 uses a single private layer, for instance.

In difference to the shared layers which consist of RNN cells, the private layers contain simple neural cells which map a vector $x \in \mathbb{R}^m$ to a vector $y \in \mathbb{R}^n$ by multiplying $x^T$ with a weight matrix $W \in \mathbb{R}^{m \times n}$ and applying an activation function $\sigma$, i.e. $y^T = \sigma(x^T W)$. The activation function $\sigma$, e.g. tanh, and number of output units of the cell $n$ can be configured by the user. The weights are updated using backpropagation.

Independent of the private layer setting, each task has a *projection layer* that maps the potentially high-dimensional input from the shared layers or the private layers to a vector whose dimensionality is equivalent to the number of available labels/classes. These projection layers are omitted for simplicity in figure 4.1.

---

[5] https://www.tensorflow.org/versions/r1.0/api_guides/python/train
[6] https://git.ukp.informatik.tu-darmstadt.de/tk-master-thesis/mt-code/blob/master/experiments/12_tensorflow_port/

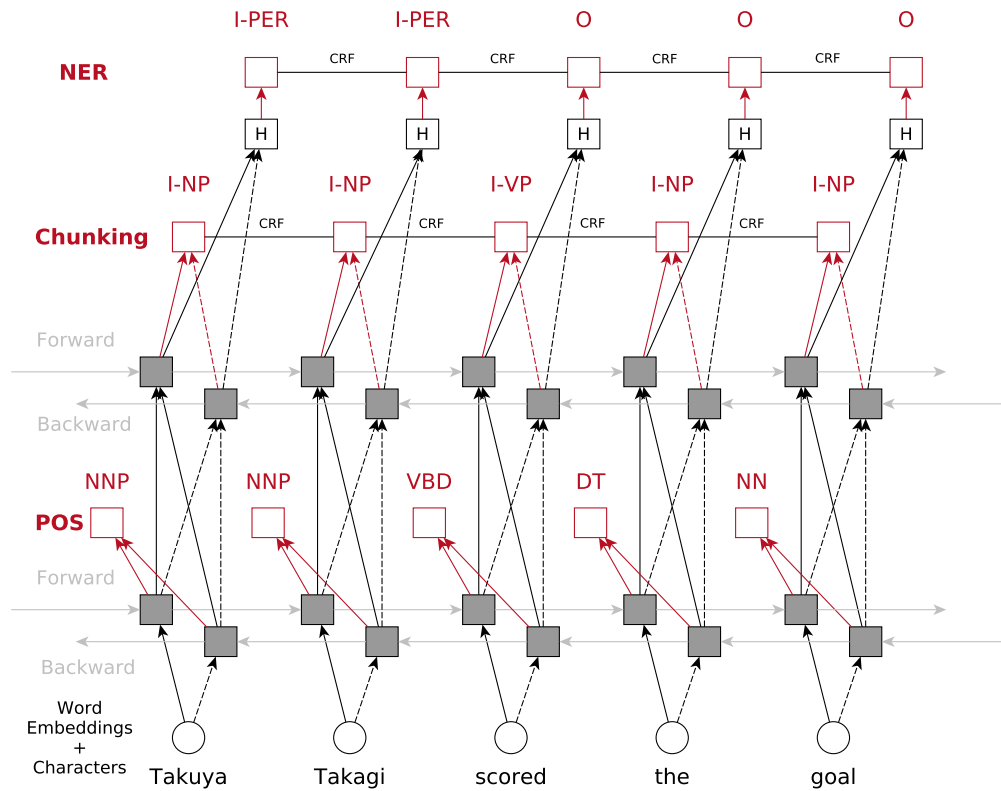

**Figure 4.1.:** This is an example for a network architecture with three tasks: POS tagging, chunking, and NER. It has two shared layers (gray rectangles) and uses character-level information. Shortcut connections have been disabled and dropout has been excluded to reduce clutter in the visualization. POS tagging is terminated at the first shared layer and chunking and NER are terminated at the second. The latter use a CRF classifier to model their label dependencies and the NER task additionally uses a private layer.

**Modeling Label Dependencies**

To enable modeling label dependencies for tasks which have interdependent labels, the user can configure for each task whether or not to use a CRF classifier. It is evident from the analysis of the related work in section 2.1 and the theoretical background provided in section 3.2 that CRFs facilitate sequence tagging when label dependencies exist. The CRF classifier, however, only models first-order dependencies. If a task is not using the CRF classifier, it uses a softmax classifier.

**Shortcut Connections**

Using "shortcut connections" had an substantial influence on the performance of the system by Hashimoto et al. (2017). Hence, the MTL sequence tagging framework also feeds the word representations into each shared layer instead of feeding them just into the first one. This behavior, however, can be disabled in the configuration.

**Character-Level Information**

Similarly, using character-level information can also be disabled because Alonso and Plank (2017) empirically show that character-level information may hurt the performance in some cases. If it is enabled, the system uses a bidirectional LSTM to extract character-level information in order to improve handling OOV words. This approach is based on Lample et al. (2016).

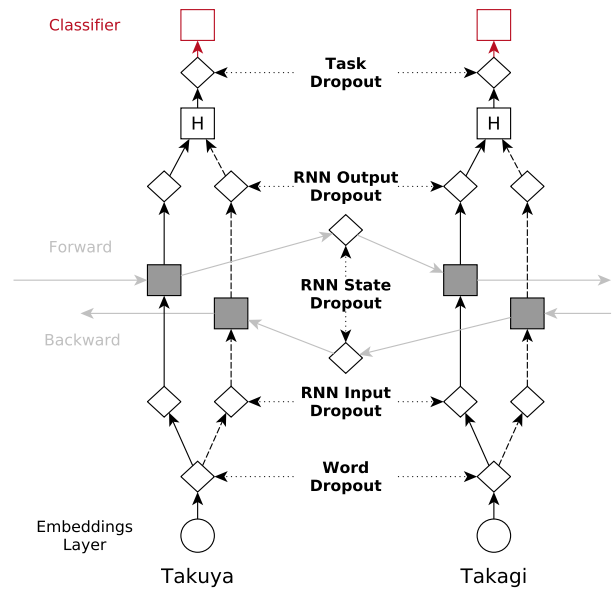

**Figure 4.2.:** All available dropout options are visualized as diamonds (rhombi) and labeled according to their purpose.

### 4.3.3 Dropout

While MTL has a regularizing effect itself, *dropout* (Srivastava et al. 2014) is supported as well. The configuration allows to specify five different dropout options:

*Word dropout* – This dropout option allows to drop word embeddings from the input sequence with the given dropout probability as described in Iyyer et al. (2015). This may yield better adaption to OOV words and prevent overfitting because the network cannot rely on the presence of specific input words (Goldberg 2017).

*RNN input dropout* – Dropout is applied to the input units of an RNN cell.

*RNN state dropout* – Dropout is applied to the recurrent connections between RNN cells of different time steps.

*RNN output dropout* – Dropout is applied to the output units of an RNN cell.

*Task dropout* – For each task, a dropout probability may be specified. This allows dropping units from the output of the projection layer with the given probability before feeding them into the classifier.

All of these options are visualized in figure 4.2. The RNN-related dropout option can also be applied using variational dropout (Gal and Ghahramani 2016). Instead of applying different dropout masks at each time step, the same mask is applied at each time step. Of course, the masks for input, state, and output dropout are still different. Using variational dropout can have superior performance as shown by Gal and Ghahramani (2016). Note that using RNN state dropout is *not recommended* without activating variational dropout because it typically results in performance deterioration (Gal and Ghahramani 2016).

### 4.3.4 Norm Clipping

To deal with exploding gradients (cf. section 3.1.2), *norm clipping* (Pascanu, Mikolov, and Bengio 2013) can be used in the MTL sequence tagging framework. The clipping or "normalization" (Reimers and Gurevych 2017a) is performed as proposed by Pascanu, Mikolov, and Bengio (2013) by using TensorFlow's `tf.clip_by_global_norm`[7].

Let $\hat{g}$ represent all gradients of the network and $||\hat{g}||$ be its global norm. The global norm[8] is the norm of all norms of the gradients in $\hat{g}$. The calculation is shown in equation (4.1).

$$||\hat{g}|| = \sqrt{\sum_{g \in \hat{g}} ||g||^2} \qquad (4.1)$$

For a given threshold which can be configured by the user, the clipping is performed as follows:

$$\hat{g} \leftarrow \frac{\text{threshold}}{\max(\text{threshold}, ||\hat{g}||)} \hat{g} \qquad (4.2)$$

As shown in equation (4.2), the gradients are only clipped if their global norm exceeds the configured threshold value.

### 4.3.5 Utilizing Pre-Trained Word Embeddings

Utilizing pre-trained word embeddings is a simple and effective method to improve the performance of machine learning systems in NLP (Turian et al. 2010). Therefore, the MTL sequence tagging framework also enables the user to work with pre-trained embeddings.

Furthermore, multiple pre-trained word embeddings can be used together by concatenating them. This concatenation is performed automatically. The user specifies which word embeddings to use and the system builds a set of concatenated word embeddings $\mathscr{E}$ as follows:

Let $E_i$ be the $i^{th}$ set of $n$ pre-trained embeddings and for a word $w$, $E_i^{(w)}$ yields the corresponding word vector from $E_i$. Further, let $W_{E_i}$ be the set of words for which $E_i$ provides word vectors.

$$\mathscr{W} = \bigcap_{1 \leq i \leq n} W_{E_i} \qquad (4.3) \qquad \mathscr{E} = \left\{ E_1^{(w)} \circ \cdots \circ E_n^{(w)} \mid w \in \mathscr{W} \right\} \qquad (4.4)$$

The set of words $\mathscr{W}$ supported by the concatenated word embeddings $\mathscr{E}$ is determined in equation (4.3). Based on that, $\mathscr{E}$ can be built as shown in equation (4.4).

Given the aforementioned building process for $\mathscr{E}$, it is evident that $\mathscr{E}$ might contain much less word vectors than the original embedding files as only the intersection of supported words is used, i.e. all words that occur in all embedding files.

Y. Zhang, Roller, and Wallace (2016) report consistent performance improvements when utilizing multiple pre-trained word embeddings even for a simple concatenation of word embeddings.

### 4.3.6 Processing Input Data

The system supports reading data in CoNLL format which is column-based and has the following properties:

- Each line may have multiple columns one of which contains the token of a sequence. The values in the other columns are usually indices or labels that are associated with the token.

- Sequences of tokens, e.g. sentences, are separated by empty lines.

---

[7] https://www.tensorflow.org/versions/r0.12/api_docs/python/train/gradient_clipping
[8] This is calculated by the TensorFlow's `tf.global_norm`. See https://www.tensorflow.org/api_docs/python/tf/global_norm.

- Columns are separated by whitespace, i.e. a tab or space character. Usually, tabs are used as column delimiters.

For each file, the token and label column indices can be specified. This enables the system to extract the relevant data from the input file and to convert it into a suitable format to serve as input for the neural network.

Once converted, the data is stored in a binary file[9] in order to speed up subsequent experiments that use the same data. Furthermore, an optimized embeddings file is created, when using pre-trained embeddings. This embeddings file only contains word vectors for words which actually occur in the input data of all tasks. This optimized file is also stored on disk.

### 4.3.7 Training

The training process, e.g. batch size, number of epochs, which optimizer to use, is mainly controlled by the user's configuration. In particular, the user can activate *early stopping*. That is, the training process stops if the performance for a specific task and metric on the development dataset stagnates for a specified number of epochs.

On the one hand, *early stopping* may prevent wasting resources on experiments or network configurations that are not promising. On the other hand, this method bears the risk of stopping prematurely while being temporarily stuck at a local minima. Therefore, the user has to carefully choose the number of epochs to wait before stopping early.

During training, the learned model is stored on disk and continuously updated. If early stopping is activated, the stored model is only updated when the performance on the development dataset increases compared to the previously best result. Otherwise, the stored model is updated after every epoch. The stored model can be used to evaluate files from the test dataset or add labels to unlabeled data without having to undergo the training process.

### 4.3.8 Post-Processing and Evaluation

Besides the training process, the MTL sequence tagging framework also manages the evaluation on development and test datasets and can optionally apply simple post-processing rules on prediction results.

After performing a prediction using the model that is currently trained or a pre-trained model loaded from a file, the predictions together with the gold labels are stored in the `ResultList` data structure. This data structure allows to perform various evaluation functions, post-process the predictions, and output the predictions in CoNLL format.

**Metrics**

The system supports the standard evaluation metrics *accuracy*, *precision*, *recall*, and *F1*. Functions from the open source library scikit-learn[10] are used to calculate them.

Furthermore, it supports the AM metrics C-F1 (50%), C-F1 (100%), R-F1 (50%), and R-F1 (100%) which have been presented in section 3.5.2. The implementation by Eger, Daxenberger, and Gurevych (2017) is used for these metrics.

Finally, the S2S metrics *word accuracy* and *edit distance* (cf. section 3.6.3) can be calculated, too. For the latter, the Python library `editdistance`[11] is used. Since the edit distance is calculated per word, the user can choose between calculating the *mean* and *median* edit distance in order to get a score for a complete dataset.

---

[9] `cPickle` (https://docs.python.org/2.3/lib/module-cPickle.html) is used for this purpose.
[10] http://scikit-learn.org/stable/
[11] https://github.com/aflc/editdistance

**Post-Processing**

Post-processing can optionally be applied to data in BIO format prior to calculating *precision*, *recall*, and *F1*. The post-processing corrects invalid "I-" labels. An "I-" label is invalid if it occurs after an "O" label or after an "I-" or "B-" label of a different class, e.g. "I-NN" after "I-VP" for chunking.

The "I-" label correction has two variants: either invalid labels are replaced by "O" labels or the first invalid "I-" label is replaced by a "B-" label. This simple post-processing step can yield performance improvements. The implementation of this process by Reimers and Gurevych (2017a) has been integrated into the system.

In contrast to the previously mentioned metrics, post-processing is mandatory for AM-related metrics because the respective evaluation functions expect a valid label structure. However, it cannot be guaranteed that the neural network produces a valid label structure. For example, a premise label might link to a component beyond the actual text (Eger, Daxenberger, and Gurevych 2017). Therefore, the following post-processing steps suggested by Eger, Daxenberger, and Gurevych (2017) have been implemented:

1. If the BIO structure is invalid, it will be corrected using the second correction variant which has been described previously.

2. If an argumentative component is heterogeneous, e.g. different relation links within a component, the majority labeling within this component is used.

3. If components link beyond the actual text, the links are changed to refer to the closest, maximum permissible component.

Moreover, the evaluation functions expect absolute links instead of the predicted relative links. An implementation for a post-processing step which converts relative links to absolute ones by Eger, Daxenberger, and Gurevych (2017) has thus been integrated as well.

Finally, the calculation of the S2S metrics requires post-processing as well. Since the training data is aligned (cf. section 3.6.2), the network also predicts special alignment symbols, in particular `EMPTY` and `_MYJOIN_`. Before calculating the metrics, these symbols are dropped and the remaining characters are concatenated.

### 4.3.9 Hyper-Parameter Optimization

Hyper-parameter optimization is a crucial part of machine learning. Often, new state-of-the-art results can be achieved with new configurations of existing systems instead of introducing entirely novel approaches (Bergstra, Bardenet, et al. 2011; Hutter, Hoos, and Leyton-Brown 2014).

Since the search space for optimal hyper-parameters can be vast (cf. section 4.3.1), manual optimization is "tedious and often impractical" (Hutter, Hoos, and Leyton-Brown 2011). Therefore, hyper-parameter optimization is usually automated. Although sophisticated approaches such as the multi-stage Bayesian optimization (Wang et al. 2015) are available, grid search is still widely used because it is simple to implement, finds better hyper-parameters compared to manual optimization, and is works well in low dimensional spaces (Bergstra and Bengio 2012).

**Grid and Random Search**

However, Bergstra and Bengio (2012) show that grid search is not optimal for hyper-parameter optimization. While the search spaces are high-dimensional, their effective dimensionality is often low, i.e. not all hyper-parameters and their respective dimensions are equally important. Grid search is inefficient since it wastes too many trials while exploring unimportant dimensions. Important dimensions, on the other hand, are only poorly covered.

Random search explores the search space more efficiently. This is visualized in figure 4.3 for a simple two-dimensional example with an important and an unimportant dimension. The random search explores the important dimension as efficient as if it would explore this dimension in isolation. While the trials are distributed less evenly in the two-dimensional search space, they are more evenly distributed within each individual dimension which results in an improved coverage.

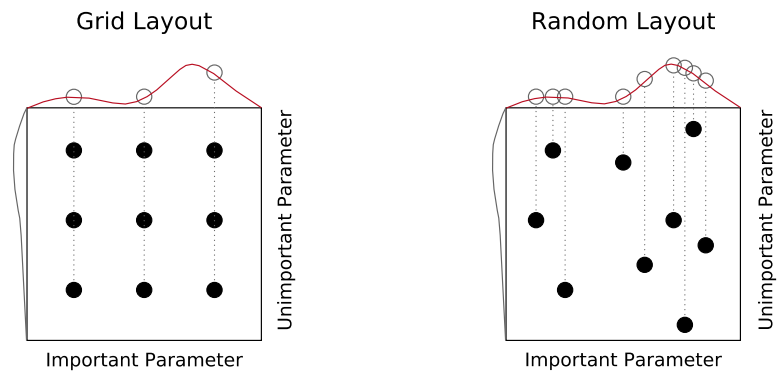

**Figure 4.3.:** Visual comparison of grid and random search in a two-dimensional space by Bergstra and Bengio (2012). For each search, nine trials are performed, but grid search only explores *three* distinct values in the important dimension (red) while random search explores *nine*.

In addition to the improved efficiency, random search allows stopping an experiment at any time and all finished trials form a complete experiment. Trials can be carried out asynchronously and if one trial fails it can be restarted or abandoned without threatening the validity of the experiment. Furthermore, additional trials can be added to the experiment without having to adjust the grid (Bergstra and Bengio 2012).

Given all the aforementioned advantages and that random search is also easy to implement, random search has been chosen for optimizing the parameters of the MTL sequence tagging framework. More advanced optimization techniques are left for future work.

**Configuration**

In order to use the hyper-parameter optimization, the user has to extend the configuration file (cf. section 4.3.1) by specifying variables and intervals from which to sample the variables' values.

**Variables**

Each variable has an individual interval of values. The user can choose between three types of intervals:

*List* – The interval consists of a set of pre-defined values, e.g. {LSTM, GRU}.

*Discrete* – The interval is a range of integers defined by a start and end value (inclusive), e.g. $[0..5]$.

*Continuous* – The interval consists of a range of real values defined by a start and end value (exclusive), e.g. $[0.0, 5.0)$.

**Trials**

Each trial is represented by a configuration file which was generated from a configuration template with variables. To create a trial configuration file, a value is randomly sampled for each variable from its respective interval.

Then, all variables in the configuration template are replaced with their sampled values resulting in a valid configuration file (cf. section 4.3.1) which can be used to train a model.

Creating a trial configuration and running the MTL sequence tagging framework with it, is also automated so that a user only needs to supply a configuration template and its variables and the system carries out the trial itself.

# 5 Experiments

## 5.1 Corpora and Preprocessing

This section provides an overview of all used corpora and performed preprocessing steps to ensure reproducibility of the experiments presented in section 5.5.

Moreover, tables 5.4 – 5.9 provide an overview of the datasets by providing the following information which we use for analyzing the experiment results in chapter 6:

*#Docs* – Depending on the dataset, the number of documents refers to the number of words (S2S datasets), sentences (e.g. POS tagging, syntactic chunking, NER), paragraphs (e.g. AM or Epistemic Segmentation (ES)) or actual documents (e.g. Argument Component Segmentation (ACS) or discourse parsing).

*#Train/Dev/Test* – This column shows how many documents are in the training, development, and test dataset.

*#Tokens* – The number of tokens is another indicator for a dataset's length. This is not the number of unique tokens. A token may be a character (S2S datasets) or a word.

*#Labels* – This is the number of unique labels used in each dataset.

*Entropy and kurtosis* – These metrics have been explained in section 3.3.5.

### 5.1.1 Auxiliary Tasks

Syntactic as well as semantic tasks are used as auxiliary tasks in the experiments. The corpora which provide data for these tasks are described in the following.

**POS Tagging, Syntactic Chunking, and NER**
Data for POS tagging, syntactic chunking, and NER is obtained from the Wall Street Journal (WSJ) portion of the Penn Treebank (PTB) (Marcus, Santorini, and Marcinkiewicz 1993; Ma and E. Hovy 2016) and the English data used in the CoNLL 2003 shared task (Tjong Kim Sang and De Meulder 2003). No additional preprocessing has been performed for these corpora. For the CoNLL 2003 data, the same splits as in the shared task are used. The splits for WSJ are as in Ma and E. Hovy (2016).

**Discourse Parsing**
Discourse parsing is also utilized as an auxiliary task. In essence, discourse parsing aims to model the understanding of coherent texts. A coherent text consists of interdependent sentences and discusses a specific topic. Such texts contain complete, distinct units of information so-called Elementary Discourse Units (EDUs) which have relationships with each other that are meaningful to the texts' topic and need to be uncovered in order to fully understand a text (Stede 2011).

As stated independently by Braud, Lacroix, and Søgaard (2017) and Stab and Gurevych (2017), discourse parsing is related to AM making it a promising candidate for suitable AM auxiliary tasks. Since discourse parsing emphasizes modeling relationships between text components, i.e. EDUs, it might be capable of improving a learners relation modeling capability which is the main motivation for employing them as auxiliary tasks.

As in Braud, Plank, and Søgaard (2016), the corpora Rhetorical Structure Theory Discourse Treebank (RST-DT) (Carlson, Marcu, and Okurowski 2001) and Penn Discourse Treebank (PDTB) (Prasad et al. 2008) are used. They differ in the discourse structure theory that underlies the annotations in the datasets: RST-DT is based on RST (Mann and Thompson 1988) while PDTB adopts "a theory-neutral approach to the annotation" (Prasad et al. 2008).

Both discourse parsing datasets are not in CoNLL format as required by the MTL sequence tagging framework (cf. section 4.3.6). Moreover, it is not straightforward to convert RST-DT and PDTB into CoNLL format. Therefore, the corpora were first preprocessed as in Braud, Plank, and Søgaard (2016)[12]. After this preprocessing, the datasets can be characterized as follows:

*RST-DT* – Each line contains an EDU labeled with its "local surrounding discourse structure" (Braud, Plank, and Søgaard 2016). That is, EDUs can either start (opening parenthesis followed by the relation name) or close (the relation name followed by a closing parenthesis) a relationship. Since EDUs can form composite EDUs by building relationships between pairs of EDUs (Stede 2011), a single EDU may open multiple relationships. A document consists of a continuous sequence of EDUs and documents are separated by empty lines. Besides the relation, the label also describes the importance of its involved EDUs, i.e. whether they are nucleus N (primary information) or satellite S (secondary information). More detailed information on this data format is provided by Braud, Plank, and Søgaard (2016). An example is given in table 5.1 and visualized in figure 5.1.

| EDU | Discourse Structure |
|---|---|
| GM officials want to get their strategy | `(SN-Background(NN-Same-unit(NS-Elaboration` |
| to reduce capacity and the work force | `NS-Elaboration)` |
| in place | `NN-Same-unit)` |
| before those talks begin . | `SN-Background))` |

**Table 5.1.:** Document-view of a sentence consisting of four EDUs labeled with the RST-DT label scheme. The EDUs form a sub-tree with four levels.

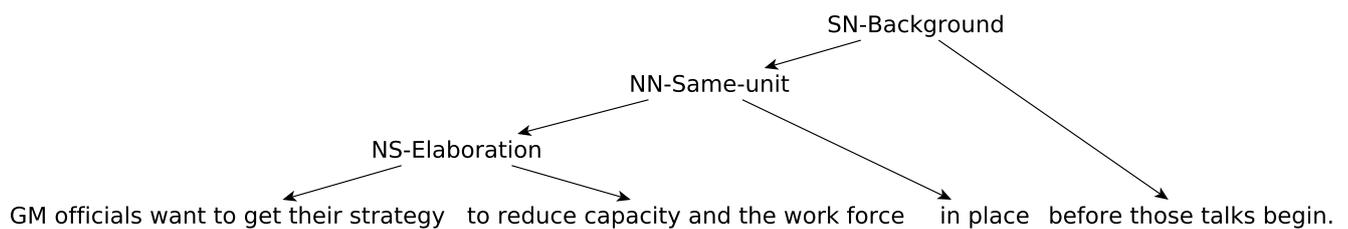

**Figure 5.1.:** Tree-view of a sentence consisting of four EDUs labeled with the RST-DT label scheme.

*PDTB* – Each line contains a sentence labeled with the relation it participates in and the span it covers in the original text. The span annotation is ignored in subsequent preprocessing steps. Relations span across multiple sentences. The labels are in BIO format. Hence, sentences that are not involved in any relation are labeled with an *O*. Documents are separated by empty lines. An example is given in table 5.2.

---

[12] RST-DT was preprocessed with the scripts available at `https://bitbucket.org/chloebt/discourse`. This repository, however, did not contain a preprocessing script for PDTB, but Chloé Braud (Braud, Plank, and Søgaard 2016) very kindly provided the preprocessed data.

| EDU | Relation | Span |
|---|---|---|
| Wang had previously forecast a loss . | `B-entrel` | 304..340 |
| The company reiterated that it expects another loss [...] . | `I-entrel` | 341..486 |
| A year ago , Wang had [...] $ 3.1 million loss from discontinued operations . | `B-entrel` | 489..642 |
| The latest period loss included a $ 12.9 pretax charge [...] | `I-entrel` | 643..720 |

Table 5.2.: Example from the PDTB dataset with two pairs of related sentences. Both pairs have the *EntRel* relationship, i.e. an entity-based coherence relation exists between their respective sentences (Prasad et al. 2008).

| RST-DT | | PDTB | |
|---|---|---|---|
| Token | Label | Token | Label |
| in | `B-NN-Same-unit)` | Wang | `B-entrel` |
| place | `I-NN-Same-unit)` | had | `I-entrel` |
| before | `B-SN-Background))` | previously | `I-entrel` |
| those | `I-SN-Background))` | forecast | `I-entrel` |
| talks | `I-SN-Background))` | a | `I-entrel` |
| begin | `I-SN-Background))` | loss | `I-entrel` |
| . | `I-SN-Background))` | . | `I-entrel` |
| | | The | `I-entrel` |
| | | company | `I-entrel` |
| | | reiterated | `I-entrel` |

Table 5.3.: Example of discourse parsing datasets in CoNLL format with BIO labels. Only an excerpt of the previous examples in tables 5.1 and 5.2 is shown. The remaining sentences are labeled analogously.

In both datasets, EDUs and sentences are already tokenized. The conversion into CoNLL format is performed by the following steps: 1) Extract the EDU/sentence and label (discourse structure for RST-DT and relation for PDTB) from each line. 2) Split the EDU/sentence into its tokens. 3) Label each token according to the EDU/sentence label. The token labels are in BIO format. Given the two examples in tables 5.1 and 5.2, excerpts from the final datasets after the conversion are shown in table 5.3.

After these steps, all datasets for auxiliary tasks are in the final format which is used in the experiments. An overview of them is provided in table 5.4.

### 5.1.2 Argumentation Mining

The Persuasive Essays (PE) dataset by Stab and Gurevych (2017) is used for AM experiments. Instead of using the segmentation into essays resulting in 402 documents (Stab and Gurevych 2017), this thesis follows Eger, Daxenberger, and Gurevych (2017), i.e. the texts are segmented into 2235 paragraphs (cf. table 5.5). Except for the relation between claims and major claims, a paragraph always contains a complete argument structure. However, predicting this structure on the paragraph level is easier than on the essay level due to fewer components and thus fewer possible combinations for relations (Eger, Daxenberger, and Gurevych 2017).

| Name | #Docs | #Train / Dev / Test | #Tokens | #Labels | Entropy | Kurtosis |
|---|---|---|---|---|---|---|
| WSJ (POS) | 49 208 | 38 219 / 5527 / 5462 | 1 173 766 | 46 | 4.339 | 2.054 |
| WSJ (chunking) | 49 208 | 38 219 / 5527 / 5462 | 1 173 766 | 23 | 2.654 | 1.478 |
| CoNLL 2003 (NER) | 20 744 | 14 041 / 3250 / 3453 | 30 289 | 9 | 1.122 | 11.001 |
| PDTB | 3309 | 1103 / 1103 / 1103 | 1 784 796 | 27 | 2.480 | 3.328 |
| RST-DT | 385 | 277 / 70 / 38 | 206 306 | 4398 | 7.770 | 2.075 |

Table 5.4.: Corpora for auxiliary tasks

| Name | #Docs | #Train / Dev / Test | #Tokens | #Labels | Entropy | Kurtosis |
|---|---|---|---|---|---|---|
| Persuasive Essays (PE) | 2235 | 1587 / 199 / 449 | 148 182 | 73 | 3.521 | 5.279 |

Table 5.5.: Corpus for Argumentation Mining

An example for the data format of this dataset has been provided in section 3.5.1. Statistics for this dataset can be found in table 5.5.

### 5.1.3 Epistemic Segmentation

Epistemic Segmentation (ES) is a task related to the research of Scientific Reasoning and Argumentation (SRA) which is concerned with the processes of understanding, evaluating, assessing, and generating scientific knowledge especially through collaboration (Fischer et al. 2014).

In SRA research the motivations of people for engaging in SRA are characterized with different *epistemic modes*, while the scientific activities of this engagement are called *epistemic activities*. Activities can be, for instance, problem identification, questioning, and generating hypotheses (Fischer et al. 2014).

The task of epistemic segmentation aims to model SRA processes which can be observed in discussions, for example. This allows to better understand the cognition and reasoning process by identifying *propositional units* in the dialogue and assigning epistemic activities to them (Csanadi, Kollar, and Fischer 2016). Propositional units do not necessarily cover a complete sentence. In fact, a single sentence may consist of several propositional units (Ghanem et al. 2016).

**Datasets**

The datasets Teacher Students (TS) (Csanadi, Kollar, and Fischer 2016), Social Workers (SW) (Ghanem et al. 2016), and Medicine (MED) (Lenzer et al. 2017) all contain transcriptions of dialogues which are segmented into propositional units and labeled with the eight epistemic activities defined by Fischer et al. (2014). The latter dataset consists of English text, while the others are in German. All of them are summarized in table 5.6.

**Preprocessing**

Since the datasets are not in CoNLL format, preprocessing steps were necessary. TS and SW are similar so that the preprocessing is almost analogous for them. They only differ slightly in their label inventories. For instance, both have fine-grained labels for the activity *evidence generation* modeling differences such as anecdotal and scientific evidence (Ghanem et al. 2016), but the label text differs, e.g. `EGscientific` and `SEG` respectively for scientific evidence.

| Name | #Docs | #Train / Dev / Test | #Tokens | #Labels | Entropy | Kurtosis |
|---|---|---|---|---|---|---|
| Medicine (MED) | 462 | 295 (1442) / 74 / 93 | 45 366 | 15 | 2.502 | 2.372 |
| Social Workers (SW) | 57 | 36 (1149) / 9 / 12 | 39 557 | 17 | 2.978 | 3.408 |
| Teacher Students (TS) | 46 | 28 (1152) / 8 / 10 | 54 512 | 19 | 2.659 | 1.430 |

**Table 5.6.:** Corpora for Epistemic Segmentation. The number of training documents in parenthesis refers to the new segmentation by sentences and their context with a context size of $n = 2$.

Both datasets consist of multiple text files each representing a document with one propositional unit per line. The units are labeled with the epistemic activities. Neutral units, i.e. units which do not represent any epistemic activity, are labeled as *non-epistemic*.

An example from the from the TS dataset with the activities *communicating and scrutinizing* (CS) and *construction and redesign of artifacts* (CA) is shown below.

| **Propositional Unit** | **Epistemic Activity** |
|---|---|
| Als Lehrerin würde ich [...] mit ihr ins Gespräch kommen, | //;CS |
| auf diese Dinge aufmerksam machen. | //;CA |

The conversion into the CoNLL format with BIO labels has been performed as follows: 1) Tokenization of the text using NLTK's `TreebankWordTokenizer`[13], 2) assignment of labels to tokens instead of propositional units, and 3) conversion of labels into valid BIO labels.

The MED dataset was provided in a single Comma Separated Values (CSV) file with multiple columns including "Text" (the propositional units), "Epistemischer Schritt" (the epistemic activities), and "Sequenz_fortlaufend" (continuous sequence numbers).

The sequences described by the sequence numbers span across multiple propositional units. Since there are no references for a segmentation of this dataset, a custom segmentation has been applied. That is, each sequence is considered a document resulting in 462 documents for this dataset. Afterwards, the conversion into CoNLL format was analogous to the other datasets.

**Labels**

As mentioned before, the label inventories of the datasets differed slightly although all of them refer to the same epistemic activities. Therefore, the dataset-specific labels have been mapped to the following unified labels: problem identification (`PI`), questioning (`Q`), evidence generation (`EG`), evidence evaluation (`EE`), hypothesis generation (`HG`), construction and redesign of artifacts (`CA`), drawing conclusions (`DC`), and communicating and scrutinizing (`CS`). Moreover, the non-epistemic label for each dataset has been associated with the `O` label. The previous example is thus labeled as follows:

| B-CS | I-CS | I-CS | I-CS | [···] | I-CS | I-CS | I-CS | I-CS | I-CS | I-CS |
|---|---|---|---|---|---|---|---|---|---|---|
| Als | Lehrerin | würde | ich | [···] | mit | ihr | ins | Gespräch | kommen | , |

| B-CA | I-CA | I-CA | I-CA | I-CA | I-CA |
|---|---|---|---|---|---|
| auf | diese | Dinge | aufmerksam | machen | . |

---
[13] http://www.nltk.org/_modules/nltk/tokenize/treebank.html#TreebankWordTokenizer

Despite the unified label inventory, not all datasets have the same number of distinct labels. Due to the BIO labeling, one would expect $8 \cdot 2 + 1 = 17$ unique labels. However, the MED dataset has no unit labeled with `CA` and the TS dataset has a single unit labeled with `NA` resulting in 15 and 19 labels respectively.

**Segmentation of Training Data**

Finally, the segmentation of the training set has been adapted. Initial tests on these datasets showed poor performance. The learner was never able to beat the majority baseline. Since the documents are very long, a new segmentation into shorter units which lowers the burden to model long range dependencies has been investigated[14].

Using NLTK's `PunktSentenceTokenizer`[15] the documents are split into sentences. For each document, new documents are created by iterating over its sentences and constructing a new document as the concatenation of the current sentence with $n$ previous and $n$ next sentences. The context size $n$ is configurable and has been chosen as $n = 2$ for the experiments in this thesis. Experiments with $n = 1$ and $n = 3$ yielded less promising results.

This conversion resulted in more, but considerably smaller documents. The labels remain unchanged. Table 5.6 lists the new number of documents for the training sets in parentheses. The development and test sets have not been changed to ensure that the learner can still be evaluated on the original segmentation proposed by the authors of the datasets – at least for TS and SW.

### 5.1.4 Argument Component Segmentation

Argument Component Segmentation (ACS) is the task of segmenting a text into argumentative and non-argumentative units. In this thesis, it is equivalent to using just the $b$-element of the AM label four-tuple (cf. section 3.5.1). Therefore, the available labels are B-Arg, I-Arg, and O. This is similar to argument component identification (Persing and Ng 2016), but without classifying the argument components.

For the experiments, a subset of the datasets used by Daxenberger et al. (2017) is employed, i.e. Wikipedia Talk Pages (WTP) by Biran and Rambow (2011), AraucariaDB (ADB) by Reed et al. (2008), and Persuasive Essays (PE:ACS) by Stab and Gurevych (2017). In difference to the PE dataset used for AM, prediction is performed on the essay level in PE:ACS.

| Name | #Docs | #Train / Dev / Test | #Tokens | #Labels | Entropy | Kurtosis |
|---|---|---|---|---|---|---|
| Wikipedia Talk Pages (WTP) | 1972 | 1428 / 358 / 186 | 187 846 | 3 (32% BI, 68% O) | 0.988 | 1.623 |
| AraucariaDB (ADB) | 502 | 360 / 91 / 51 | 59 053 | 3 (40% BI, 60% O) | 1.125 | 1.250 |
| Persuasive Essays (PE:ACS) | 402 | 257 / 65 / 80 | 147 271 | 3 (36% BI, 64% O) | 1.131 | 1.492 |

Table 5.7.: Corpora for argument component segmentation. In addition to the total number of labels (three in each dataset), the ratio between `O` labels and the other BIO labels (`BI`) is provided.

While Daxenberger et al. (2017) predict whether or not a sentence contains a claim, all types of components, i.e. major claims, claims, and premises (cf. section 3.5.1), are included and the learner has to predict the exact span of a component in this thesis, i.e. the prediction task is more fine-grained.

---

[14] The proposal for the new segmentation and a program to perform it was kindly provided by Steffen Eger. The program has been adapted to allow the configuration of the context size instead of a fixed context size of 1.

[15] `http://www.nltk.org/_modules/nltk/tokenize/punkt.html#PunktSentenceTokenizer`

Components may span across sentences, consist of a single sentence or cover only a part of a sentence. An overview of the three datasets is provided in table 5.7.

### 5.1.5 Sequence-to-Sequence – Grapheme-to-Phoneme Conversion

Finally, the datasets for the S2S experiments are G2P (cf. section 3.6.1) datasets on the one hand and on the other hand a dataset of word segmentations extracted[16] from the Combilex dataset (Richmond, R. A. J. Clark, and Fitt 2009). The G2P datasets are CMU by the Carnegie Mellon University[17], Celex by Baayen, Piepenbrock, and Gulikers (1993), and Combilex by Richmond, R. A. J. Clark, and Fitt (2009).

Since the MTL sequence tagging framework can only solve sequence tagging problems (cf. chapter 4), the training data of the G2P tasks has to be aligned (cf. section 3.6.2). The alignment has been performed with a script provided by Steffen Eger[18]. This script uses the aligner by Jiampojamarn, Kondrak, and Sherif (2007) to align graphemes and phonemes as in Schnober et al. (2016), i.e. we disallow many-to-1 or many-to-many matches between the grapheme and phoneme string because supporting them would require our system to segment the input string into grapheme groups with a length $\geq 1$. This cannot be done reliably and therefore would be an additional source of errors. Development and test sets remain unchanged. As an example, the aligned and unaligned phoneme strings for the word "exiting" are presented in table 5.8.

|  | Unaligned | Aligned |
|---|---|---|
| **Celex** | E g z @ t I N | E g_z @ t I $\epsilon$ N |
| **CMU** | EH G Z AH T IH NG | EH G_Z AH T IH NG $\epsilon$ |
| **Combilex** | E g z @ t I N | E g_z @ t I $\epsilon$ N |

**Table 5.8.:** Aligned and unaligned phoneme strings for the word "exiting". $\epsilon$ is a special alignment symbol representing the empty string and _ is used to merge two phonemes.

Furthermore, the word segmentation dataset required preprocessing as well. The dataset provides information about the morphemic (morph), phonetic (phon), and syllable (syll) segmentation of words. This information is encoded in a tree structure. For example, the word "vandalise" (British English spelling) is visualized in figure 5.2. To frame word segmentation as a sequence tagging problem, the segmentation information can be encoded as a binary string (Eger 2013). This representation can be derived directly from the tree structure by finding all letters of the phonemes resulting in the phonetic segments, then finding the phonemes for each syllable, and finally finding all syllables for the morphemes.

The resulting segmentations can be written with the common hyphen-notation, i.e. "vandal-ise" (morphemic segmentation), "van-dal-ise" (syllable segmentation), and "v-a-n-d-al-i-s-e" (phonetic segmentation), but also be encoded as binary by representing each character by a bit and marking the start of each new segment with a 1 except for the first segment. That is, `000000100`, `000100100`, and `011110111` respectively.

For all G2P datasets, a custom split into training, development, and test set has been performed using 64%, 16%, and 20% of the words respectively. The word segmentation dataset was already split into training, development, and test sets.

---

[16] The Ubiquitous Knowledge Processing (UKP) lab at Technische Universität Darmstadt (TUD) extracted the word segmentations and converted them into a tree structure.
[17] `http://www.speech.cs.cmu.edu/cgi-bin/cmudict`
[18] `https://github.com/UKPLab/coling2016-pcrf-seq2seq`

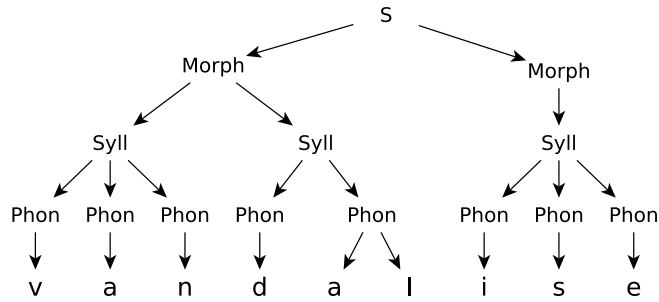

**Figure 5.2.:** Morphemic, syllable, and phonetic segmentation of the word "vandalise" (British English spelling) visualized as a tree.

| Name | #Docs | #Train / Dev / Test | #Tokens | #Labels | Entropy | Kurtosis |
|---|---|---|---|---|---|---|
| CMU | 113 413 | 72 576 / 18 150 / 22 687 | 849 273 | 299 | 4.699 | 1.982 |
| Celex | 29 998 | 19 198 / 4800 / 6000 | 245 342 | 224 | 4.988 | 1.587 |
| Combilex | 128 604 | 82 306 / 20 577 / 25 721 | 1 112 624 | 319 | 4.982 | 1.617 |
| Split (Morph) | 126 479 | 78 655 / 21 330 / 26 494 | 1 109 378 | 2 | 0.512 | 6.898 |
| Split (Syll) | 126 479 | 78 655 / 21 330 / 26 494 | 1 109 378 | 2 | 0.815 | 2.300 |
| Split (Phon) | 126 479 | 78 655 / 21 330 / 26 494 | 1 109 378 | 2 | 0.807 | 2.368 |

**Table 5.9.:** Corpora for Sequence-to-Sequence tasks

## 5.2 Word Embeddings

In the experiments, three different word embeddings are utilized. However, no performance comparison of different embeddings has been attempted. For a comparison see Reimers and Gurevych (2017b).

**Extended Dependency Skipgram Embeddings**

Using word embeddings trained with the model by Komninos and Manandhar (2016) has shown to be superior to the use of other word embeddings for sequence tagging tasks in the experiments by Reimers and Gurevych (2017b). Therefore, all mono-lingual, i.e. English, experiments employ these embeddings. They are trained on the English Wikipedia dump (August 2015) which consists of 2 billion words. The resulting vectors have a dimensionality of 300 (Komninos and Manandhar 2016).

**Bilingual Embeddings**

The bilingual embeddings trained by Lee (2017) are used for all experiments which involve German and English data. The embeddings have been trained on the *Europarl* corpus (Koehn 2005) with the BilBOWA model (Gouws, Bengio, and Greg Corrado 2015) for 5 epochs resulting in 200 dimensional vectors. All German words are suffixed with "_de" and all English words with "_en".

**Character Embeddings**

Since the S2S datasets provide characters instead of words as inputs, conventional word embeddings are of no use. Therefore, character embeddings are utilized in these experiments. Woolf (2017) created these embeddings based on the pre-trained GloVe embeddings (Pennington, Socher, and Manning 2014) by extracting character embeddings as the average of the vectors of all words in which a specific character occurs. For example, the vector for the character "t" contains, among others, the vectors of the words "the" and "training". In particular, he used the 300 dimensional vectors trained on 840 billion tokens of web data because they contain uppercase letters (Pennington, Socher, and Manning 2014; Woolf 2017).

## 5.3 Hyper-Parameter Optimization and Evaluation

Section 5.5 will present all experiment setups. To ensure that every setup has equal chances of finding a good hyper-parameter configuration, each setup is granted *ten* trials in the random search resulting in *ten* different trial hyper-parameter configurations. Each trial hyper-parameter configuration is run *three* times, i.e. *three* different random initializations of the model's weights are evaluated. The evaluation results on the development dataset are averaged.

After finishing all trials, the hyper-parameter configuration with the best results on the development dataset is used for the actual experiment setup evaluation. Using this configuration, the experiment setup is trained *ten* times with the same hyper-parameter configuration, but different weight initializations.

Since neural networks are non-deterministic learners due to the random initialization of weights, the learned models will have varying evaluation results (cf. Reimers and Gurevych 2017b). After the training, each model is evaluated on the test dataset and the evaluation results are averaged.

## 5.4 Different Architectures

To evaluate not only potential performance differences between MTL and STL, but also whether the MTL sequence tagging framework achieves competitive performance, the experiments are also performed on other architectures namely the STL LSTM-CRF by Lample et al. (2016) and the MTL architecture of Reimers and Gurevych (2017a).

The hyper-parameter optimization of the MTL sequence tagging framework has been adapted to be capable of supporting these architectures as well. However, the architectures have different ranges for the optimization trials because they support different configuration options. Moreover, the trials in the hyper-parameter optimization are only run once instead of thrice (cf. section 5.3). The evaluation of the best configuration, nonetheless, is averaged across ten runs in order to report more reliable scores (cf. Reimers and Gurevych 2017b).

Furthermore, minor changes regarding logging, handling input and embedding files, and the output format have been implemented in order to streamline the experimentation process with all architectures.

## 5.5 Experiment Setups

This section provides an overview of all experiment setups. An experiment setup is one combination of a main task and potentially several auxiliary tasks or no auxiliary tasks in case of STL. The setups are grouped into four experiment settings which we present in sections 5.5.1 – 5.5.4. We evaluate each setup on an individual hyper-parameter combination which is determined by the hyper-parameter optimization process for this setup (cf. section 5.3).

An experiment setup is referred to with the following notation: $\texttt{Setting}_{\text{Auxiliary Tasks}}^{\text{Main Task}}$. The setting is one of `AM`, `ES`, `ACS`, and `G2P`, the main task is specified as the name of the dataset which provides the data for the main task, and the auxiliary tasks are a comma separated list of either dataset names or task names. Task names are used for POS tagging, syntactic chunking, and NER (PCN), natural subtasks, and word segmentation tasks, i.e. whenever the dataset name alone would be ambiguous[19]. In an STL setup, no auxiliary tasks are listed.

### 5.5.1 Multi-Task Argumentation Mining

The first investigated experiment setting is multi-task Argumentation Mining (AM). This is also the most exhaustive of all experiment settings in this thesis and to our knowledge the most comprehensive investigation of the MTL paradigm for `AM` yet. The main task in this setting is `AM` on the PE dataset.

---
[19]  For instance, the natural subtasks in the multi-task AM setting are all from the PE dataset.

**Auxiliary Tasks**

In this experiment setting, various auxiliary tasks are used. The required datasets have been explained in section 5.1.1. First of all, the common NLP tasks POS tagging, syntactic chunking, and NER (PCN) are used as auxiliary tasks. Since these tasks are fundamental and thus omnipresent in NLP, they could prove to be beneficial to a wide range of NLP problems including AM. As elaborated in section 2.2.2, these tasks were already helpful auxiliary tasks in related work even for semantic tasks.

Further, discourse parsing is used as an auxiliary task because it may enable the learner to model relationships between argumentative components more effectively (cf. section 5.1.1). Both datasets, PDTB and RST-DT, are utilized.

Finally, natural subtasks (cf. section 2.2.5) are used. We leverage the AM label scheme's suitability for deriving natural subtasks to create the following natural subtasks[20]:

*Argument Component Segmentation (ACS)* – Use only the $b$-element of the AM label four-tuple. The labels are `B-Arg`, `I-Arg`, `O`.

*Argument Component Identification (ACI)* – Use the elements $b$ and $t$ of the four-tuple, e.g. `I-Claim` or `B-Premise`.

*Argument Relation Segmentation (ARS)* – Find the spans of argument relations. This is similar to ACS, but it excludes major claims as they have no outgoing relations. The labels are `B-Rel`, `I-Rel`, `O`.

*Argument Relation Identification (ARI)* – This task is derived by using the elements $b$, $t$, and $s$ of the four-tuple. In other words, it is equivalent to the main task except for the prediction of distances between the components. Labels are, for instance, `I-Claim:For`, `B-MajorClaim` or `I-Premise:Support`.

These auxiliary tasks are used in different combinations in the experiment setups presented in table 5.10.

**Metrics**

The `AM` metrics C-F1 (50%), C-F1 (100%), R-F1 (50%), and R-F1 (100%) that have been discussed in section 3.5.2 are used to evaluate the main task. The auxiliary tasks are evaluated using F1, precision, recall, and accuracy. However, a detailed report of the scores achieved by the subtasks is not in the scope of this thesis because only the performance of the main task is of interest in this experiment setting.

### 5.5.2 Cross-Lingual Epistemic Segmentation

Since there are one English and two German `ES` datasets, a cross-lingual experiment can be conducted. For this purpose, the bilingual embeddings (cf. section 5.2) are used.

In order to reduce the required time for the hyper-parameter search, it is only performed on 30% of the training data. Moreover, this experiment setting is only evaluated on the MTL sequence tagging framework and the architecture by Lample et al. (2016).

**Auxiliary tasks**

MTL can also utilize other languages as auxiliary tasks (cf. section 2.2.1). In this case, the English MED dataset is used as an auxiliary task while the German datasets TS and SW are used as main tasks, but our MTL setups only use one German dataset at a time. Although it is only treated as an auxiliary task, the MED task could also benefit from being learned together with either main task.

---

[20] In their MTL experiments, Eger, Daxenberger, and Gurevych (2017) use two natural subtasks: the first one uses the label set $\{(b,t) \mid (b,t,d,s) \in Y\}$, i.e. is equivalent to Argument Component Identification (ACI), and the second one uses $\{(d,s) \mid (b,t,d,s) \in Y\}$. The definition of $Y$ is provided in section 3.5.1.

| Name | Auxiliary Tasks | Remarks |
|---|---|---|
| $\text{AM}^{\text{PE}}$ | – | – |
| $\text{AM}^{\text{PE}}_{\text{PCN}}$ | POS tagging, syntactic chunking, and NER (PCN) | – |
| $\text{AM}^{\text{PE}}_{\text{PDTB}}$ | Discourse parsing with the dataset PDTB | Only 10% of the PDTB training data is used so that the number of documents for the auxiliary and the main task is approximately equivalent. |
| $\text{AM}^{\text{PE}}_{\text{PDTB, RST-DT}}$ | Discourse parsing with the datasets PDTB and RST-DT | Only 10% of the PDTB and 30% of the RST-DT training data are used. |
| $\text{AM}^{\text{PE}}_{\text{Subtasks}}$ | All natural subtasks, i.e. ACS, ACI, ARS, and ARI. | – |
| $\text{AM}^{\text{PE}}_{\text{Subtasks, PCN}}$ | All natural subtasks, POS tagging, chunking, and NER | – |
| $\text{AM}^{\text{PE}}_{\text{Subtasks, PDTB}}$ | All natural subtasks and discourse parsing with PDTB | Only 10% of the PDTB training data is used. |

**Table 5.10.:** Setups for the multi-task AM experiment using the Persuasive Essays (PE) dataset. POS tagging, syntactic chunking, and NER (PCN) (WSJ and CoNLL 2003 datasets), discourse parsing (Penn Discourse Treebank (PDTB) and Rhetorical Structure Theory Discourse Treebank (RST-DT) datasets), and natural subtasks are used as auxiliary tasks.

Using the notation introduced before, $\text{ES}^{\text{TS}}_{\text{MED}}$ refers to the experiment setup with ES on the TS dataset as the main task and ES on the MED dataset as the auxiliary task. Since the early stopping regularization only takes the main task into account, the potential setup $\text{ES}^{\text{MED}}_{\text{TS}}$ is not entirely equivalent because it indicates MED being the main task. For notational convenience in the result tables in section 5.6, we refer to the setup $\text{ES}^{\text{TS}}_{\text{MED}}$ as $\text{ES}^{\text{MED*}}_{\text{TS}}$ if the performance on the MED dataset is evaluated, i.e. the asterisks (*) indicates that MED is *not the main task*. $\text{ES}^{\text{SW}}_{\text{MED}}$ and $\text{ES}^{\text{MED*}}_{\text{SW}}$ refer to each other analogously.

Table 5.11 lists all setups in this setting. In addition to the default STL baseline, i.e. training the main task in isolation, a second STL baseline is added: the main task, i.e. TS or SW is trained not only on its own training data, but on the union of its own training data and the training data of MED, i.e. the auxiliary task. This training data union setup is indicated by +MED in the superscript of the experiment setup notation. This baseline is used to verify that MTL is not just successful because noise is added to the training data or the amount of training data has been increased. This is similar to the $S+A$ baseline of Bollmann and Søgaard (2016).

**Metrics**

The metrics F1, precision, and recall are used for this experiment setting.

### 5.5.3 Cross-Domain Argument Component Segmentation

Multiple datasets from different domains are available for the task of ACS. This enables cross-domain experiments which have been introduced in section 2.2.1.

To ensure that the number of documents used for training is roughly equivalent for all domains, 100%, 70%, and 20% of the training data of PE:ACS, ADB, and WTP are used.

| Name | Auxiliary Tasks | Remarks |
| --- | --- | --- |
| $\text{ES}^{\text{TS}}$ | – | – |
| $\text{ES}^{\text{TS}}_{\text{MED}}$ | ES on the medicine dataset | Also referred to as $\text{ES}^{\text{MED}*}_{\text{TS}}$. |
| $\text{ES}^{\text{SW}}$ | – | – |
| $\text{ES}^{\text{SW+MED}}$ | – | Baseline trained on the union of the SW and the MED training data. |
| $\text{ES}^{\text{SW}}_{\text{MED}}$ | ES on the medicine dataset | Also referred to as $\text{ES}^{\text{MED}*}_{\text{SW}}$. |

**Table 5.11.:** Setups for the cross-lingual ES experiment using the datasets Teacher Students (TS), Social Workers (SW), and Medicine (MED).

This experiment setting is only evaluated on the MTL sequence tagging framework and the architecture by Reimers and Gurevych (2017a) to reduce the required computation time and memory consumption[21] for this experiment.

**Auxiliary tasks**

In addition to using other languages as auxiliary tasks (cf. section 5.5.2), utilizing different domains is a potential application area of MTL as well (cf. section 3.3.2). In this experiment, the corpora ADB by Reed et al. (2008), WTP by Biran and Rambow (2011), and PE:ACS by Stab and Gurevych (2017) are used because they have a similar label distribution. In particular, they have comparably few out-of-span, i.e. 0, labels (cf. table 5.7).

In the experiment setups, PE:ACS and WTP are used as main tasks similarly to TS and SW in the multi-lingual ES setting. ADB is used as an auxiliary task in both cases. Again, the asterisks notation is used when the auxiliary task is evaluated, e.g. $\text{ACS}^{\text{WTP}}_{\text{ADB}}$ and $\text{ACS}^{\text{ADB}*}_{\text{WTP}}$ refer to the same setup, but the latter notation is used in the evaluation on the ADB dataset. Analogous to the previous experiment setting (cf. section 5.5.2), additional STL baselines which use a union of training data from different domains are evaluated. The different experiment setups emerging from these tasks are listed in table 5.12.

**Metrics**

The metrics F1, precision, and recall are used for this experiment setting.

### 5.5.4 Multi-Task Grapheme-to-Phoneme Conversion

The final experiment setting is quite different from the previous ones. It shall evaluate the effectiveness of MTL for S2S tasks that have been framed as sequence tagging problems. A major goal of extending the thesis' scope to include this task which is rather unrelated to AM is to assess whether the MTL paradigm generalizes well, i.e. whether it is applicable to a diverse set of NLP problems.

In particular, we investigate G2P conversion using the datasets presented in section 5.1.5, i.e. Celex, CMU, and Combilex. Since the G2P and word segmentation datasets are large, only 10% of each dataset's training data (40% for Celex) is used for the hyper-parameter optimization.

Moreover, we evaluate the best hyper-parameter configuration on different amounts of training data to investigate potential correlations between the number of training samples and the success of the MTL paradigm. This comparison is not optimal because it may not be entirely fair. Larger training datasets might profit from different hyper-parameter configurations, e.g. less regularization. For a fair comparison, an additional hyper-parameter search would be required for each training dataset size.

---

[21] The system by Lample et al. (2016) at times consumed up to 250 GB of memory in this experiment setting. This slowed down the evaluation server. Analyzing and solving this problem is not in the scope of this thesis.

| Name | Auxiliary Tasks | Remarks |
| --- | --- | --- |
| ACS$^{\text{ADB}}$ | – | – |
| ACS$^{\text{ADB+PE:ACS}}$ | – | Baseline trained on the union of the ADB and the PE:ACS training data. |
| ACS$^{\text{WTP}}$ | – | – |
| ACS$^{\text{WTP+ADB}}$ | – | Baseline trained on the union of the WTP and the ADB training data. |
| ACS$^{\text{WTP}}_{\text{ADB}}$ | ACS on the ADB dataset | Also referred to as ACS$^{\text{ADB}*}_{\text{WTP}}$. |
| ACS$^{\text{PE:ACS}}$ | – | – |
| ACS$^{\text{PE:ACS+ADB}}$ | – | Baseline trained on the union of the PE:ACS and the ADB training data. |
| ACS$^{\text{PE:ACS}}_{\text{ADB}}$ | ACS on the ADB dataset | Also referred to as ACS$^{\text{ADB}*}_{\text{PE:ACS}}$. |

**Table 5.12.:** Setups for the cross-domain ACS experiment using the datasets AraucariaDB (ADB) by Reed et al. (2008), Wikipedia Talk Pages (WTP) by Biran and Rambow (2011), and Persuasive Essays (PE:ACS) by Stab and Gurevych (2017).

Repeating these experiment setups with individual hyper-parameter optimizations is, however, left for future work since it would exceed the scope of this thesis.

**Auxiliary tasks**

In this experiment setting, the word segmentation dataset will be the provider of auxiliary tasks. We leave possible multi-domain setups with the used datasets for future work. The intuition behind using morphemic, syllable, and phonetic segmentations as auxiliary tasks is that segmentations are highly informative about the pronunciation of words and in particular for groups of letters which are frequently joined into one phoneme, e.g. "sh".

| Word | Syllable Split | IPA | Arpabet |
| --- | --- | --- | --- |
| archbishop | arch-bishop | ˌɑːtʃˈbɪʃəp | AA R CH B IH SH AH P |
| grasshopper | grass-hop-per | ˈɡrɑːsˌhɒpə(r) | G R AE S HH AA P ER |

**Table 5.13.:** Syllable segmentations and pronunciations of the words "archbishop" and "grasshopper". The syllable segmentations are from Weis, Enz, and Schnorr (2003) and the pronunciations are in the Arpabet format (Klatt 1977) as used in the CMU dataset and the IPA format (International Phonetic Association 1999). The latter pronunciations are also obtained from Weis, Enz, and Schnorr (2003).

Consider the example in table 5.13. Both words, "archbishop" and "grasshopper", contain the grapheme group "sh". In the example, this group is converted to the single phoneme ʃ (*esh*) in the International Phonetic Alphabet (IPA) format (International Phonetic Association 1999) and SH in the Arpabet (Klatt 1977) for the first word and into two phonemes, sˌh and S HH in the used alphabets, for the second word. Without knowing about the syllable segmentation, both pronunciations, sˌh and ʃ, are possible. If, however, this information is available, it is evident that "sh" in "grasshopper" must not be pronounced as ʃ (*esh*), but as sˌh, i.e. *s* and *h* are mapped to two separate phonemes. Therefore, a learner may benefit from learning how to segment a word into morphemes, syllables, and phonemes when working on G2P conversion as its main task. An overview of all experiment setups is provided in table 5.14.

**Metrics**

The results of this experiment setting are evaluated using the S2S metrics word accuracy and mean edit distance (cf. section 3.6.3).

| Name | Auxiliary Tasks |
|---|---|
| G2P$^{\text{Celex}}$ | – |
| G2P$^{\text{Celex}}_{\text{morph, syll, phon}}$ | Morphemic (morph), syllable (syll), and phonetic (phon) segmentation |
| G2P$^{\text{CMU}}$ | – |
| G2P$^{\text{CMU}}_{\text{morph, syll, phon}}$ | Morphemic (morph), syllable (syll), and phonetic (phon) segmentation |
| G2P$^{\text{Combilex}}$ | – |
| G2P$^{\text{Combilex}}_{\text{morph, syll, phon}}$ | Morphemic (morph), syllable (syll), and phonetic (phon) segmentation |

**Table 5.14.:** Setups for the multi-task G2P conversion experiment using the datasets Celex, CMU, and Combilex. Morphemic (morph), syllable (syll), and phonetic (phon) word segmentation are used as auxiliary tasks. The word segmentation dataset has been described in section 5.1.5.

## 5.6 Experiment Results

We present the results of the previously explained experiments in the following. This section only describes the results while the analysis of them is provided in chapter 6. Note that the prediction post-processing (cf. section 4.3.8) has **only** been used in the AM setups, i.e. the BIO labels in other experiments have not been corrected automatically.

### 5.6.1 Conventions

Tables 5.15 – 5.18 list the evaluation results. To facilitate the comparison of experiment results, the tables adhere to the following conventions.

The results in each table form groups. First, they are ordered by the experiment, then by the evaluated dataset, i.e. the superscript in the setup name, and finally by the percentage of used training samples. The latter is omitted for settings where we have not varied the amount of training data. This grouping forms *blocks* of experiments which are indicated by the horizontal borders in the tables.

Furthermore, results of STL setups are marked in blue as they represent the baseline against which our and Reimers and Gurevych (2017a)'s system compete with MTL. If the results within a line are green, this MTL setup has been able to beat the respective STL baseline. The respective STL baseline for an MTL setup is the STL setup in the same *block*. Note that training data union setups are never colored.

Finally, we highlight the best results in each block in **bold** and underline the best results for an experiment setup across all architectures. We do this only for experiment setups which have been evaluated with the same amount of training data for multiple architectures.

### 5.6.2 Multi-Task Argumentation Mining

Table 5.15 shows the results for setups in the multi-task AM experiment setting. First of all, the scores show that the MTL sequence tagging framework is capable of achieving similar results to the state-of-the-art sequence tagging system by Lample et al. (2016) in the C-F1 metric. In fact, it outperforms the latter in this evaluation criterion. However, the scores for the R-F1 metric are considerably lower suggesting an inferiority of the MTL sequence tagging framework when it comes to modeling relationships between argumentative components in the STL scenario.

| Architecture | Setup | C-F1 (50%) | C-F1 (100%) | R-F1 (50%) | R-F1 (100%) |
|---|---|---|---|---|---|
| MTL sequence tagging framework | $\text{AM}^{\text{PE}}$ | 72.050 % | 63.690 % | 39.705 % | 35.153 % |
| | $\text{AM}^{\text{PE}}_{\text{PCN}}$ | 71.340 % | 63.258 % | 39.825 % | 34.805 % |
| | $\text{AM}^{\text{PE}}_{\text{PDTB}}$ | 71.169 % | 62.739 % | 41.057 % | 35.804 % |
| | $\text{AM}^{\text{PE}}_{\text{PDTB, RST-DT}}$ | 70.864 % | 61.995 % | 39.548 % | 34.462 % |
| | $\text{AM}^{\text{PE}}_{\text{Subtasks}}$ | 74.203 % | 66.620 % | 44.865 % | 39.725 % |
| | $\text{AM}^{\text{PE}}_{\text{Subtasks, PCN}}$ | 72.945 % | 64.685 % | 42.890 % | 37.769 % |
| | $\text{AM}^{\text{PE}}_{\text{Subtasks, PDTB}}$ | 73.456 % | 65.018 % | 44.243 % | 38.604 % |
| Lample et al. (2016) | $\text{AM}^{\text{PE}}$ | 71.075 % | 61.440 % | 42.047 % | 36.444 % |
| Reimers and Gurevych (2017a) | $\text{AM}^{\text{PE}}$ | 70.536 % | 61.727 % | 37.114 % | 31.924 % |
| | $\text{AM}^{\text{PE}}_{\text{PCN}}$ | 65.794 % | 56.680 % | 26.421 % | 22.183 % |
| | $\text{AM}^{\text{PE}}_{\text{PDTB}}$ | 68.013 % | 59.170 % | 34.562 % | 30.118 % |
| | $\text{AM}^{\text{PE}}_{\text{Subtasks}}$ | 69.319 % | 61.284 % | 29.985 % | 25.964 % |
| | $\text{AM}^{\text{PE}}_{\text{Subtasks, PCN}}$ | 74.208 % | 66.468 % | 39.420 % | 34.795 % |
| | $\text{AM}^{\text{PE}}_{\text{Subtasks, PDTB}}$ | 69.743 % | 58.795 % | 37.254 % | 30.452 % |

**Table 5.15.:** Results for multi-task AM. Section 5.6.1 explains the use of color codes and bold markers to highlight results.

By utilizing MTL, the MTL sequence tagging framework is able to achieve substantial improvements compared to the STL baseline and also the architecture by Lample et al. (2016) although not all our MTL setups are beneficial for the learner.

The largest gains in performance due to MTL are achieved when modeling relations. Almost all our MTL setups score higher in this subtask of AM. Improvements of more than 5% have been attained on the 50% level, i.e. the R-F1 (50%) metric.

Using POS tagging, syntactic chunking, and NER (PCN), PDTB alone, and PDTB and RST-DT combined as auxiliary tasks is harmful to the performance w.r.t. modeling components. Nevertheless, PDTB is able to improve the R-F1 metrics, thereby supporting the hypothesis that discourse parsing is beneficial for the learners capability to model relationships (cf. section 5.5.1). However, using PDTB and RST-DT together as auxiliary tasks reduces the performance in all evaluation metrics.

Natural subtasks turn out to be the most successful auxiliary tasks by a wide margin. Combining them with other auxiliary tasks still leads to substantial improvements compared to the STL baseline, but decreases the performance when comparing the scores with using only natural subtasks.

The evaluation results from the architecture of Reimers and Gurevych (2017a) also show that MTL can yield performance improvements, but are quite different to the results of our MTL sequence tagging framework. While the setups *PCN* and *subtasks* alone reduced the performance in all evaluation criteria, combining these auxiliary tasks has yielded the best performance for this architecture. It has even outperformed the best setup in the MTL sequence tagging framework w.r.t. the metric C-F1 (50%).

Utilizing the PDTB dataset with or without subtasks as an auxiliary task is superior to using PCN or subtasks alone for this architecture, but is still worse than the STL baseline. Moreover, PDTB does not improve the R-F1 scores as in our MTL sequence tagging framework.

In fact, Reimers and Gurevych (2017a)'s architecture seems to have difficulties with learning relationships between argumentative components in general. All setups fail to achieve good scores in the R-F1 metrics, i.e. the architecture is not capable of outperforming the STL baselines of the other architectures in modeling relations.

All setups and architectures evaluated in this thesis, however, cannot compete with the state-of-the-art results achieved by Eger, Daxenberger, and Gurevych (2017) with the LSTM-ER architecture by Miwa and Bansal (2016) on the PE dataset. Therefore, the relative differences between the STL and MTL results will be investigated further in the following chapters.

### 5.6.3 Cross-Lingual Epistemic Segmentation

The results for the cross-lingual setting which are listed in table 5.16 indicate that MTL is helpful when using other languages as auxiliary tasks.

| Architecture | Setup | Used Training Samples | F1 | Precision | Recall |
|---|---|---|---|---|---|
| MTL sequence tagging framework | $ES^{SW}$ | 30 % | 23.032 % | 23.495 % | 23.644 % |
| | $ES^{SW+MED}$ | 30 % | 18.416 % | 18.042 % | 21.080 % |
| | $ES^{SW}_{MED}$ | 30 % | 23.211 % | 23.853 % | 24.002 % |
| | $ES^{SW}$ | 100 % | 28.753 % | 28.481 % | 31.230 % |
| | $ES^{SW+MED}$ | 100 % | 23.950 % | 23.695 % | 25.889 % |
| | $ES^{SW}_{MED}$ | 100 % | 28.927 % | 28.961 % | 30.793 % |
| | $ES^{TS}$ | 30 % | 16.904 % | 16.622 % | 19.346 % |
| | $ES^{TS}_{MED}$ | 30 % | 18.553 % | 18.774 % | 21.739 % |
| | $ES^{MED}$ | 30 % | 23.474 % | 24.318 % | 26.270 % |
| | $ES^{MED*}_{TS}$ | 30 % | 24.125 % | 24.696 % | 27.385 % |
| | $ES^{MED*}_{SW}$ | 30 % | 24.119 % | 24.600 % | 25.583 % |
| | $ES^{MED*}_{SW}$ | 100 % | **22.767 %** | **23.399 %** | **24.379 %** |
| Lample et al. (2016) | $ES^{SW}$ | 30 % | 24.614 % | 24.209 % | 28.410 % |
| | $ES^{SW\dagger}$ | 100 % | 29.742 % | 27.727 % | 36.994 % |

**Table 5.16.:** Results for cross-lingual ES with the datasets Medicine (MED), Social Workers (SW), and Teacher Students (TS). Section 5.6.1 explains the use of color codes and bold markers to highlight results.
† The presented scores are only the result of a single run instead of the average of ten runs.

However, the improvements in our MTL setup for the SW dataset are far from substantial, e.g. 0.179% higher F1 when using 30% of the training data. The performance gains due to MTL remain stable in this setup when the amount of training data is increased to 100%. Note that no additional hyper-parameter search has been performed for the setups with 100% of the training data. This bears the risk of an unfair comparison as elaborated for the G2P experiments (cf. section 5.5.4).

Nevertheless, it is evident that MTL is superior to simply using training data from both languages together. Our setups with the union of SW's and MED's training data have considerably lower F1 scores than the respective STL and MTL setups independent of the training dataset's size.

For the TS dataset, utilizing MTL results in a more substantial performance gain. The F1 score is improved by 1.629 %. However, an STL baseline trained on the union of the training data from TS and MED is left for future work. We expect to observe the same trend which has been shown for the SW dataset and the training data union baseline, i.e. the F1 score is lower when training on the training data of both datasets.

When using 30% of the training data, the performance on MED is also improved by MTL. Both datasets, TS and SW, have similar effects on the F1 score of the MED dataset.

Interestingly, the F1 score is smaller after increasing the training set size to 100% in our MTL setup with the SW dataset. Since there is no evaluation with 100% of the training data in an STL setup of the MED dataset, it cannot be determined whether this trend is a peculiarity of this dataset, caused by MTL or due to non-suitable hyper-parameters. This is left for future work, too.

Finally, comparing the results for the SW dataset with the results achieved with the architecture by Lample et al. (2016) shows that the latter architecture is able to achieve notable improvements in the F1 score.

Closer investigation of this improvement shows that it is caused by a substantially higher recall, e.g. a difference of 4.408 % for the setup $ES_{MED}^{SW}$ with 30% of the training data, while the improvements in precision are small. In fact, the precision is even reduced by using Lample et al. (2016)'s architecture for 100% of the training data.

### 5.6.4 Cross-Domain Argument Component Segmentation

Similar to the results of the cross-lingual setting, the results for the cross-domain setting are partially inconclusive. Table 5.17 presents the experiment results. While ACS on the ADB and WTP datasets slightly benefited from MTL (+0.411% and +0.635% F1 respectively), the F1 score dropped for PE:ACS (−0.154%).

| Architecture | Setup | F1 | Precision | Recall |
|---|---|---|---|---|
| MTL sequence tagging framework | $ACS^{ADB}$ | 59.805 % | 59.176 % | **63.099 %** |
| | $ACS^{ADB+PE:ACS}$ | 59.667 % | **60.825 %** | 62.105 % |
| | $ACS^{ADB*}_{PE:ACS}$ | 60.216 % | 60.399 % | 62.018 % |
| | $ACS^{ADB*}_{WTP}$ | 59.450 % | 59.662 % | 62.702 % |
| | $ACS^{PE:ACS}$ | **88.759 %** | **88.022 %** | **89.655 %** |
| | $ACS^{PE:ACS+ADB}$ | 85.316 % | 84.069 % | 86.939 % |
| | $ACS^{PE:ACS}_{ADB}$ | 88.605 % | 88.000 % | 89.347 % |
| | $ACS^{WTP}$ | 46.305 % | 46.094 % | 47.964 % |
| | $ACS^{WTP+ADB}$ | 45.311 % | 45.047 % | 46.457 % |
| | $ACS^{WTP}_{ADB}$ | 46.940 % | 46.906 % | 48.087 % |
| Reimers and Gurevych (2017a) | $ACS^{ADB}$ | 60.328 % | 61.826 % | 62.538 % |
| | $ACS^{ADB+PE:ACS}$ | 59.140 % | 60.791 % | 60.874 % |
| | $ACS^{ADB*}_{PE:ACS}$ | 61.578 % | 62.543 % | 63.673 % |
| | $ACS^{PE:ACS}$ | 86.943 % | 86.015 % | **88.192 %** |
| | $ACS^{PE:ACS+ADB}$ | 63.276 % | 65.840 % | 72.499 % |
| | $ACS^{PE:ACS}_{ADB}$ | 87.246 % | 86.606 % | 88.124 % |

**Table 5.17.:** Results for cross-domain ACS with the datasets AraucariaDB (ADB), Persuasive Essays (PE:ACS), and Wikipedia Talk Pages (WTP). Section 5.6.1 explains the use of color codes and bold markers to highlight results.

With the architecture by Reimers and Gurevych (2017a), more substantial improvements can be seen for the ADB dataset (+1.250%). In addition, the F1 score for the PE:ACS dataset is increased (+0.303%) as well by applying the MTL paradigm.

This experiment, however, shows once more that simply training on the union of the training data is detrimental to the performance. The F1 score for the PE:ACS dataset even drops by more than 23% in the training data union setup with Reimers and Gurevych (2017a)'s architecture.

### 5.6.5 Multi-Task Grapheme-to-Phoneme Conversion

Finally, the results of the fourth experiment are shown in table 5.18. As mentioned in section 3.6.1, the hyper-parameter optimization has been performed on smaller training sets. In the evaluation, we have used larger training sets, too. This allows an observation of MTL success or failure for different amounts of training data.

In this experiment setting, G2P conversion has been evaluated on three different datasets. The results for these datasets vary greatly. In short, MTL is beneficial for only one of them. This is the Combilex dataset and the improvements in word accuracy (+6.296%) and mean edit distance (−0.122) are substantial compared to the STL baseline. On the other side, MTL is harmful to the G2P conversion performance on the datasets Celex (−3.138%) and CMU (−15,432%). Especially, the performance drop on the CMU dataset is considerable.

While similar trends can be observed with the architecture by Reimers and Gurevych (2017a) for the datasets Celex and CMU, i.e. −0.198% and −5.754% word accuracy respectively, the performance is also decreased by MTL for the Combilex dataset with this architecture (−4.013%). Hence, MTL is harmful for all datasets when using Reimers and Gurevych (2017a)'s system. Nevertheless, this architecture performs best for Celex, CMU, and Combilex when comparing the performance on the smallest training datasets.

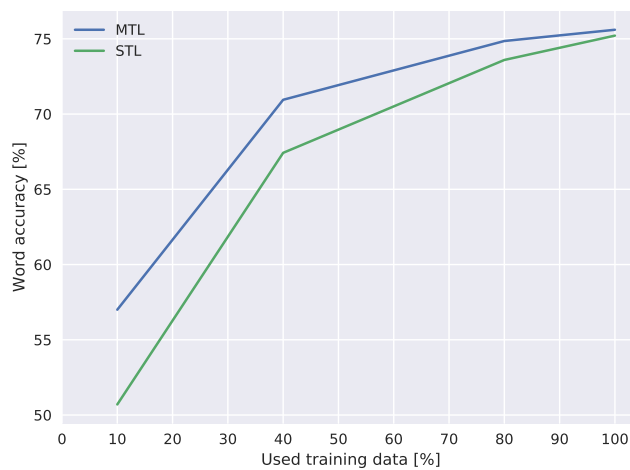

**Figure 5.3.:** Word accuracy for the Combilex dataset with STL and MTL and different training dataset sizes. The training dataset size is given as a fraction (in percent) of all available training samples (cf. table 5.9), i.e. 100% indicates that all training samles are used.

When observing the dependency between the effect of MTL on the performance and the amount of training data with the MTL sequence tagging framework, it is evident that the difference in performance between STL and MTL decreases with an increasing size of the training dataset. This holds for all evaluated datasets independent of the success or failure of MTL. This effect is visualized to for the Combilex dataset in figure 5.3.

| Architecture | Setup | Used Training Samples | Word Accuracy | Mean Edit Distance |
|---|---|---|---|---|
| | G2P$^{Celex}$ | 40 % | **59.240 %** | **0.641** |
| | G2P$^{Celex}_{morph, syll, phon}$ | 40 % | 56.102 % | 0.696 |
| | G2P$^{Celex}$ | 80 % | **66.952 %** | **0.497** |
| | G2P$^{Celex}_{morph, syll, phon}$ | 80 % | 64.263 % | 0.542 |
| | G2P$^{Celex}$ | 100 % | **68.995 %** | **0.459** |
| | G2P$^{Celex}_{morph, syll, phon}$ | 100 % | 66.507 % | 0.500 |
| | G2P$^{CMU}$ | 10 % | **32.745 %** | **1.513** |
| | G2P$^{CMU}_{morph, syll, phon}$ | 10 % | 17.322 % | 2.210 |
| MTL sequence | G2P$^{CMU}$ | 40 % | **47.871 %** | **1.079** |
| tagging framework | G2P$^{CMU}_{morph, syll, phon}$ | 40 % | 32.727 % | 1.480 |
| | G2P$^{Combilex}$ | 10 % | 50.703 % | 0.807 |
| | G2P$^{Combilex}_{morph, syll, phon}$ | 10 % | **56.999 %** | **0.685** |
| | G2P$^{Combilex}$ | 40 % | 67.424 % | 0.482 |
| | G2P$^{Combilex}_{morph, syll, phon}$ | 40 % | **70.949 %** | **0.423** |
| | G2P$^{Combilex}$ | 80 % | 73.596 % | 0.376 |
| | G2P$^{Combilex}_{morph, syll, phon}$ | 80 % | **74.856 %** | **0.355** |
| | G2P$^{Combilex}$ | 100 % | 75.214 % | 0.346 |
| | G2P$^{Combilex}_{morph, syll, phon}$ | 100 % | **75.604 %** | **0.340** |
| | G2P$^{Celex}$ | 40 % | **62.429 %** | **0.589** |
| Lample et al. (2016) | G2P$^{CMU}$ | 10 % | **31.837 %** | **1.569** |
| | G2P$^{Combilex}$ | 10 % | **54.412 %** | **0.730** |
| | G2P$^{Celex}$ | 40 % | 65.415 % | 0.532 |
| | G2P$^{Celex}_{morph, syll, phon}$ | 40 % | 65.217 % | 0.529 |
| Reimers and | G2P$^{CMU}$ | 10 % | **34.411 %** | **1.458** |
| Gurevych (2017a) | G2P$^{CMU}_{morph, syll, phon}$ | 10 % | 28.657 % | 1.684 |
| | G2P$^{Combilex}$ | 10 % | **61.065 %** | **0.609** |
| | G2P$^{Combilex}_{morph, syll, phon}$ | 10 % | 57.052 % | 0.667 |

**Table 5.18.:** Results for multi-task G2P conversion. Section 5.6.1 explains the use of color codes and bold markers to highlight results.

# 6 Discussion

## 6.1 Architecture Comparison

The experiments presented in chapter 5 have been performed with three different architectures (cf. section 5.4). While experiment-specific differences in the evaluation results have been pointed out in section 5.6, this section compares the architectures across all experiments. Furthermore, a runtime comparison is presented in section 6.1.2.

### 6.1.1 Evaluation Performance

First of all, there is no architecture that consistently outperforms any of the other architectures. In the first experiment, our MTL sequence tagging framework achieves the best results by utilizing the MTL paradigm. Lample et al. (2016)'s architecture achieves the best results in the ES experiments even compared to our MTL setups. While the multi-lingual ACS experiments suggest a tie between our MTL sequence tagging framework and the architecture by Reimers and Gurevych (2017a), the latter system outperforms the others in the G2P conversion experiments with substantially better results.

| Architecture | Setting | STL better | MTL better | Count |
|---|---|---|---|---|
| MTL sequence tagging framework | AM | 3 | 3 | 6 |
| | ES | 0 | 5 | 5 |
| | ACS | 2 | 2 | 4 |
| | G2P | 5 | 4 | 9 |
| | Count | 10 | 14 | 24 |
| Reimers and Gurevych (2017a) | AM | 4 | 1 | 5 |
| | ACS | 0 | 2 | 2 |
| | G2P | 3 | 0 | 3 |
| | Count | 7 | 3 | 10 |
| **Overall Count** | | 17 | 17 | 34 |

**Table 6.1.:** Overview of the evaluation w.r.t. MTL results. This table summarizes how many MTL setups have beaten the STL baseline. Only if an MTL setup is better than the STL baseline in the majority of all applicable metrics, we consider the MTL setup as having outperformed STL.

With our MTL sequence tagging framework, applying the MTL paradigm frequently results in better predictions w.r.t. the applied metrics. In contrast, MTL with Reimers and Gurevych (2017a)'s architecture often caused substantially worse predictions than STL (cf. table 5.15 and 5.18). This is summarized in table 6.1. Therefore, trends observed with one architecture cannot be seen with the other and vice versa although both architectures are very similar to each other (cf. section 4.1).

In the multi-task `AM` experiment settings, the MTL sequence tagging framework and the system by Reimers and Gurevych (2017a) were able to outperform the strong STL baseline set by Lample et al. (2016)'s architecture through utilizing MTL. This result is interesting because on the one hand it shows that even an inferior base architecture, i.e. inferior w.r.t. the STL performance, can outperform a seemingly superior architecture when applying MTL. On the other hand, the importance of implementation details becomes evident since the three systems (MTL sequence tagging framework, Reimers and Gurevych (2017a), and Lample et al. (2016)) all use very similar network architectures. That is, all of them have LSTM layers, utilize character-level information, and use a CRF layer for prediction.

Improvements in the actual implementation may thus result in a better STL performance. Moreover, the performance in MTL setups may be increased as well. A comparison of implementation details is left for future work. This comparison may even result in a framework comparison because the systems use different frameworks, i.e. TensorFlow (MTL sequence tagging framework), Keras[22] (Reimers and Gurevych 2017a), and Theano[23] (Lample et al. 2016). Moreover, it may yield a better understanding of the somewhat divergent trends regarding the use of MTL when comparing Reimers and Gurevych (2017a)'s and our architecture.

### 6.1.2 Runtime

To compare the runtime characteristics of the different architectures, an STL and an MTL experiment have been performed on the same machine[24]. In addition to the previously presented architectures (cf. section 5.4), the runtime comparison also includes the architecture by Søgaard and Goldberg (2016). It has been excluded from the experiments in chapter 5 because its model is less powerful, e.g. no CRF layer, and therefore was not able to achieve competitive results on the experiment tasks.

The STL experiment is $\text{AM}^{\text{PE}}$ and the MTL experiment is $\text{AM}^{\text{PE}}_{\text{ACS}}$, i.e. Argument Component Segmentation (ACS) is used as the sole auxiliary task for AM on the Persuasive Essays (PE) dataset. We choose the AM setting as this thesis is focused on MTL for AM and perform both experiments on 20% of PE's training data (cf. table 5.5). Each training procedure has at most 50 epochs, but may terminate early due to early stopping. Table 6.2 provides the runtime results.

| System | $\text{AM}^{\text{PE}}$ (STL) | | | $\text{AM}^{\text{PE}}_{\text{ACS}}$ (MTL) | | |
|---|---|---|---|---|---|---|
| | **Training Epochs** | **Training Duration** | **Epoch Duration** | **Training Epochs** | **Training Duration** | **Epoch Duration** |
| MTL sequence tagging framework | 28 | 2385 s | 85.179 s | 25 | 15 847 s | 633.880 s |
| Lample et al. (2016) | 14 | 5736 s | 409.714 s | – | – | – |
| Reimers and Gurevych (2017a) | 50 | 3235 s | 64.700 s | 12 | 3079 s | 256.583 s |
| Søgaard and Goldberg (2016) | 50 | 9174 s | 183.480 s | 50 | 7784 s | 155.680 s |

**Table 6.2.:** Runtime comparison between the MTL sequence tagging framework and the architectures by Lample et al. (2016), Reimers and Gurevych (2017a), and Søgaard and Goldberg (2016). The experiments are AM on the PE dataset. The MTL setup uses the natural subtask ACS as the only auxiliary task.

Since the number of epochs varies, comparing the total training duration is not sensible. However, the epoch duration can be compared. In the STL setup the architecture by Reimers and Gurevych (2017a) is the fastest. Despite its simple architecture, Søgaard and Goldberg (2016)'s architecture is comparably slow. Only the architecture by Lample et al. (2016) requires substantially more time per epoch.

---

[22] https://keras.io/
[23] http://deeplearning.net/software/theano/
[24] AMD Phenom<sup>TM</sup> II X4 965 processor (4 cores) and 16 GB RAM (DDR3).

Interestingly, the system by Søgaard and Goldberg (2016) requires less time per epoch in the MTL setup while the runtime is substantially increased for the other architectures that are capable of using MTL. It is also the overall fastest system in the MTL setup. While the runtime difference of our MTL sequence tagging framework and Reimers and Gurevych (2017a)'s system is small in the STL setup, it is substantial for MTL. The reason may be that the latter system is "optimized for performance"[25]. Since both systems are similar w.r.t. their features, the suggested implementation comparison (cf. section 6.1.1) could provide clues to increase the training speed of our MTL sequence tagging framework.

## 6.2 Analysis of Evaluation Results

The previous chapter has only described the evaluation results. This section analyzes these results, thereby investigating the reasons for a good or bad prediction performance of different setups, highlighting implications for future work regarding the MTL paradigm, and providing insights on the learning procedure of the learners in different settings and setups.

### 6.2.1 Sensitivity to Random Seeds

The non-deterministic nature of neural networks (cf. section 5.3) causes the results on the test set to vary across the ten evaluation runs. Reimers and Gurevych (2017b) have investigated this phenomenon extensively. In this section, a similar investigation is applied to our experiments. In particular, the sensitivity to different random seeds and thus different weight initializations of the used architectures is analyzed.

First of all, the dispersion is analyzed on the architecture and experiment level. The experiments use different metrics, e.g. C-F1 (50%), word accuracy, and F1. Hence, it is not optimal to compare their variance or standard deviation. Instead, the *coefficient of variation* cv is used to compare the results for different experiments. This coefficient is calculated by dividing the standard deviation $\sigma$ by the mean $\mu$, i.e. $\text{cv} = \frac{\sigma}{\mu}$. This enables a dimensionless comparison (Brown 1998).

In figure 6.1a, the variation for each architecture is visualized and figure 6.1b shows the variation for each experiment setting. Each point in the visualizations represents a single experiment setup. The number of setups evaluated with each architecture is very imbalanced. Thus, it is difficult to find substantial differences between the different architectures w.r.t. dispersion. However, Lample et al. (2016)'s architecture appears to have the smallest degree of dispersion. The other architectures are similar to each other and both have setups with a very high degree of dispersion.

In contrast, the visualization allows to draw clear conclusions regarding differences between the experiment settings in their sensitivity to different weight initializations. Multi-task `AM` and multi-task `G2P` conversion setups are less sensitive while setups in the cross-lingual `ES` setting have a higher degree of dispersion. Therefore, good results are rather achieved by chance than by a suitable model. This lack of resilience to varying weight initializations indicates that the evaluated systems are incapable of modeling the `ES` task properly. The low F1 scores (cf. table 5.16) indicate this, too. In the cross-domain `ACS` setting, Reimers and Gurevych (2017b)'s architecture has results with less variant results compared to the MTL sequence tagging framework.

After analyzing the sensitivity to random seeds on a higher level, we now investigate some of the experiment settings in detail. Since the experiment setups within the same experiment setting are evaluated with the same metrics, the coefficient of variation is not required. The results, thus, are compared directly using violin plots. A violin plot is created by estimating the probability density function from

---

[25] `https://github.com/UKPLab/emnlp2017-bilstm-cnn-crf`

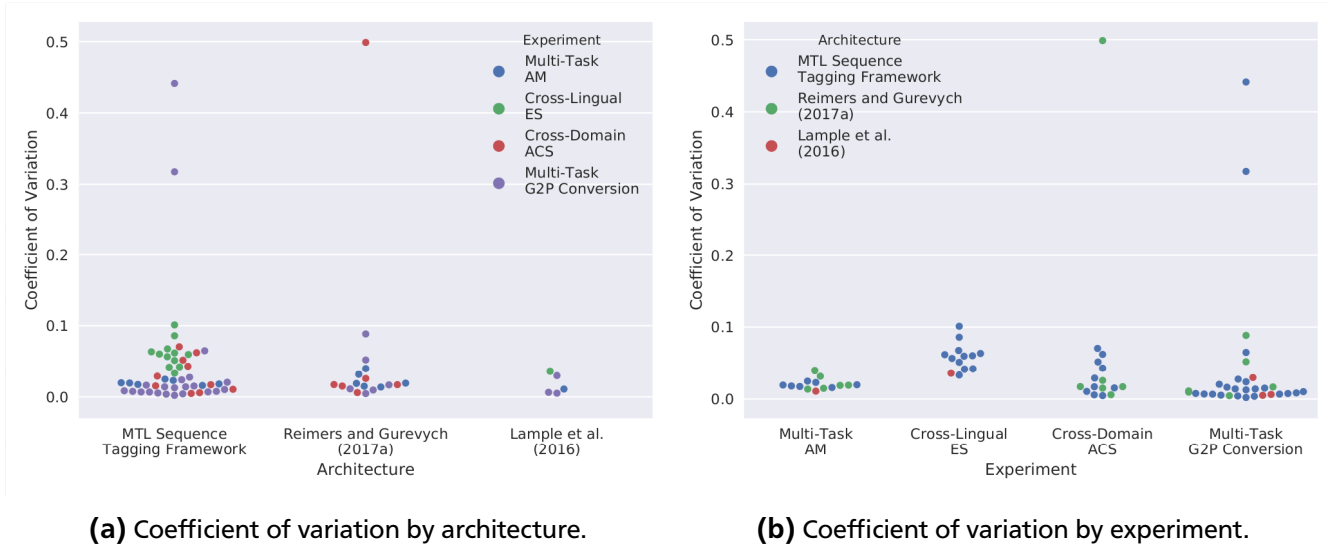

**(a)** Coefficient of variation by architecture.

**(b)** Coefficient of variation by experiment.

**Figure 6.1.:** Coefficient of variation for the experiment results. Each point represents an experiment setup. Each of them is run ten times and the coefficient of variation is calculated across these runs. The coefficient has been calculated for each setup only for a single metric, i.e. C-F1 (50%) for `AM`, F1 for `ES` and `ACS`, and word accuracy for `G2P` conversion.

the given samples and plotting it to the y-axis (Reimers and Gurevych 2017a). These samples are the evaluation results for each setup. Hence, each violin is created from ten samples.

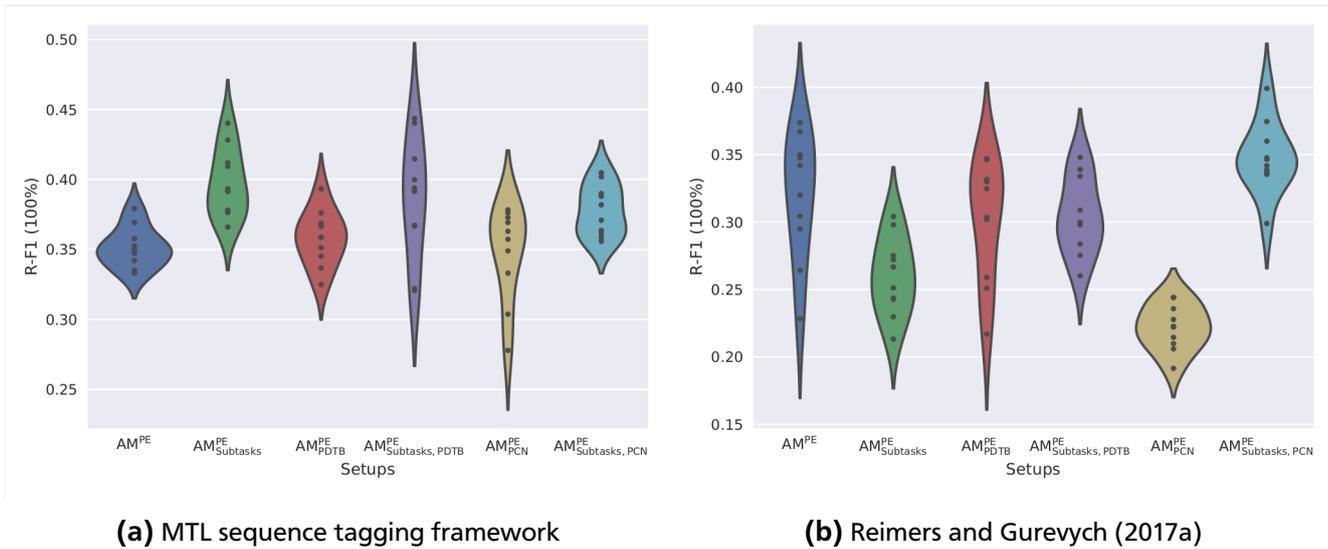

**(a)** MTL sequence tagging framework

**(b)** Reimers and Gurevych (2017a)

**Figure 6.2.:** Violin plots for multi-task `AM` with our MTL sequence tagging framework and the system by Reimers and Gurevych (2017a) using the metric R-F1 (100%). $\text{AM}^{\text{PE}}_{\text{PDTB, RST-DT}}$ is excluded from this visualization.

Figure 6.2 shows violin plots for the setups of the multi-task `AM` experiment setting. Section 5.6.2 has already provided a description of the visualized results. In contrast, this analysis focuses on the variance in the results. The STL results for our MTL sequence tagging framework have less variance than with the system by Reimers and Gurevych (2017a). The reverse is true for the results of the setup $\text{AM}^{\text{PE}}_{\text{PCN}}$. In general, however, the latter architecture's results have a higher variance and more extreme outliers which are often below the mean, i.e. decrease the average score. This explains why this architecture struggles to outperform our MTL sequence tagging framework (cf. table 5.15) in the first experiment setting.

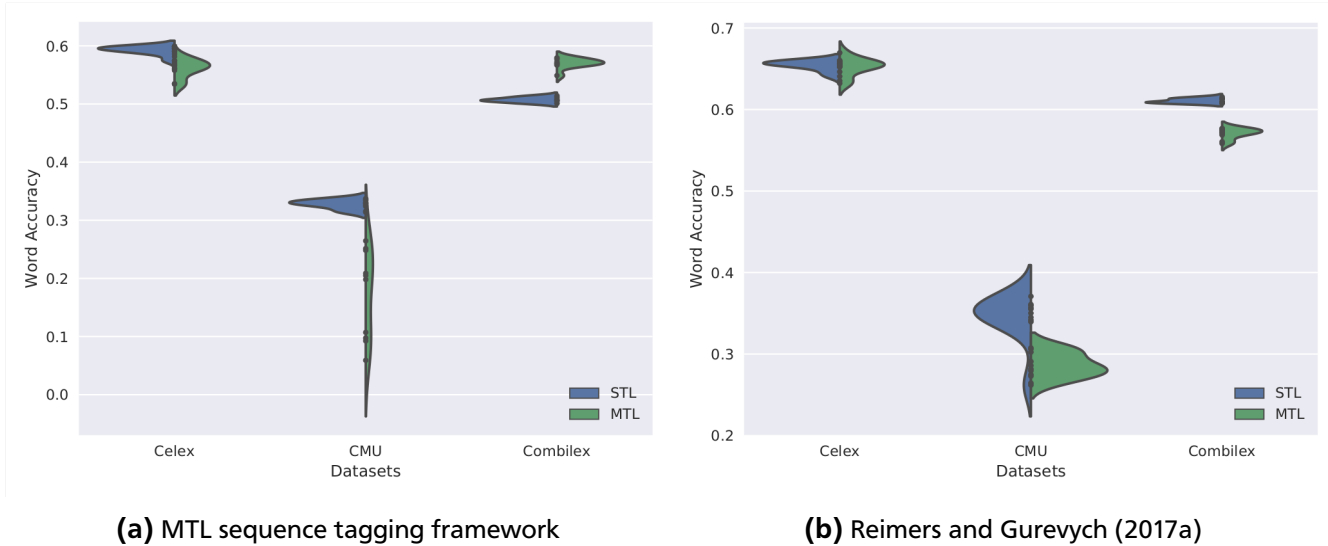

**(a)** MTL sequence tagging framework

**(b)** Reimers and Gurevych (2017a)

**Figure 6.3.:** Violin plots for multi-task G2P conversions with our system and the system by Reimers and Gurevych (2017a).

Furthermore, it can be seen that the best setups, i.e. $\text{AM}^{\text{PE}}_{\text{Subtasks}}$ and $\text{AM}^{\text{PE}}_{\text{Subtasks, PCN}}$ for our and Reimers and Gurevych (2017a)'s architecture respectively, are characterized by a comparably small degree of dispersion. This indicates that their robustness to different weight initializations is a reason for their superiority to other setups.

The violin plots of the G2P experiments in figure 6.3 show that the models are very stable w.r.t. changing weight initializations for the Celex and Combilex datasets. In contrast, the variance of the results for the CMU dataset is very high. In particular, our CMU MTL setup is characterized by a high degree of dispersion which explains why the reported average word accuracy is substantially lower compared to the STL setup. Overall, MTL is more stable when faced with varying random initializations than STL in this experiment setting.

When comparing the violin plots for Celex and Combilex, it becomes evident that they have very similar patterns in both architectures despite different means. This indicates that both architectures have produced models which are alike in terms of resilience to different weight initializations. With both datasets variance of the evaluation results is not the cause for better or worse scores with the MTL setup.

### 6.2.2 Sensitivity to Hyper-Parameters

After analyzing the sensitivity to random seeds in the last section, the sensitivity to varying hyper-parameters is investigated. We do this by visualizing the dispersion of the evaluation results in the hyper-parameter search. Figure 6.4 visualizes the coefficient of variation by architecture (6.4a), experiment setting (6.4b), and paradigm (6.4c).

First of all, the models created by our MTL sequence tagging framework are more sensible to changes in the hyper-parameter configuration, i.e. good results are more dependent on choosing the correct hyper-parameters than they are with the other architectures.

The highest degree of dispersion can, again, be found in the second experiment setting. Not only is there an overall higher variation in the results, extreme outliers are more common than in any of the other experiment settings, too. This is a further indicator for the difficulty of the ES task. On the one hand, the best models' results are susceptible to varying weight initializations (cf. section 6.2.1) and on the other hand, it is difficult to find a good model in the first place because of the strong influence of the hyper-parameters.

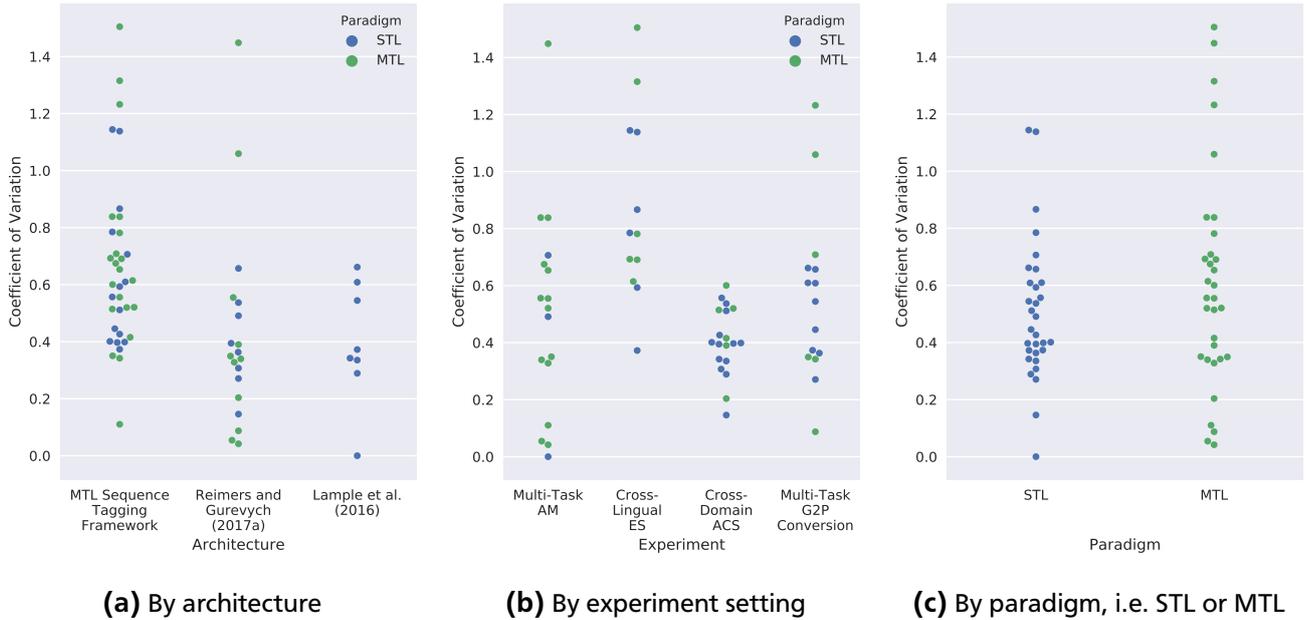

**(a)** By architecture  **(b)** By experiment setting  **(c)** By paradigm, i.e. STL or MTL

**Figure 6.4.:** Coefficient of variation for the results in the hyper-parameter optimization. Each point represents an experiment setup which has ten trials. The coefficient is calculated across these runs and is only calculated for a single metric per setup, i.e. C-F1 (50%) for `AM`, F1 for `ES` and `ACS`, and word accuracy for `G2P` conversion.

Finally, figure 6.4c shows that MTL setups are more sensitive to different hyper-parameter configurations. This is independent of the architecture and experiment setting (cf. figures 6.4a and 6.4b). This insight, however, is not unexpected since MTL setups have more hyper-parameters, e.g. which task is terminated on which shared layer, and thus a larger parameter space with potentially more *important* hyper-parameters, i.e. hyper-parameters with a substantial influence on the performance. Moreover, different task have different degrees of sensitivity to hyper-parameters (cf. figure 6.4b). Due to the parameter sharing in MTL, combining multiple tasks instead of learning a single task is more likely to have an increased sensitivity to changes in the hyper-parameters as one of many auxiliary tasks, for instance, may be highly sensitive which influences the sensitivity of the complete model in the MTL setup.

### 6.2.3 Learning Procedures

Figure 6.5 shows the training progress for two experiment settings by visualizing the performance on the development dataset calculated with a task-specific metric in each training epoch. In the following, we use *learning procedure* and *learning curve* interchangeably. While the STL setups have curves with a similarly steep or even steeper slope in the beginning, MTL is more successful in the long run. In the STL setups, either the slope becomes less steep faster than the slope of the MTL curve (figure 6.5a) or the learner gets stuck in a local optimum (figure 6.5b)

Taking a closer look at figure 6.5b, one can see that the STL curve for $ES^{MED}$ continues to grow while the STL curve for $ES^{SW}$ hits a plateau. This indicates that using MED as an auxiliary task helps to overcome the local optimum.

However, MTL has a substantial impact on the runtime of the system. For our MTL sequence tagging framework, the runtime is increased by almost 100% on average when MTL is utilized. As shown in table 6.2, the runtime is also increased for Reimers and Gurevych (2017a)'s architecture.

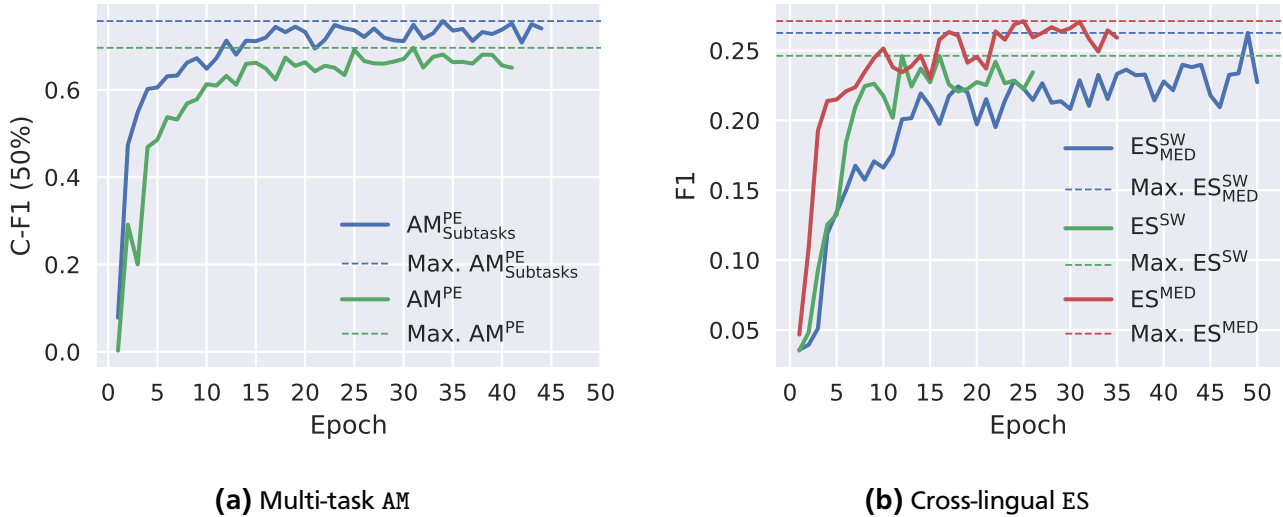

**(a)** Multi-task AM

**(b)** Cross-lingual ES

**Figure 6.5.:** Training progress for multi-task AM and cross-lingual ES on the respective development datasets.

### 6.2.4 Suitability of Auxiliary Tasks

As explained in section 3.3.5 assessing the suitability of auxiliary tasks is still an open research question. Based on our evaluation results, we provide the findings from related work with further empirical evidence in this section.

**Entropy and Kurtosis**

Analyzing the auxiliary tasks by means of dataset properties is particularly feasible for the experiment settings cross-lingual ES and cross-domain ACS because in each MTL setup only a single auxiliary task has been used which allows a direct comparison of the auxiliary tasks. Moreover, the label entropy of the auxiliary tasks can be compared among each other for either setting because they have (almost) the same number of labels.

First, performing ES on the MED dataset has been learned in isolation (STL), together with SW, and together with TS. As documented in table 5.16, the latter setup achieves the best performance on the MED dataset. While both datasets TS and SW have a similar entropy, TS' kurtosis is substantially smaller. In the cross-domain ACS setting, we have made a similar observation. The performance of ACS on the ADB dataset is best when the learner trains ACS on ADB and PE:ACS jointly via MTL. Actually, training ACS with MTL on ADB and WTP hurts the performance on ADB. When revisiting the properties of these datasets from table 5.7, one sees that the label distribution of PE:ACS has higher entropy and lower kurtosis than the label distribution of WTP.

These findings suggest that entropy and kurtosis are useful for predicting a suitable auxiliary task in advance. Indeed, the findings are consistent with Alonso and Plank (2017) and Bingel and Søgaard (2017).

**Learning Curves**

Bingel and Søgaard (2017) conclude that the STL learning curves of auxiliary tasks are predictive of the MTL performance (cf. section 3.3.5). The analysis of the learning procedure of $ES^{SW}_{MED}$ in the previous section supports this finding as well. Further analysis of the learning curves is, however, left for future work because more experiments in which the auxiliary tasks are trained in isolation are necessary.

**Size of the Label Inventory**

Besides entropy and kurtosis, the size of the label inventory is an interesting dataset property according to Alonso and Plank (2017). They conclude that smaller label inventories are preferable for auxiliary tasks. The discourse parsing datasets PDTB and RST-DT have significant differences in this category with 27 and 4398 unique labels respectively (cf. table 5.4). While PDTB alone has been a useful auxiliary task to improve the learners relation modeling capability, utilizing PDTB and RST-DT together has resulted in the worst `AM` performance of the MTL sequence tagging framework. RST-DT's huge label inventory is a potential cause for this performance deterioration. Performing an experiment with RST-DT as the sole auxiliary task is, however, necessary to legitimate this conclusion, but left for future work.

**Natural Subtasks**

Finally, natural subtasks have proven to be effective auxiliary tasks (cf. table 5.15). In our multi-task `AM` setting, they are the best auxiliary tasks for the MTL sequence tagging framework and are also better than PCN and PDTB for the system by Reimers and Gurevych (2017a). In the latter system, using natural subtasks together with PCN is the most successful approach. The success of natural subtasks is consistent with the related work presented in section 2.2.5. The reasons for their success have been introduced in section 3.3.5. Their label inventories are smaller by design and they operate on the same input features as the main task. In case of the PE dataset, the natural subtasks all have considerably lower kurtosis than the main task, too. Furthermore, natural subtasks address label sparsity as they effectively reduce the number of infrequent labels.

### 6.2.5 Network Architecture and Features

Due to the flexibility of the MTL sequence tagging framework, the actual network architecture is highly dependent on the used configuration file which is controlled to a large extent by the values of the random search trials. This section discusses whether there are features and architectural decisions, e.g. task hierarchies, which have been chosen to be included in the best configurations by the hyper-parameter optimization more often. In total, the hyper-parameter search has resulted in 27 configurations by selecting the best configurations out of 270 trial configurations. This investigation is limited to configurations of the MTL sequence tagging framework because its configuration options are the most comprehensive of all evaluated architectures.

**Task Hierarchies**

First of all, we observe a tendency towards hierarchical models as introduced by Søgaard and Goldberg (2016). It is, however, important to note that the configuration of the hyper-parameter search made hierarchical setups more likely to ensure that the main task is terminated last, i.e. it is more likely that the main task is terminated on a higher layer than the auxiliary tasks. We have ensured this by selecting the index of the main task's termination layer from a range of higher values compared to the auxiliary tasks.

Nevertheless, it can be seen that the auxiliary tasks form a hierarchy among each other. In the first experiment setting, POS tagging is always predicted from a lower layer than syntactic chunking and NER. Morphemic segmentation is always terminated on the fourth shared layer while phonetic and syllable segmentation are predicted from lower layers in the multi-task G2P conversion setting. Moreover, in two out of three of our MTL setups, the phonetic segmentation task is terminated before the syllable segmentation. In the second and third experiment settings, main and auxiliary task are always terminated on the same shared layer by our configuration.

**Features**

In the investigated configuration files, preferences for specific features can be detected. Variational dropout is used seldom, i.e. only in 5 out of 27 configurations. Approximately half of the configurations use shortcut connections, but the uses are distributed unevenly. While this feature is used heavily in the cross-lingual ES and multi-lingual G2P conversion settings, it is almost completely ignored in the multi-task AM setups. We hypothesize that while the shared layers mostly capture language-agnostic features, the shortcut connections allow to maintain language-specific information in the cross-lingual setting as the word embeddings for German and English words are slightly different even though we use bilingual embeddings. Moreover, we believe the reason for the shortcut connections' success in the G2P setting is that the dependency between an input character and its phonemes is stronger than the dependency between tokens and their labels in the other sequence tagging experiments. In G2P, a character can only be mapped to a small set of possible phonemes, e.g. "a" may never be mapped to ʃ (*esh*). In contrast, most tokens can mapped to any AM, ES or ACS label in the other experiments. Therefore, maintaining the character embedding in the input of the shared layers is more helpful than maintaining word embeddings and thus shortcut connections are more useful.

Approximately every second configuration uses character-level information, but they are used rarely in the first experiment setting. In contrast, all setups of the cross-lingual ES setting use character-level information. Since the German datasets TS and SW contain substantially more out of vocabulary words compared to the other used datasets, e.g. 2.440 % of MED's training data tokens are OOV while the OOV rate for SW is 8.770 %, character-level information are especially useful (cf. section 4.3.2) even though their use is counter-intuitive in a cross-lingual experiment as characters are used differently in different languages.

As expected given related work (e.g. Reimers and Gurevych 2017a), the majority of the configurations (20 out of 27) has a CRF classifier for the main task. This preference exists for the auxiliary tasks as well.

### 6.2.6 Error Analysis

We analyze the results of the experiments in detail in this section. In particular, causes for performance improvements and deteriorations shall be revealed. However, not all potential sources of errors or success can be analyzed within the scope of this thesis. The analysis is therefore not exhaustive. While the analysis covers some results related to the architectures by Lample et al. (2016) and Reimers and Gurevych (2017a), we focus primarily on the results of our MTL sequence tagging framework.

#### Multi-Task Argumentation Mining

Considerable performance improvements due to MTL have been achieved in the multi-task AM setting. Figure 6.6 shows that these improvements are caused by being more successful when predicting longer argumentative components. In particular, the improved performance on components of length 20 to 35 helps to improve the overall performance since these components occur substantially more frequently than extremely long components, i.e. components with a length of more than 35 tokens. The worst MTL setup, i.e. $AM^{PE}_{PDTB,\ RST-DT}$, on the other hand, has the most severe performance deterioration with extremely long components which, however, occur infrequently. Potential causes for the reduced performance of this setup compared to STL have already been discussed in section 6.2.4.

As mentioned in section 5.6.2, the setup $AM^{PE}_{PDTB}$ successfully improved the capability of the learner to model argument relations as intended. Unfortunately, it reduced the performance in the C-F1 evaluation metrics. PDTB has only relation labels after the performed preprocessing (cf. section 5.1.1). Therefore, we hypothesize that learner is too focused on modeling relations and fails to segment and classify components correctly. Given the inherent dependency between components and relationships in the AM metrics, i.e. predicting a correct relation requires predicting its components correctly (Eger, Daxenberger, and Gurevych 2017), the improvement achieved by using PDTB becomes more surprising. Despite fewer correctly predicted components and thus, a performance penalty for the relation prediction performance, improvements in the relation modeling capability have been attained. This implicates that the discourse parsing auxiliary task has enabled a more efficient use of the correctly predicted components compared to the STL setup.

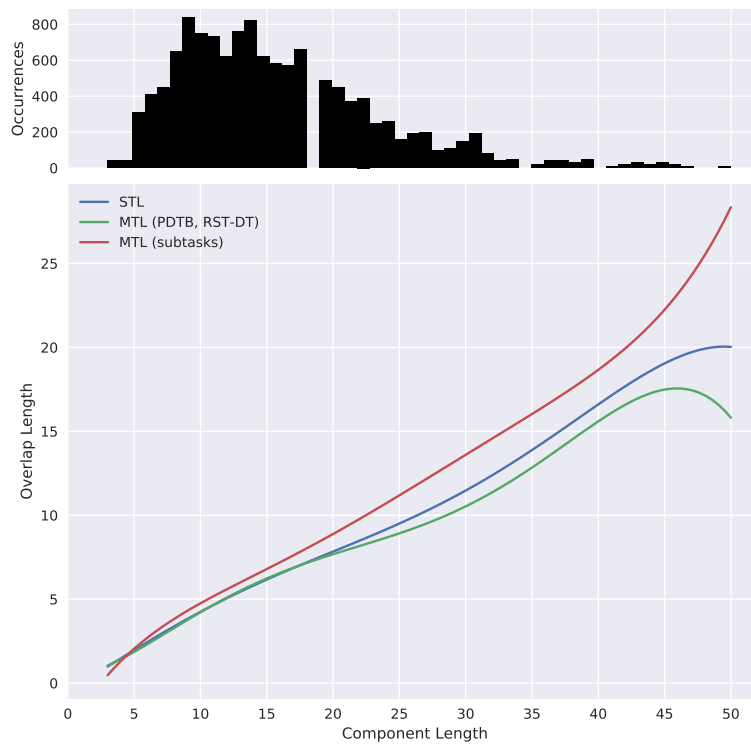

**Figure 6.6.:** Correlation between component length and correctness of predictions for the PE dataset with the MTL sequence tagging framework. Each line in the visualization represents an experiment setup. To prevent overplotting, only the STL setup and the best and worst MTL setups are visualized. Higher values on the $y$-axis refer to more correct predictions for argumentative components of the length $x$. The correctness of a prediction primarily depends on assigning the correct type to a component, but also on selecting the correct span of text, i.e. the correctness is not measured on the token level, but rather on the component level. We provide a detailed explanation of this figure's creation process in appendix B.

### Cross-Lingual Epistemic Segmentation

In general, the performance on the ES datasets is not good. The datasets, however, are comparably small (cf. table 5.6). Larger datasets may yield better results. Moreover, the results are at least substantially better than the majority baseline. For instance, the most frequent label in the test set of the SW dataset is `I-EG` which is used for 2636 out of 7437 tokens, i.e. 35.444 %. Predicting only this label results in 3.079 % F1 score. Using a simple random baseline system which randomly assigns one of the 17 labels of the SW dataset to a token, achieves 6.027 % F1 score. The probability distribution used by this system is equivalent to the label distribution in the test dataset, e.g. the label `I-EG` is assigned to a label with a probability of 35.444 %.

An analysis of the predictions reveals that the learner fails to adhere to the BIO labeling. For the SW dataset, more than 40 % of all components are inconsistent according to the BIO scheme (cf. section 3.4). Moreover, the component spans are frequently incorrect. That's why applying the automated BIO error correction (cf. section 4.3.8) does not result in better F1 scores either. Finally, a high number of incorrect classifications, i.e. low accuracy, can be observed as well. The achieved accuracy of approximately 36 % is

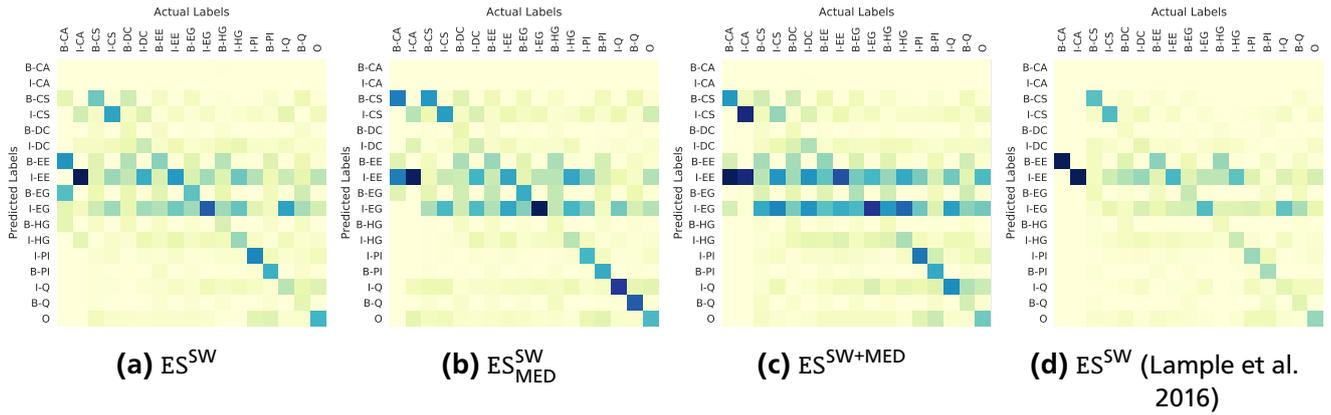

(a) ES$^{SW}$  (b) ES$^{SW}_{MED}$  (c) ES$^{SW+MED}$  (d) ES$^{SW}$ (Lample et al. 2016)

**Figure 6.7.:** Confusion matrices for the SW dataset. Darker squares indicate that the predicted label occurs more frequently. The matrices have been normalized by dividing the number of prediction occurrences by the number of label occurrences in the gold standard data, i.e. each value is divided by the sum of its column's values. Figures 6.7a - 6.7c summarize the predictions of the ten evaluations on test with the MTL sequence tagging framework. Figure 6.7d shows the confusion matrix for the best evaluation with the system by Lample et al. (2016).

only slightly above the majority baseline. These observations hold true for predictions of both evaluated architectures, i.e. the MTL sequence tagging framework and Lample et al. (2016)'s architecture.

The most frequent labels in the test split of the SW dataset are `I-EG` (35.444 %), `I-EE` (19.040 %), and `I-DC` (9.735 %). In all setups, the false positive rate for `I-EG` and `I-EE` is high (see figure 6.7). However, the training data union setup ES$^{SW+MED}$ suffers the most from this misclassification. `I-DC`, on the other side, has only a few false positives, but does not have many true positives either. Since the systems fail to adhere to the BIO labeling, the infrequent `B-` labels are often classified incorrectly.

Figure 6.8 shows that MTL, again, achieves prediction improvements for longer components, but the difference between STL and MTL is not as clear as in the multi-task `AM` setting. However, predicting a long component correctly in this setting has a more substantial impact on the evaluation metrics because the F1 score, precision, and recall are calculated on the token level. In contrast, C-F1 (100%), for example, is calculated on the component level. Therefore, MTL is still able to achieve performance improvements compared to STL setups (cf. table 5.16), but they are not as substantial as in the first experiment setting.

### Cross-Domain Argument Component Segmentation

In this experiment setting, MTL has been most successful for ACS on the ADB dataset. When revisiting table 5.7, the main difference to the other ACS datasets is apparent: the ADB dataset is significantly smaller w.r.t. the number of tokens. As other authors concluded in related work (cf. section 2.2.1) and the analysis of different training dataset sizes in section 5.6.5 showed, MTL is particularly useful for smaller datasets. The results in this experiment setting, therefore, are further indicators for this characteristic of MTL.

Figure 6.9 shows that MTL improves the performance on longer components in the cross-domain ACS setting, too. However, the improvements are even more marginal than in the cross-lingual ES setting. Visualized in figure 6.9a, one can see that only the ACS$^{ADB*}_{PE:ACS}$ setup achieves better performance. Moreover, the training data union setup ACS$^{ADB+PE:ACS}$ is better than the ACS$^{ADB*}_{WTP}$ setup. These findings are consistent with the experiment results presented in table 5.17. Similarly, the lack of performance differences for longer components on the PE:ACS dataset displayed in figure 6.9b is consistent with the lack of performance improvements due to MTL in the experiment results.

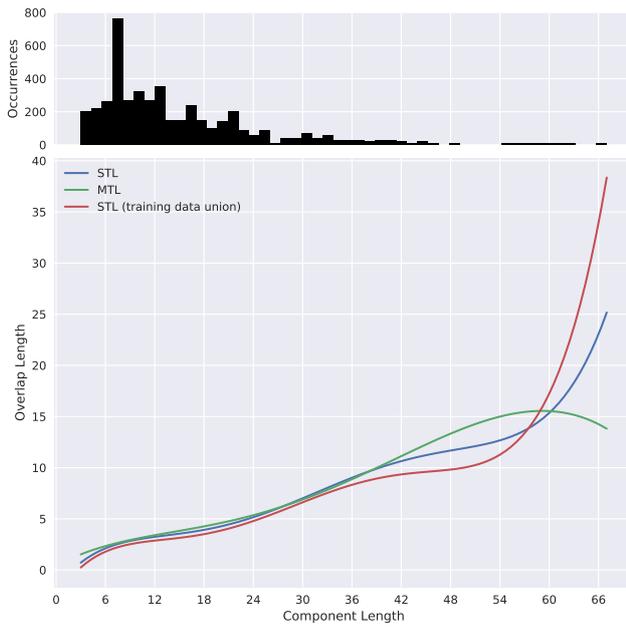
**(a)** Social Workers (SW)

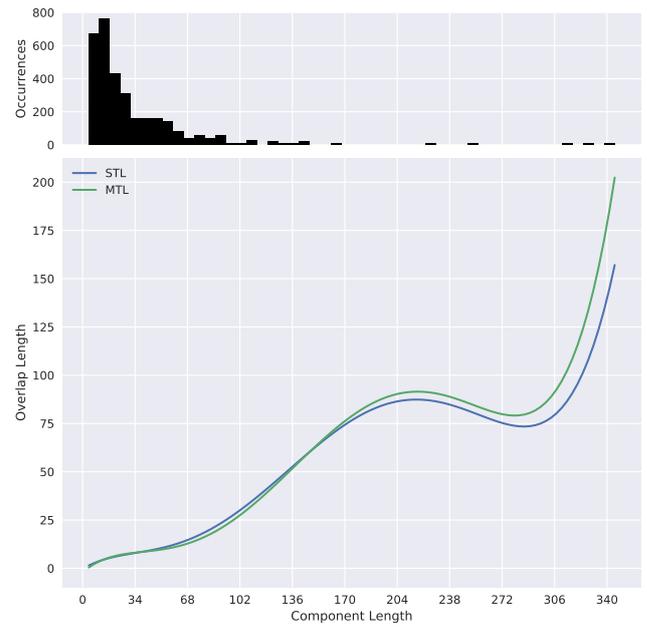
**(b)** Teacher Students (TS)

**Figure 6.8.:** Correlation between component length and correctness of predictions for the SW and TS datasets analogous to figure 6.6. We provide a detailed explanation of this figure's creation process in appendix B.

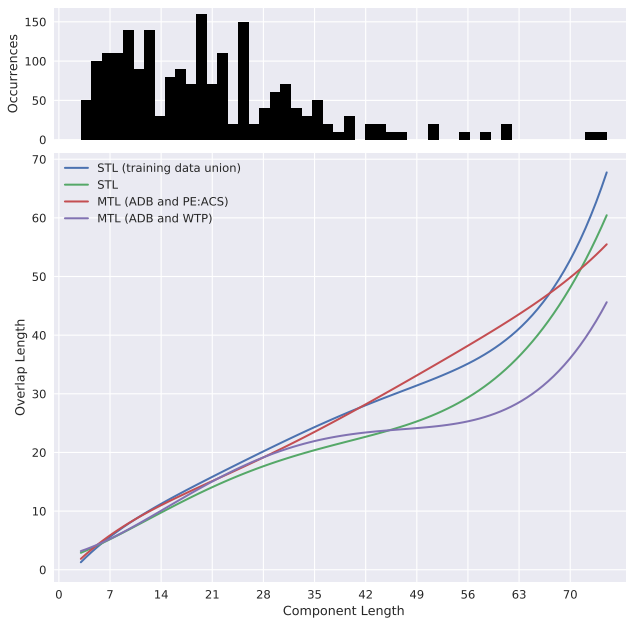
**(a)** AraucariaDB (ADB)

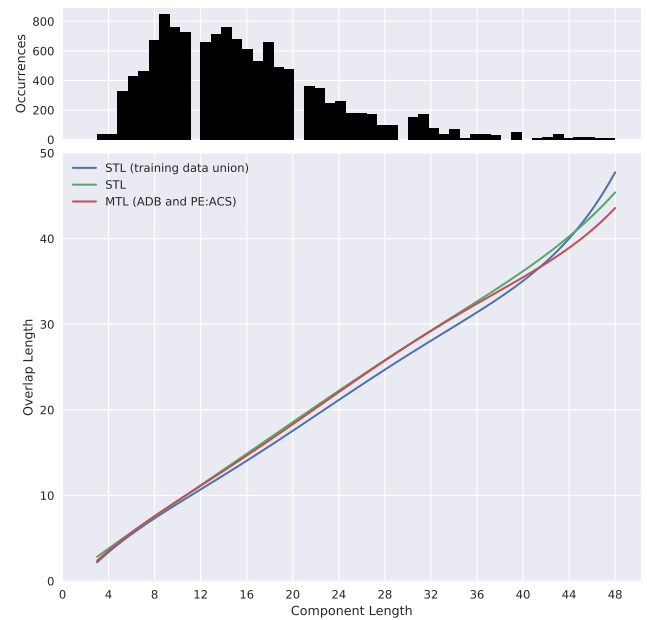
**(b)** Persuasive Essays (PE:ACS)

**Figure 6.9.:** Correlation between component length and correctness of predictions for the ADB and PE:ACS datasets analogous to figure 6.6. We provide a detailed explanation of this figure's creation process in appendix B.

When learning ACS on the ADB dataset alone, the MTL sequence tagging framework often fails to adhere to the BIO labeling scheme. Almost 60 % of all components contain inconsistencies. This is even worse than the inconsistencies registered for the SW dataset in the previous section. However, in the MTL setups, this score is reduced to less than 10 %. Since this reduction can also be observed with the training data union setup, the increased amount of training data seems to be the cause for this substantial improvement regarding the adherence to valid BIO labels.

The same analysis for the datasets PE:ACS and WTP, however, yields the following insights. In the STL setting, ACS on PE:ACS adheres almost perfectly to the BIO scheme with only 0.010 % of inconsistent components. This score is slightly increased in the MTL setup (0.750 %), but substantially increased in the training data union setup (24.870 %). On WTP, the fraction of inconsistent components is 68.080 %, 77.350 %, and 9.970 % for STL, training data union, and MTL respectively.

In fact, all these findings are consistent with the experiment results presented in table 5.17. For instance, the F1 score on the PE:ACS dataset is best for the STL setup, slightly decreased for the MTL setup, and substantially decreased for the training data union setup. Hence, the ability to model the label dependencies in the BIO scheme correctly is beneficial to the overall performance and MTL can have a positive effect on this ability, but may also be harmful for tasks which already have large quantities of training data.

### Multi-Task Grapheme-to-Phoneme Conversion

In the final experiment setting, MTL has only been beneficial for G2P conversion on the Combilex dataset and only with our MTL sequence tagging framework. With Reimers and Gurevych (2017a) architecture, STL has consistently outperformed MTL. One major reason for the differences in the MTL success is probably the origin of the auxiliary dataset. As explained in section 5.1.5, the word segmentations are derived from the Combilex dataset.

All MTL setups utilized all types of word segmentations as auxiliary tasks. To determine whether some word segmentations are more helpful than others, three additional experiment setups have been evaluated. However, no additional hyper-parameter optimization has been performed. The three new setups are $\text{G2P}_{morph}^{\text{Combilex}}$, $\text{G2P}_{syll}^{\text{Combilex}}$, and $\text{G2P}_{phon}^{\text{Combilex}}$. All of them are trained with the hyper-parameters of $\text{G2P}_{morph,syll,phon}^{\text{Combilex}}$ and on 10 % of the training data. The resulting word accuracies are 55.875 %, 56.868 %, and 56.025 % respectively. Hence, all word segmentations are similarly helpful; syllable segmentation is the most helpful auxiliary task and the combination of all auxiliary tasks still achieves the best results. The latter may change if the hyper-parameters of these ablation tests are tuned individually. This is left for future work.

The success of syllable segmentations is a highly useful insight because correct syllable splits can easily be generated using, for instance, the Knuth-Liang algorithm (Liang 1983) which is used in the TEX typesetting system (Knuth 1984) or a traditional dictionary can be used as a source of syllable segmentations. This would allow to repeat the G2P experiments of this thesis with generated word segmentations from the Celex or CMU dataset or a dictionary resource such as Wiktionary[26].

Intuitively, word segmentation information becomes more useful as words get longer because longer words are more likely to consist of multiple segments. This is particularly true for morphemic and syllable segmentation. Thus, such words should benefit most from the MTL setup. Indeed, this can be observed in figure 6.10b. As the words get longer, the average edit distance increases. This correlation is inherent to the edit distance metric. However, the *difference* in edit distance grows as well, thereby proving that MTL is increasingly useful for longer words. Although less prominent, this effect remained visible while we increased the training set size. For the CMU dataset (see figure 6.10a), this trend cannot be observed. The STL and MTL curves are almost parallel to each other. The model trained for this dataset seems incapable of utilizing the word segmentation information, i.e. the knowledge provided by the auxiliary tasks in the MTL setup, to its advantage. Instead, the performance deteriorates, but the difference in edit distance remains constant. Hence, the trend is specific to MTL.

---

[26] https://www.wiktionary.org/

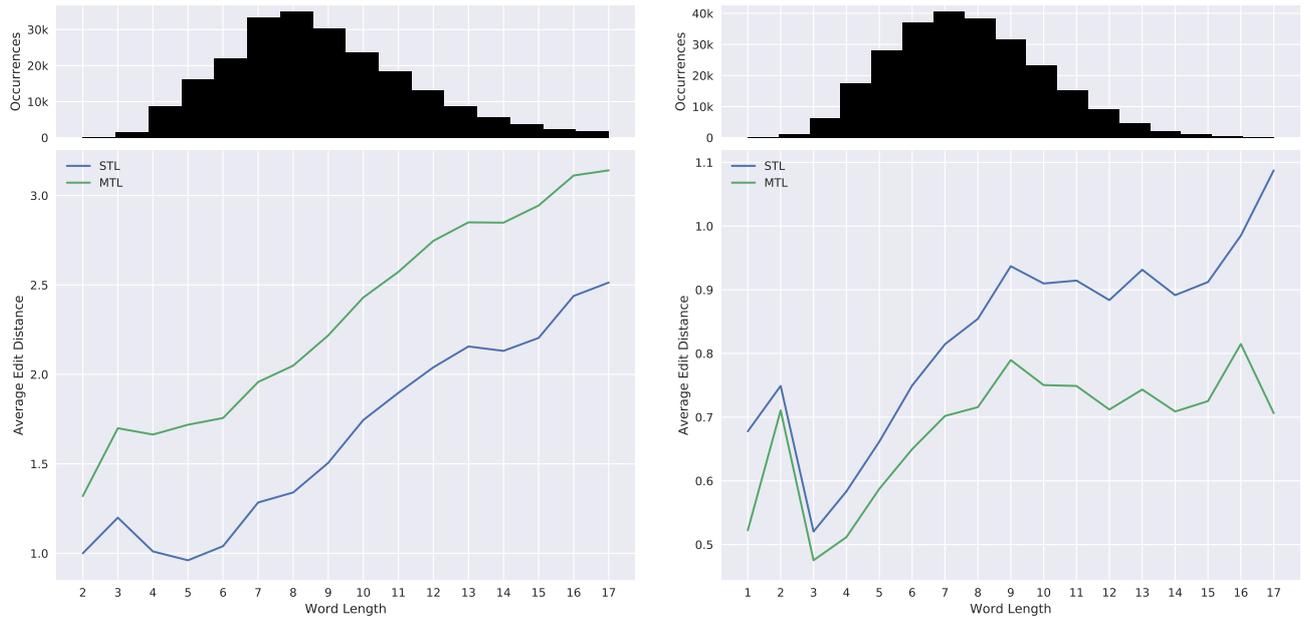

**Figure 6.10.:** Correlation between word length and average edit distance for the datasets CMU and Combilex. The models which are responsible for these predictions and the achieved edit distances have been trained on 10 % of their respective training datasets. Words with a length of more than 17 characters have not been visualized as they occur very infrequently.

Furthermore, we have investigated which words in the Combilex dataset have benefited from the MTL setup and have found that the word segmentation information, in fact, often helps to decide whether two graphemes should be converted to one or two phonemes. For instance, "dis-har-monic" is correctly converted to "d I s h A r m A n I k" by our MTL setup while our STL setup predicts "d I S A r m A n I k", i.e. "sh" is converted to the single phoneme "S". Given the word segmentation information (indicated by the hyphens), this conversion is not sensible. In contrast, our STL setup converts "diph-thongs" to "d I f t h A N z" instead of the correct "d I f T A N z" which has been predicted by our MTL setup. In this case, the split indicates that a conversion into two phonemes is less likely to be correct. The model which has been trained with MTL utilizes this knowledge. We obtained the syllable segmentations (hyphens) in these examples from Weis, Enz, and Schnorr (2003).

Finally, we analyze the comparably bad results on the CMU dataset. First of all, the split into training, development, and test dataset is flawed because the data was sorted alphabetically before the split and has not been shuffled in the process of creating the different splits. Therefore, the test dataset only contains words starting with S, T, U, V, ..., Z. This complicates generalization for the systems.

As shown in figure 6.3a, the variance in the CMU results for the MTL setup is high. When comparing the predictions of the best and the worst MTL results, the main source of errors is apparently the first phoneme. For instance, "SECTORS" is correctly converted to "S EH K T ER Z" by the best model and incorrectly converted to "B EH K T ER Z" by the worst model. Furthermore, the first phoneme is often completely omitted in the worst model, e.g. "SEEKING" is converted into "IY K IH NG" instead of "S IY K IH NG". When the best STL model is added to this comparison, a further error pattern emerges. An additional phoneme is prepended to the phoneme string, e.g "EH S IY K IH NG" for "SEEKING". These patterns indicate that the conversion of the first grapheme is difficult for the learner. This is probably a consequence of the flawed split which we have explained previously.

In addition to that, we have hypothesized that our MTL setup has intensified this systematic failure because all word segmentation binary strings start with a `0` as 1s are only used to represent intra-word segmentation boundaries (cf. section 5.1.5). Since a rerun of the MTL setup with auxiliary data where each word segmentation binary string starts with 1 instead of `0` further decreased the average word accuracy to 16.275 %, we rejected this hypothesis.

In future work, the CMU experiments should be repeated with a better split into training, development, and test sets, i.e. the data should be shuffled prior to the splitting. Moreover, the introduction of special "start of word" symbol (e.g. Bollmann and Søgaard 2016) could also help to solve the detected problems.

# 7 Conclusion

The motivation for this thesis was an investigation of the MTL paradigm based on the findings in related work and the results of the new MTL experiments which have been conducted in this thesis (cf. section 1.1). Given the results in section 5.6, MTL cannot be considered a *silver bullet*. It can be beneficial for a learner's performance, but there is no guarantee for success. However, such a guarantee has never been assumed and related work already showed that MTL may fail as well (cf. section 2.2). Instead, we have found indicators when, i.e. in which scenarios and for which types of tasks, MTL is helpful and have provided reasons why it increases or decreases the performance (cf chapter 6). One major benefit of MTL is its ability to fight data sparseness (cf. section 3.3.2). This has been confirmed by the experiments. MTL is particularly useful for tasks with a small number of training samples. Hence, MTL is a recommended technique for tasks that suffer from not having large quantities of training data. The runtime overhead due to MTL is not as severe in these cases, too.

**Choosing Auxiliary Tasks**

Even in a scenario where MTL can be successful, its success is not certain because the right choice of auxiliary tasks is crucial. This can be observed especially in the first experiment setting (cf. table 5.15). It is, however, difficult to find suitable auxiliary tasks. Moreover, one has to be careful not to resort to feature engineering since carefully selecting auxiliary tasks by hand utilizing domain-specific knowledge is just a new kind of feature engineering. Instead, the insights of Alonso and Plank (2017) and Bingel and Søgaard (2017) which have been confirmed in this thesis (cf. section 6.2.4) should be used. It is possible to find good auxiliary task by means of dataset properties or the STL learning curves.

In addition to that, natural subtasks have shown to be effective. They may serve as a good starting point when investigating MTL for a specific task given that the task has fairly complex labels which allow the derivation of simpler subtasks, of course. Since the data for natural subtasks can be generated automatically rather than requiring manual annotation or purchasing another dataset, the barrier of entry for MTL is very low. While natural subtasks are particularly promising auxiliary task candidates, the barrier of entry for MTL is also low in general. The main cost factor is the increased processing time. As it is possible to use any existing dataset, e.g. datasets from shared tasks, and even leverage datasets from other languages and domains, data sparseness is not an obstacle to MTL.

**Applicability of MTL**

In the related work (cf. section 2.2), MTL has already been shown to be successful across a wide range of tasks. The experiments conducted within the scope of this thesis extend this range. In each experiment setting, MTL was at least to some degree successful. It was not only helpful for pure sequence tagging tasks, but also generalized to an S2S task.

Besides the theoretical reasons for improvements resulting from the application of MTL such as leveraging multiple tasks as mutual sources of bias and reducing the likelihood of overfitting due to a regularizing effect (cf. section 3.3), we found that MTL allows to improve the performance for longer input sequences. This has been very helpful for the analyzed tasks as `AM` and `ES` datasets often contain argumentative components/propositional units which consist of a long sequence of tokens. Similarly, MTL was able to perform better, i.e. more accurate, conversions of graphemes to phonemes for long words. Furthermore, MTL can be helpful to escape local minima by leveraging the training signal in the auxiliary task. Finally, MTL was able to help the learner to better model label dependencies which can be observed as an improved adherence to the BIO scheme, i.e. there are less violations of the BIO scheme with MTL.

Since label dependencies are important for sequence tagging tasks, this success has a positive effect on the performance.

# 8 Future Work

In the previous chapters of this thesis, we have pointed out several tasks for future work. We discuss only the most important of them in this chapter and extend this discussion by new ideas for future work.

**Improvement of the MTL Sequence Tagging Framework**

Our MTL sequence tagging framework implements many features which have proven successful in related work. Nevertheless, there are still various features which may have a positive impact on the performance of the system. In the current implementation, character-level information can only be extracted by a bidirectional LSTM network, i.e. utilizing the approach of Lample et al. (2016). Reimers and Gurevych (2017a), however, found that the CNN-based approach by Ma and E. Hovy (2016) performs equally good, but is more efficient w.r.t. the computational requirements. Hence, implementing the CNN approach might be helpful to reduce the runtime of the MTL sequence tagging framework. Instead of first-order CRFs, higher-order CRFs could be used. This would allow to model dependencies between arbitrarily many consecutive labels (cf. section 2.1.1), thereby being more effective in preventing violations of the BIO scheme (cf. section 6.2.6), for instance.

Label embeddings which we have discussed in section 3.3.4 could also prove effective as they allow tasks that are terminated on a higher layer to use the prediction results of lower-level tasks. This might yield performance improvements. At first, a simpler solution without additional embedding layers could suffice to prove the effectiveness of using label information from lower-level tasks. This solution can be implemented by feeding the output of a softmax classifier for a lower-level task to the subsequent shared layers. As explained in section 4.3.2, our MTL sequence tagging framework uses hard parameter sharing. This is not optimal as the learner cannot distinguish task-specific from task-invariant features (cf. section 2.2.3). Adapting either the adversarial loss and orthogonality constraints approach by Liu, Qiu, and X. Huang (2017) or the "Sluice Networks" approach by Ruder et al. (2017) or even both can resolve this weakness of our system.

The MTL sequence tagging framework's applicability is limited to sequence tagging problems and problems which can be framed as sequence tagging problems such as G2P conversion. Utilizing other types of machine learning problems as auxiliary tasks, however, might prove beneficial as in the architecture by Kaiser et al. (2017) (cf. section 2.2.4). For example, sentence-level auxiliary tasks could be used to aid token-level main tasks.

**Other Architectural Approaches**

In the related work, most authors use RNNs for sequence tagging tasks. This approach has also been chosen for the MTL sequence tagging framework (cf. section 4.3.2). It would be interesting to compare the MTL performance with a similar system which, however, is based on a CNN. A CNN-based sequence tagging system has been proposed by Strubell et al. (2017), but it does not support MTL. To the best of our knowledge, a CNN-based, MTL-capable sequence tagging system would be entirely novel. It may prove superior to the RNN-based variants for scenarios in which runtime is essential.

Comparing MTL to a pipeline approach (cf. section 3.3.6) may yield interesting results. Such an approach, however, may suffer from error propagation as the auxiliary tasks are solved independently (Eger, Daxenberger, and Gurevych 2017). This comparison becomes even more interesting when the MTL sequence tagging framework implements label embeddings as they effectively provide the main task with the same input features as the pipeline approach, i.e. tokens and labels of lower-level tasks, but without solving the tasks individually, thereby mitigating the problem of error propagation.

**More Experiments and Analysis**

First of all, more experiment setups can be evaluated. In the multi-task `AM` setting, all natural subtasks were used together. Evaluating the performance for each subtask individually would allow to decide which subtask or which combination of subtasks works best for `AM`. The RST-DT dataset has only been used together with the PDTB dataset which resulted in the worst MTL performance for `AM`. An evaluation of `AM` with discourse parsing on RST-DT as the sole auxiliary task is necessary for a final conclusion regarding whether or not RST-DT is solely responsible for the performance deterioration in the $\text{AM}^{\text{PE}}_{\text{PDTB,RST-DT}}$ setup. Moreover, repeating the multi-task `G2P` conversion experiments for Celex and CMU with word segmentation data from another source could yield performance improvements via MTL in contrast to the current results.

As mentioned in section 5.5, no individual hyper-parameter optimization has been performed for the experiment setups with varying amounts of training data. This is not optimal as setups with more training samples are likely to require different hyper-parameter configurations. Hence, the experiments should be repeated with individual hyper-parameter optimizations. Furthermore, the other experiment settings, i.e. multi-task `AM` and cross-domain `ACS`, could be evaluated with varying amounts of training data, too. Thereby, the trend of a decreasing effect of MTL when increasing the amount of training data could be verified.

Finally, more research regarding the suitability of auxiliary tasks is necessary in order to make MTL more feasible and prevent the threat of resorting to feature engineering (cf. chapter 7) as suitable auxiliary tasks can be selected automatically, e.g. by dataset properties, without evaluating them first.

# A  Argumentation Mining Example

Stab and Gurevych (2017) provide the following example which contains all types of argumentative components, i.e. major claim, claim, and premise:

> Ever since researchers at the Roslin Institute in Edinburgh cloned an adult sheep, there has been an ongoing debate about whether cloning technology is morally and ethically right or not. Some people argue for and others against and there is still no agreement whether cloning technology should be permitted. However, as far as I'm concerned, **[cloning is an important technology for humankind]**$_{MajorClaim1}$ since [it would be very useful for developing novel cures]$_{Claim1}$.
>
> First, [cloning will be beneficial for many people who are in need of organ transplants]$_{Claim2}$. [Clone organs will match perfectly to the blood group and tissue of patients]$_{Premise1}$ since [they can be raised from cloned stem cells of the patient]$_{Premise2}$. In addition, [it shortens the healing process]$_{Premise3}$. Usually, [it is very rare to find an appropriate organ donor]$_{Premise4}$ and [by using cloning in order to raise required organs the waiting time can be shortened tremendously]$_{Premise5}$.
>
> [...]
>
> Admittedly, [cloning could be misused for military purposes]$_{Claim5}$. For example, [it could be used to manipulate human genes in order to create obedient soldiers with extraordinary abilities]$_{Premise9}$. However, because [moral and ethical values are internationally shared]$_{Premise10}$, [it is very unlikely that cloning will be misused for militant objectives]$_{Premise11}$.

Claim 1 and 2 support the the major claim. Premise 1 and 3 support claim 2 while premise 2 supports premise 1 and premise 4 and 5 support premise 3. Claim 5 attacks the major claim and is supported by premise 9. Premise 11, however, attacks claim 5, i.e. it defends the author's stance. Premise 10 supports premise 11. Compared to Stab and Gurevych (2017), we abridged this example, but the excerpt shown here is sufficient to understand the AM structure within the scope of this thesis.

# B Span Overlap Visualization

While the meaning of the figures 6.6, 6.8a, 6.8b, 6.9a, and 6.9b is easy to understand given the figures' captions, their creation process is non-trivial. Therefore, we describe it in this chapter more detailed. The procedure to create these visualization which we refer to as *span overlap visualizations* in the following is as follows. Let $g_{a,b}$ be an argumentative component from the gold data $\mathbb{G}$ spanning from index $a$ to index $b$ in the text with $a < b$ and $p_{c,d}$ an argumentative component spanning from $c$ to $d$ with $c < d$ from the predictions $\mathbb{P}$. The subscripts are omitted when this causes no confusion.

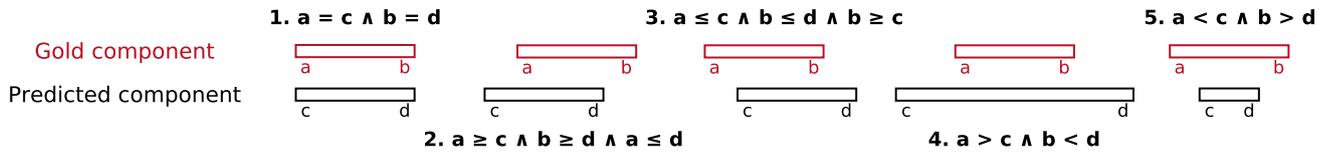

**Figure B.1.:** All overlap scenarios for a gold component $g_{a,b}$ starting at index $a$ and ending at index $b$ and a predicted component $p_{c,d}$ starting at $c$ and ending at $d$.

We say that $g_{a,b}$ and $p_{c,d}$ *overlap* if either of the following conditions is satisfied:

1. $cond_1 : a = c \land b = d$

2. $cond_2 : a \geq c \land b \geq d \land a \leq d$

3. $cond_3 : a \leq c \land b \leq d \land b \geq c$

4. $cond_4 : a > c \land b < d$

5. $cond_5 : a < c \land b > d$

Figure B.1 visualizes the five scenarios which are represented by these conditions. If, for instance, $a \geq d$, the components do *not overlap*. For a gold component $g \in \mathbb{G}$, $\Omega(g)$ is the set of all predicted components $p \in \mathbb{P}$ which *overlap* with $g$.

Given the `AM` label definition from section 3.5.1, let $\tau : \mathbb{G} \cup \mathbb{P} \to \{(t,d,s) \mid (b,t,d^*,s) \in Y\}$ be the function which returns a component's `AM` label without the BIO prefix as these labels are component- and not token-level labels. Note that $d^*$ refers to the absolute distance between components while $d$ in the original definition of $Y$ refers to the relative distance. The function $\tau$ differs slightly from experiment setting to experiment setting as they are using different label set. Its definition, however is analogous. It is just a projection from the token-level label to the component-level label.

The *overlap length function* $\omega$ is then defined as in equation (B.1). It utilizes the *length function* specified in equation (B.2) which calculates the overlap length from the start and end indices of two components.

$$\omega : \mathbb{G} \times \mathbb{P} \to \mathbb{N}, \ g_{a,b}, p_{c,d} \mapsto \omega(g_{a,b}, p_{c,d}) = \begin{cases} \text{length}(a,b,c,d) & \text{if } \tau(g_{a,b}) = \tau(p_{c,d}) \\ 0 & \text{else} \end{cases} \quad (B.1)$$

$$\text{length}: \mathbb{N}^4 \to \mathbb{N}, \ a,b,c,d \mapsto \text{length}(a,b,c,d) = \begin{cases} b-a & \text{if } cond_1 \\ d-a & \text{if } cond_2 \\ b-c & \text{if } cond_3 \\ b-a & \text{if } cond_4 \\ d-c & \text{if } cond_5 \\ 0 & \text{else} \end{cases} \quad \text{(B.2)}$$

Finally, equation (B.3) defines $\mathbb{O}$ as the multiset of all tuples containing the length of a gold component and the length of the longest overlap with a correct predicted component.

$$\mathbb{O} := \left\{ (x,y) = \left( \overbrace{\text{length}(a,b,a,b)}^{\text{Component Length}}, \overbrace{\max_{p_{c,d} \in \Omega(g_{a,b})} \omega(g_{a,b}, p_{c,d})}^{\text{Overlap Length}} \right) \ \Bigg| \ g_{a,b} \in \mathbb{G} \right\} \quad \text{(B.3)}$$

The multiset $\mathbb{O}$ is calculated for each experiment setup and represented in the span overlap visualizations. Instead of plotting each tuple onto a Cartesian coordinate plane resulting in a cluttered scatterplot, we have decided to fit a polynomial of degree 5 to the tuples in $\mathbb{O}$ using NumPy's `chebfit`[27]. We have verified that these polynomials agree with the actual data distribution.

---

[27] https://docs.scipy.org/doc/numpy-1.13.0/reference/generated/numpy.polynomial.chebyshev.chebfit.html

86